\documentclass{article}

\usepackage[nonatbib,final]{neurips_data_2023}

\usepackage[utf8]{inputenc} 
\usepackage[T1]{fontenc}    
\usepackage{hyperref}       
\usepackage{url}            
\usepackage{booktabs}       
\usepackage{amsfonts}       
\usepackage{nicefrac}       
\usepackage{microtype}      

\usepackage[capitalize]{cleveref}
\crefname{section}{Sec.}{Secs.}
\Crefname{section}{Section}{Sections}
\Crefname{table}{Table}{Tables}
\crefname{table}{Tab.}{Tabs.}

\usepackage[dvipsnames,table]{xcolor}
\usepackage{multirow}
\usepackage{caption}
\usepackage{enumitem}
\usepackage{graphicx}
\usepackage{arydshln}

\usepackage{mdframed}
\mdfdefinestyle{MyFrame}{%
    outerlinewidth=2pt,
    roundcorner=10pt,
    innerbottommargin=0.5\baselineskip,
    innertopmargin=0.5\baselineskip,
    splittopskip=0.5\baselineskip,
    }

\newcommand{\crowdsourced}{crowd-sourced{ }}

\newcommand{\etal}{et al.}
\newcommand\sbullet[1][.75]{\mathbin{\vcenter{\hbox{\scalebox{#1}{$\bullet$}}}}}

\newcolumntype{L}[1]{>{\raggedright\let\newline\\\arraybackslash\hspace{0pt}}p{#1}}

\newcommand{\new}[1]{{#1}}

\newcommand{\aaron}[1]{{\color{pink} Aaron: #1}}

\newcommand{\vikram}[1]{{\color{ForestGreen} Vikram: #1}}

\newcommand{\sectioncolor}{violet}

\newcommand{\smallsec}[1]{\vspace{1pt} \noindent \textbf{#1.}}
\newcommand{\appen}{{GeoDE }}
\newcommand{\appenn}{GeoDE}

\begin{document}

\title{GeoDE: a Geographically Diverse Evaluation Dataset for Object Recognition}

\author{Vikram V. Ramaswamy$ ^1$, \ \  
Sing Yu Lin$ ^1$, \ \    Dora Zhao$ ^2$*, \ \   Aaron B. Adcock$ ^3$, \\  \textbf{Laurens van der Maaten}$ ^3$,  \ \  \textbf{Deepti Ghadiyaram}$ ^4$,  \ \   \textbf{Olga Russakovsky}$ ^1$ \\ \\
$^1$Princeton University \ \ \ \  
$^2$Stanford University \ \ \ \ $^3$Meta AI \ \ \ \ $^4$Runway\\ 
\footnotesize{*Work done as a student at Princeton University}
}
\maketitle

\begin{abstract}
   Current dataset collection methods typically scrape large amounts of data from the web. While this technique is extremely scalable, data collected in this way tends to reinforce stereotypical biases, can contain personally identifiable information, and typically originates from Europe and North America. In this work, we rethink the dataset collection paradigm and introduce \appen, a geographically diverse dataset with 61,940 images from 40 classes and 6 world regions, with no personally identifiable information, collected by soliciting images from people around the world. We analyse \appen to understand differences in images collected in this manner compared to web-scraping. We demonstrate its use as both an evaluation and training dataset, allowing us to highlight and begin to mitigate the shortcomings in current models, despite GeoDE's relatively small size. We release the full dataset and code at \url{https://geodiverse-data-collection.cs.princeton.edu/} 
\end{abstract}

\section{Introduction}
\label{sec:intro}

The creation of large-scale image datasets has enabled advances in the performance of computer vision models. Previously limited by internal manual collection efforts~\cite{feifei2004caltech101,griffin2007caltech,Everingham10pascal}, in the past 15 years the size of these datasets has rapidly grown. This growth has been empowered by a new data collection framework: scraping web images at scale. These images are either human-labelled (e.g., ImageNet~\cite{imagenet_cvpr09,imagenet_ijcv14}), use tags (e.g., CLIP-400M~\cite{radford_clip}) or use self-supervision (e.g., PASS~\cite{asano21pass}). 

However, these web-scraped datasets come with their downsides. One of these downsides is that these datasets can often contain pernicious social and cultural biases. For example, gender and racial biases can manifest through underrepresentation and/or through stereotypical depictions of certain demographic groups~\cite{noble2018algorithms,buolamwini18agender,zhao2017men,revisetool_extended,birhane2021large}. There is also \textit{geographic bias}: works of e.g., Shankar \etal~\cite{shankar2017no} and de Vries \etal~\cite{devries2019objectrecognition} show that web-scraped datasets consist of images mostly from North America and Western Europe. 

The other common downsides are copyright, consent and compensation. Dataset creators frequently do not obtain full permission of the content creators and of the people featured in the content~\cite{birhane2021large}.\footnote{While images used are sometimes under the most permissive Creative Commons license, it is unclear if creators know the full impacts of their images being used in the training of large scale models.} 
While annotators are compensated, content creators and image subjects rarely are~\cite{birhane2021large}.
Though there have been efforts to balance datasets~\cite{dubey2021adaptive}, clean datasets~\cite{yang2020filtering}, and protect privacy of depicted subjects by blurring~\cite{yang2021imagenetfaces}, methods that rely on web-scraping cannot fully eliminate these issues~\cite{birhane2021large,jo2000archives}. 

To tackle these issues, an exciting new dataset DollarStreet~\cite{DollarStreet} was recently introduced (licence: CC). Instead of web-scraping, DollarStreet sources data from the Gapminder foundation. It comprises of images taken by volunteer and professional photographers in different countries to illustrate households with different income statuses. This results in 38,479 images from 63 different countries, tagged with 289 labels. DollarStreet overcomes issues of consent, and is in many ways the first truly  geographically diverse dataset (Tab.~\ref{tab:dsets}). 

{However, DollarStreet curates a computer vision dataset from images that have \emph{previously been collected} and released on the web. In contrast, we present an alternative geographically-diverse data collection process (more details in Sec.~\ref{sec:dset_choices}), which allows us to explicitly target a different object-centric image distribution. 




Further, with the advent of multimodal foundation models~\cite{radford_clip,dalle,gpt4,touvron2023llama,driess2023palme}, revising the idea of manual collection of test datasets may be in order. With these models, direct access to the training data is frequently impossible, although we know they were trained on trillions of web-scraped examples. Thus, any dataset comprising of previously available images may actually have been included in the model training, violating the core machine learning tenet of separate train/test splits. 

\smallsec{Contributions} We collected a \textbf{geographically diverse dataset of common objects} which overcomes many issues described above. Concretely, we \emph{commission} photos of different objects from people across the world from Appen (\url{www.appen.com})'s global workforce. 
This  naturally resolves \textbf{consent} concerns, similar to DollarStreet~\cite{DollarStreet} and ensures that the data is (at least temporarily)  an \textbf{unseen test set}. 
We \textbf{own the copyright} to all of the images in \appenn, \textbf{have explicit permission} from creators to use these images for computer vision, verified that the images \textbf{do not show identifiable people} or other personally identifiable information (PII), and ensured \textbf{ compensation} to the content creators.

\begin{figure}[t]
    \centering
    \resizebox{\linewidth}{!}{
    \begin{tabular}{cc}
    GeoYFCC distribution~\cite{dubey2021adaptive} &     \appen distribution (ours)\\
        \includegraphics[width=0.65\linewidth]{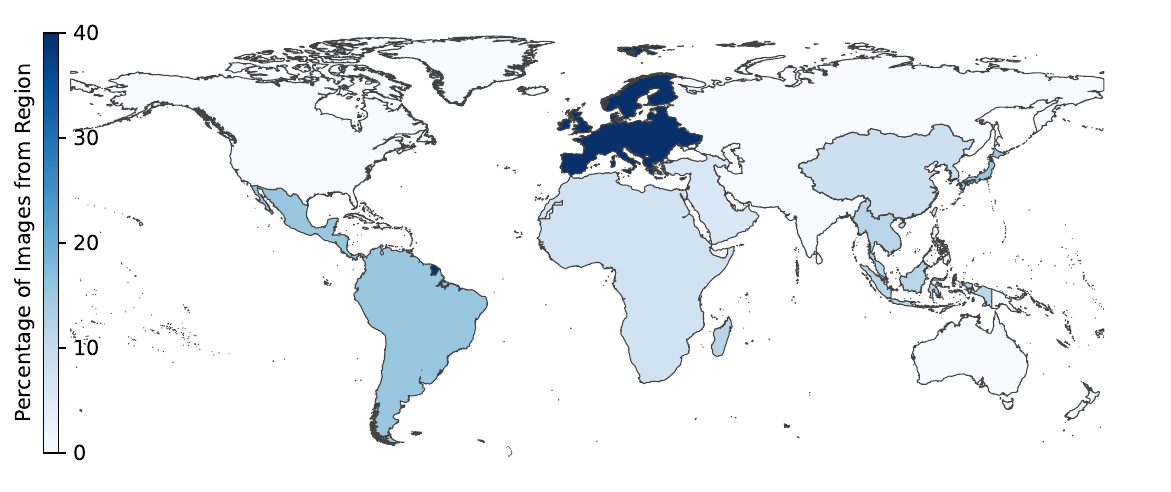} &
    \includegraphics[width=0.65\linewidth]{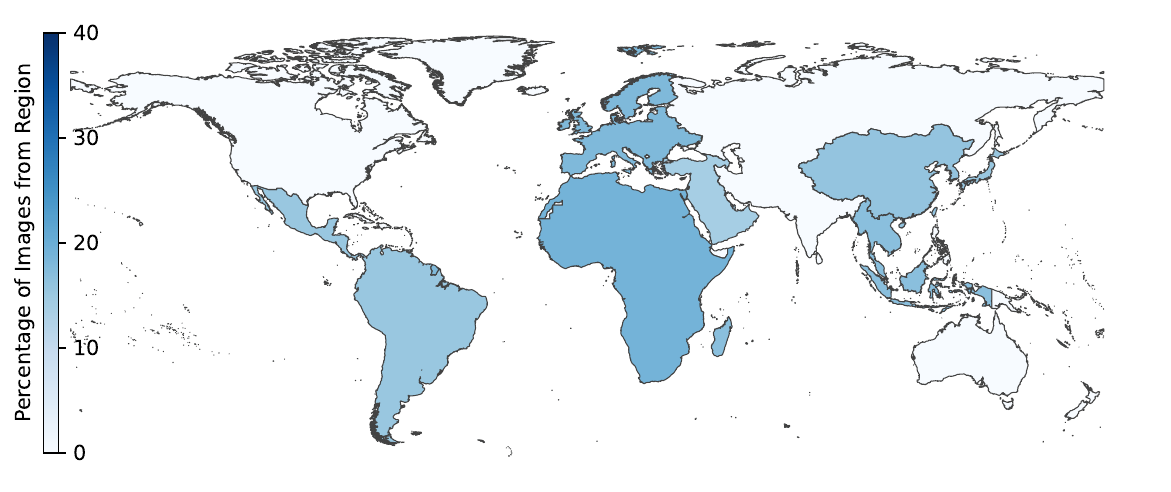}
    \end{tabular}}
    \caption{We construct a geographically diverse dataset \appen that is approximately balanced across 6 world regions. We visualize the images per region, and compare our distribution (\emph{right}) to that of a previously created diverse dataset GeoYFCC~\cite{dubey2021adaptive} (\emph{left}).}
    \label{fig:pullfig}
\end{figure}

    The resulting {Geo}graphically Diverse Evaluation (\appenn) dataset\footnote{\appen is owned and maintained by Princeton; Meta AI was not involved with the data collection.} contains 61,940 images roughly balanced across 40 object categories and 6 geographic regions. We find that the object recognition problem becomes surprisingly challenging since the GeoDE images represent the \textbf{diverse appearance} of common objects across six global regions: Africa, the Americas, East Asia, Europe, Southeast Asia, and West Asia. 
    Similar to de Vries \etal~\cite{devries2019objectrecognition}, we show that modern object recognition models perform poorly on recognizing objects from Africa, East Asia, and SE Asia. Augmenting current training datasets (like ImageNet~\cite{imagenet_cvpr09,imagenet_ijcv14}) with training images from \appen yields an improvement of 9\% on DollarStreet~\cite{DollarStreet} and 21\% on the test split of \appen. 
    
    Further, requesting images with specific content \textbf{mitigated selection bias}: web-scraped images are typically uploaded by creators with different incentives, e.g. to generate engagement with exciting/unique content~\cite{Sigurdsson2016Charades}, and disincentives for mundane everyday content.
    We show that the distribution of images in \appen is different to that in other datasets, even when controlling for world region and object class.   
    


Unfortunately, a key drawback of our method is the \textbf{\emph{cost}}, which is partially the result of aiming for fair compensation to the content creators and curators. The method  can also lead to other biases (e.g., lack of economic diversity, since workers are required to have a smart phone). However, we demonstrate that even small amounts of data collected in this way can be beneficial in  remedying some of the concerns with large-scale web-scraped datasets. 
Our work challenges the current paradigm for dataset collection and illustrates the process of manually curating image datasets. In addition, we introduce the \appen{} dataset which offers a more geodiverse and ethically created alternative for object recognition. 
Data and code can be found at \url{https://geodiverse-data-collection.cs.princeton.edu/}

\section{Related Work}
\label{sec:related}

\begin{table*}[t]
    \centering
    \caption{We compare approaches to dataset collection, along with the distribution and sizes of each. Although \appen{} is smaller than standard datasets, we ensure that images are sourced with permission, contain no identifiable people, and are balanced across both regions and object classes.} 
    \label{tab:dsets}
        \resizebox{\linewidth}{!}{
    \begin{tabular}{p{0.12\linewidth}L{0.18\linewidth}L{0.35\linewidth}L{0.2\linewidth}L{0.17\linewidth}}
    \toprule
     Dataset & Size;\newline distribution & Collection process; \newline annotation process & Geographic coverage & Personally Identifiable Info (PII) \\ \midrule
     ImageNet \newline \cite{imagenet_cvpr09,imagenet_ijcv14} & 14.2M; mostly balanced \newline across classes & Scraped images from the web based on the class label; crowd-sourced annotations & Mostly North America \& Western Europe \cite{shankar2017no} & Contains people, subset with faces blurred~\cite{yang2021faceobfuscation} \\\arrayrulecolor{gray}\midrule
PASS\newline\cite{asano21pass} & 1.4M; N/A (no  labels)  & Random images from Flickr; no annotations & Flickr, thus mostly North America \& Western Europe & No people 
\\\midrule
     OpenImages \newline \cite{OpenImages} & 9.1M ;  long-tailed class distribution  & Flickr images with CC-BY  licenses; automatic labels with some human verification &  Mostly North America \& Western Europe~\cite{shankar2017no} & Contains people \\ \midrule
     OpenImages Extended\newline\cite{OpenImagesExtended}& 478K; long-tailed class distribution  & Crowd-sourced gamified app to collect images; automatic labels and manual descriptions & More than $80\%$ of images are from India  & People are blurred \\\midrule
     GeoYFCC\newline\cite{dubey2021adaptive} & 330K; long-tailed class distribution & Flickr images subsampled to be geodiverse; noisy tags& 62 countries, but  concentrated in Europe (Fig.~\ref{fig:pullfig}) & Contains people  \\\midrule
DollarStreet\newline\cite{DollarStreet} & 38,479; mostly balanced across topics & Images by professional and volunteer photographers; manual labels including household income &  63 countries in Africa, America, Asia \& Europe & Yes, with  permission \\ 
     \midrule
     \appen{} \newline (ours) & 61,940; balanced across classes and regions & Crowd-sourced collection using paid workers; manual annotation  & Even coverage over six geographical regions (\cref{tab:countries}) & No identifiable people and no other PII  \\ \bottomrule
    \end{tabular}}
    
\end{table*}

There are three key research directions that inspired this work. The first is the call to increase geographic diversity in visual datasets~\cite{shankar2017no,devries2019objectrecognition}.  In response, there have been attempts to construct such datasets~\cite{dubey2021adaptive,OpenImagesExtended}, summarized in \cref{tab:dsets}. However, these datasets are still geographically concentrated (in Europe for GeoYFCC~\cite{dubey2021adaptive} or India for OpenImages Extended~\cite{OpenImagesExtended}). The dataset most similar to ours is \textbf{DollarStreet}~\cite{DollarStreet}, also mentioned in the introduction. Both DollarStreet and our GeoDe were collected to improve geographic diversity of image datasets, are relatively small scale datasets (62k for GeoDE, 39k for DollarStreet), and are collected through crowd-sourcing. However, DollarStreet was repurposed as a computer vision dataset by curating the images through GapMinder, a non-profit organization that collected these images to showcase differences in how people live around the world. Thus, the images in DollarStreet were collected through a more social science perspective: for example, images are collected to showcase different everyday actions such as ``washing hands'' as opposed to objects such as ``hand soap''. 
On the other hand, GeoDE is more focused on understanding how objects across the world are visually different.





Second, in using participants to \emph{generate} visual content, we follow  video datasets  Charades~\cite{Sigurdsson2016Charades}, Epic Kitchens~\cite{damen2018scaling} and Ego4d~\cite{graumen2021ego4d}. 
We similarly leverage paid workers to provide examples that fall outside the common web-scraped distribution. 
However, we differ in that our key goal is to ensure geographic diversity. This poses unique challenges in recruitment and dataset scope (more in Sec.~\ref{sec:dset_choices}). 

Finally, in our data collection efforts we take into account the extensive literature around selection bias in computer vision datasets~\cite{torralba2011unbiased,buolamwini18agender,zhao2017men,revisetool_extended,stock2018convnets,dulhanty2019auditing,yang2020towards,birhane2021large} and ensure that our dataset is collected responsibly, with attention to privacy, consent, copyright and worker compensation~\cite{birhane2021large}.

\section{Collecting \appen}
\label{sec:dset_choices}

We describe our data collection process, including our selection of object classes and world regions. 

\smallsec{Selecting the object classes}
We focus on object classes that are likely to be visually distinct in different parts of the world. Selecting such objects is a chicken-and-egg problem: without a geographically diverse dataset at our disposal, it is unclear which objects are diverse. We adopt a number of heuristics using existing datasets to find a plausible set. We use simple computer vision techniques (linear models and visual clustering, using features extracted from self-supervised PASS-pretrained models~\cite{asano21pass}) along with manual examination to identify a set of candidate tags from DollarStreet~\cite{DollarStreet} and GeoYFCC~\cite{dubey2021adaptive} (e.g., ``chili,'' ``footstool,'' ``stove''). To prune these tags, we remove those that are not objects (e.g., ``arctic'', ``descent''), remove wild animals not found in all regions (e.g., ``gnu'', ``camel'') and group variants of objects (e.g., ``stupa'', ``temple'', ``church'' and ``mosque'').  The final set of objects is in \cref{tab:objs}, and the full process is in the supp. mat.

\begin{table}[t]
    \centering
        \caption{\appen consists of 40 object classes, loosely organized into 4 groups.}
\label{tab:objs}
    \resizebox{\linewidth}{!}{
    \begin{tabular}{L{0.25\linewidth} p{3pt} L{0.25\linewidth} p{3pt} L{0.25\linewidth} p{3pt} L{0.25\linewidth}}
    \toprule
     \emph{Indoor common} && \emph{Indoor rare} && \emph{Outdoor common} && \emph{Outdoor rare} \\ \midrule
bag, chair, dustbin, hairbrush/comb, hand soap, hat, light fixture, light switch, toothbrush,  toothpaste/toothpowder  && candle, cleaning equipment, cooking pot, jug,  lighter, medicine, plate of food, spices, stove, toy && backyard, car, fence, front door, house, road sign, streetlight/lantern, tree, truck, waste container && bicycle, boat, bus, dog, flag, monument,  religious building, stall, 
storefront, wheelbarrow  \\ \bottomrule
      
    \end{tabular}}

\end{table}

\begin{table}[t]
    \centering
    \caption{\appen consists of images from six world regions. Within each, there are 3-4 countries contributing to most of the images. Participants from other countries within the region were accepted.}
    \label{tab:countries}
    \resizebox{\linewidth}{!}{
    {\small
    \begin{tabular}{p{0.1\linewidth}@{}p{0.35\linewidth}p{0.1\linewidth}@{}p{0.35\linewidth}}
    \toprule
    \emph{West Asia:}& Saudi Arabia, UAE, Turkey &
\emph{Africa:}& Egypt, Nigeria, South Africa \\ 
\emph{East Asia:}& China, Japan, South Korea &
\emph{SE Asia:}& Indonesia, Philippines, Thailand \\ 
\emph{Americas:} & Argentina, Colombia, Mexico &
\emph{Europe:} & Italy, Romania, Spain, United Kingdom \\
\bottomrule
\end{tabular}
}}
    
\end{table}

\smallsec{Selecting diverse geographic regions} 
We chose six regions: Africa, Central and South America (``Americas''), East Asia, Europe, SouthEast (``SE'') Asia and West Asia. Within each, we targeted 3-4 countries (\cref{tab:countries}). The regions were chosen due to the lack of available images from them in most public datasets~\cite{shankar2017no,devries2019objectrecognition,revisetool_eccv}; the countries were chosen based on the presence of participants within Appen's database\footnote{{While we do not currently find this to be the case empirically, we acknowledge that the regions themselves are quite broad, and certain objects might look different within the region.}}. We obtain a roughly even distribution of images across each class and region pair. 

\smallsec{Image collection and worker demographics}
Workers were asked to upload images for a given object class (Fig~\ref{fig:inst_demographics}). There were 4,500+ workers, representing a range of genders, ages and races (see supp. mat.). All images submitted were manually checked by Appen's quality assurance (QA) team.


\begin{figure}[t]
    \centering

\fbox{%
    \parbox{0.95\linewidth}{%
    {\footnotesize
        \textbf{\emph{General Instructions}} \\
        In this task, you will submit up to \textbf{3 photos} of the \textbf{same type of object} (e.g., upload 3 photos of \textbf{3 different bags}; please \textbf{do not} upload 3 photos of the same bag from different angles).
\begin{enumerate}[topsep=0pt,itemsep=0ex,partopsep=0ex,parsep=0ex,leftmargin=15pt]
    \item Please make sure the location function is enabled for the camera.
    \item The photo resolution should be at least 640 x 480.
    \item All images should be new photos captured with Appen Mobile.
\item Please make sure it's a single object per image.
\item Please make sure it's a well-lit environment and the object is clearly visible in the photos.
\item Please make the object occupy at least 25\% of the image.
\item Objects captured are foregrounded and not occluded.
\item Objects should not be blurred, e.g., motion blur.
\item No effects or filters added (cropping is acceptable).
\item Please try to avoid capturing people in the images (it's OK if people are blurry in the background and far from the camera).
\item Please try to avoid capturing vehicle license plates in images.
\end{enumerate}
    }}%
    }
    
    \caption{Image collection instructions given to workers.}
    \label{fig:inst_demographics}
\end{figure}

{

\section{Lessons learned from collecting \appen}

A key part of this study was to understand if manually taking photographs is a viable alternative to web-scraping. In this section we detail the lessons learned in constructing the \appen dataset.

\smallsec{Getting sufficient images of all object classes}
While some object classes expectedly proved more difficult than others (e.g., ``monument'' or ``flag'' were simply hard for workers to find), others surprised us. For example, ``stove'' was originally underrepresented until the definition was clarified to ``any cooking surface either electric, gas, induction.'' Workers' perception of their cooking appliance as ``stove'' varied, highlighting a vocabulary challenge unique to geographically diverse data collection.


\smallsec{Multiple copies of images} 
Two most common types of error were incorrect images (i.e, not belonging to the class selected) and multiple copies of an image. The QA team found that participants often submitted multiple copies of the same object instance from different angles despite instructions not to do so. Workers also sometimes submitted very similar objects (examples in supp. mat.), for example, three hats by the same worker with slightly varying colors). We filtered out such images. 

\smallsec{Multiple objects per image} For some of the (especially larger and outdoor) object categories, it was difficult to ensure that other objects were not present, particularly ``trees''. For example, we found that images of ``fences'' often have ``trees'' present, and it was not always possible to discern between objects in the foreground and background (examples in supp. mat.). Thus, we requested that images that had a significant portion of the image covered by trees be explicitly tagged. These additional annotations can be used to filter and remove such images and/or to analyze errors made by a model. 

\smallsec{Other} Beyond these, the rest of the data collection went smoothly. Following instructions, only 0.78\% images contained identifiable information. Some images contain non-identifiable incidental people in the background (especially for larger object classes, like ``monument''). All such images are tagged in \appenn. We were also able to 
ensure that the number of images per region is roughly equal, although it was harder to obtain an even number of images per country within each region.


\smallsec{Cost} Collecting images in this way was expensive: each image cost roughly \$0.87 for a total cost of \$54,000, not including researcher time. This allowed us to  compensate photographers as well as the management and QA teams for their labour. 



\section{Comparing \appen to current datasets}
\label{sec:comp}
We compare \appen with three datasets: the canonical ImageNet~\cite{imagenet_cvpr09}, the more geographically diverse (but still webscraped) GeoYFCC~\cite{dubey2021adaptive} and finally, the recently curated DollarStreet dataset~\cite{DollarStreet}.

\begin{figure*}
    \centering
    \includegraphics[width=\textwidth]{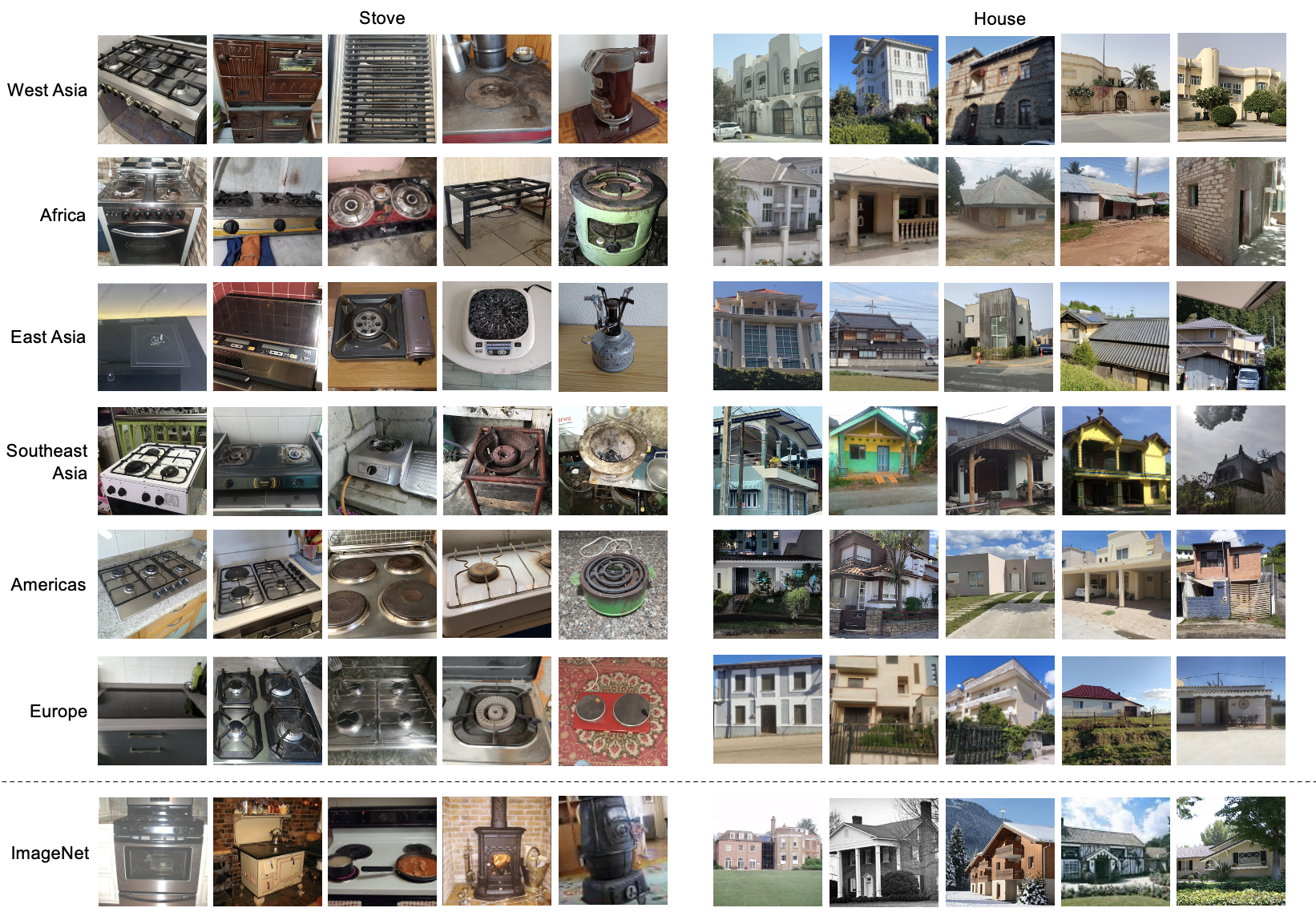}
    \caption{Sample images of two object classes in different regions within  \appen (and ImageNet in the bottom row, for comparison). Note the variety of stoves and houses across geographic regions in \appen -- and also the fact that the stoves are more \emph{used} (thus arguably more realistic) than in ImageNet. Product labels on images in the figure have been blurred.} 
    \label{fig:sample}
\end{figure*}

\smallsec{Qualitative} In \cref{fig:sample}, we show a subset of 60 \appen images of ``stoves'' and ``houses'' (more in the appendix). Compared to images from ImageNet, we see a larger variety of stoves: e.g., induction coils, single and two burner stoves. The stoves also appear more \emph{used} than those in ImageNet. Similarly, for ``house,'' we see a larger range in terms of materials used and size. In the \cref{sec:eval,sec:training} we examine the impact of this diversity on visual recognition models. 

\smallsec{Statistics} 
GeoYFCC~\cite{dubey2021adaptive} (license: CC) is a webscraped dataset subsampled from YFCC100M~\cite{yfcc100m} to be geographically diverse; thus, by raw counts it is a much larger dataset compared to both DollarStreet and GeoDE, with over 1M images from 62 countries. However, looking at the regions (Fig.~\ref{fig:pullfig}) reveals imbalances, with most countries from Europe. Moreover, this dataset does not have curated labels, just tags, and the distribution among tags is also long-tailed (the top 20 of 1,197 tags comprise 34\% of the dataset). Comparatively, \appen is balanced across both regions and classes. DollarStreet~\cite{DollarStreet} is a much smaller dataset with only 38,479 photos, comparable to GeoDE with 61,940 photos. DollarStreet has more classes but with fewer images per class compared to GeoDE: DollarStreet averages only 133 images per each of its 289 classes (with 382 images on average for its 40 most common classes), while GeoDE average 1,548 images per each of its 40 classes.

\begin{figure*}[t]
   \resizebox{0.99\linewidth}{!}{ \setlength{\tabcolsep}{2pt} \begin{tabular}{L{1.2cm}cccccc}
    & {\large West Asia} & {\large Africa} & {\large East Asia} & {\large Southeast Asia} & {\large Americas} & {\large Europe} \\
{\large Plate of food} & \raisebox{-0.6\totalheight}{\includegraphics[width=0.2\textwidth]{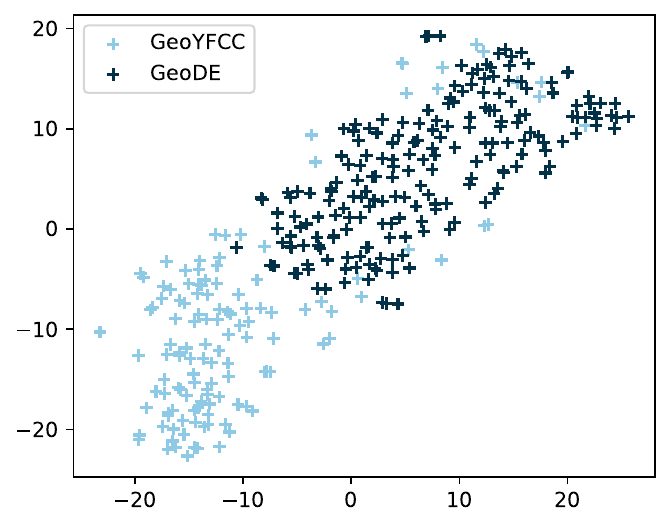}} & 
\raisebox{-0.6\totalheight}{\includegraphics[width=0.2\textwidth]{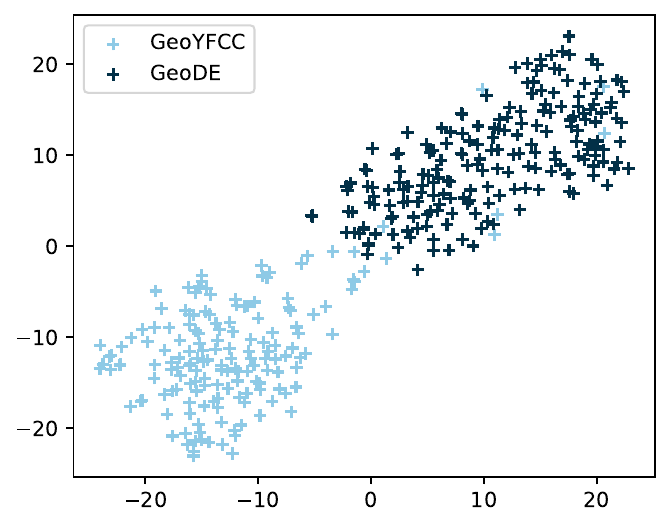}} & 
\raisebox{-0.6\totalheight}{\includegraphics[width=0.2\textwidth]{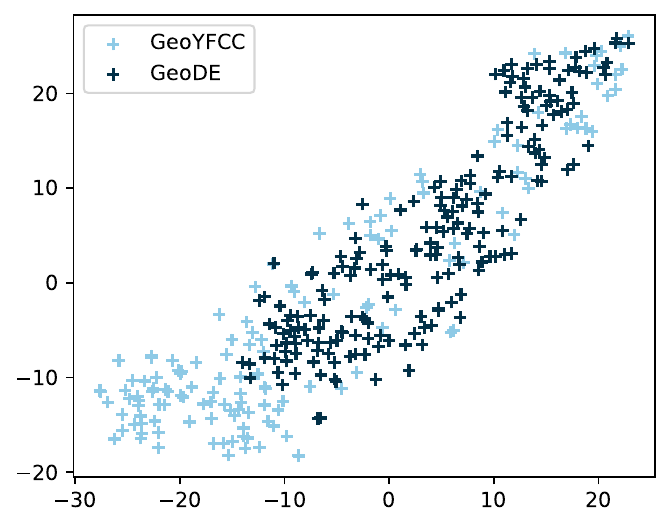}} & 
\raisebox{-0.6\totalheight}{\includegraphics[width=0.2\textwidth]{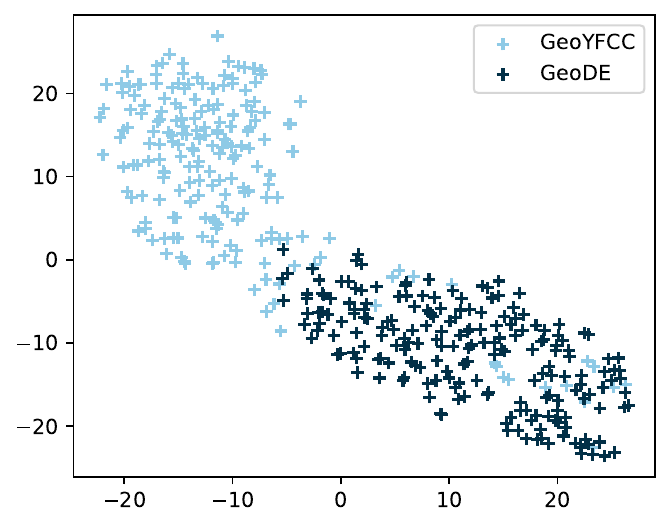}} & \raisebox{-0.6\totalheight}{\includegraphics[width=0.2\textwidth]{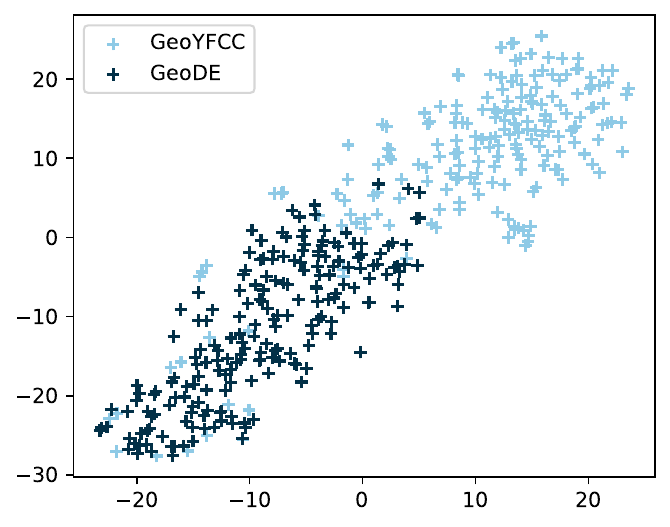}} & \raisebox{-0.6\totalheight}{\includegraphics[width=0.2\textwidth]{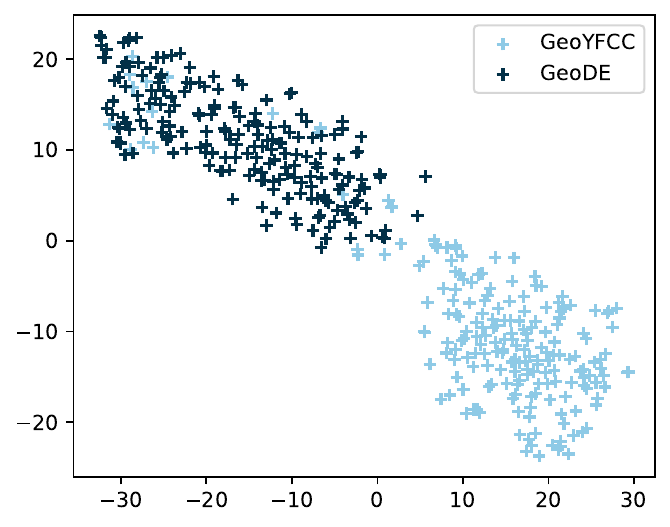}} \\
{\large Storefront} & \raisebox{-0.6\totalheight}{\includegraphics[width=0.2\textwidth]{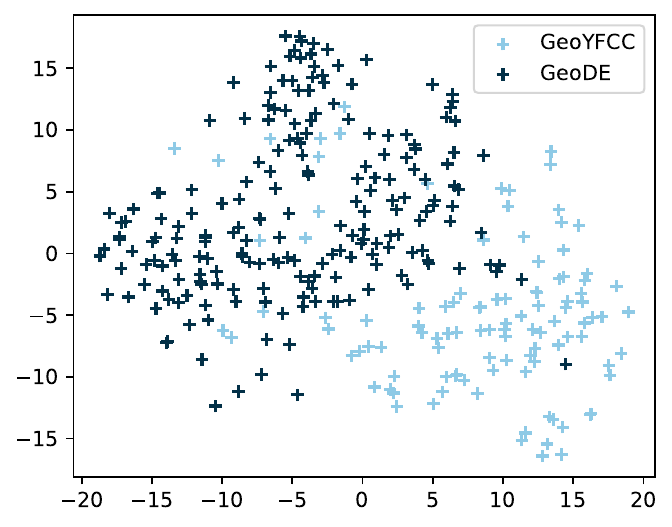}} & 
\raisebox{-0.6\totalheight}{\includegraphics[width=0.2\textwidth]{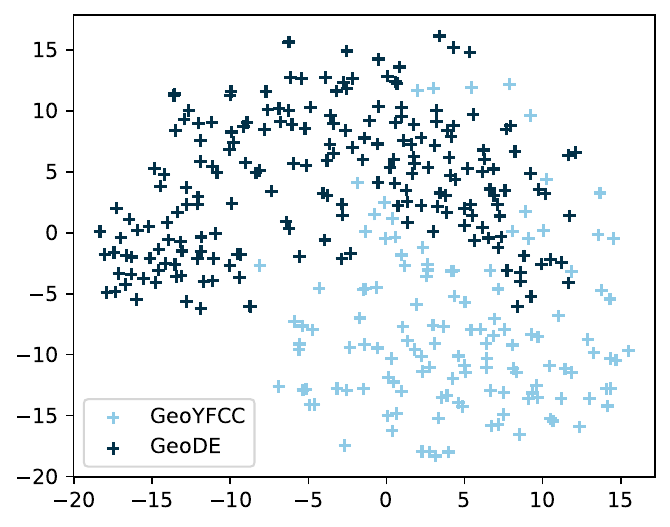}} & 
\raisebox{-0.6\totalheight}{\includegraphics[width=0.2\textwidth]{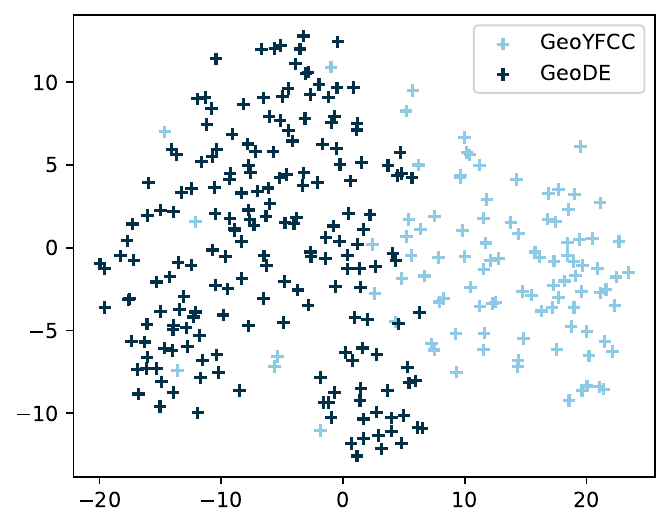}} & 
\raisebox{-0.6\totalheight}{\includegraphics[width=0.2\textwidth]{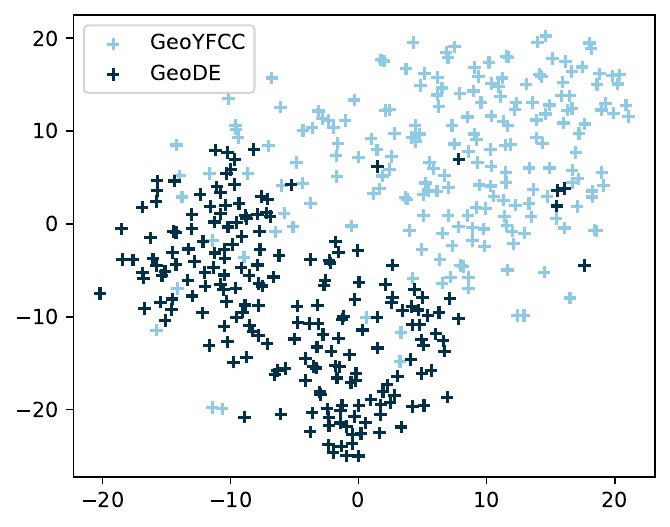}} & \raisebox{-0.6\totalheight}{\includegraphics[width=0.2\textwidth]{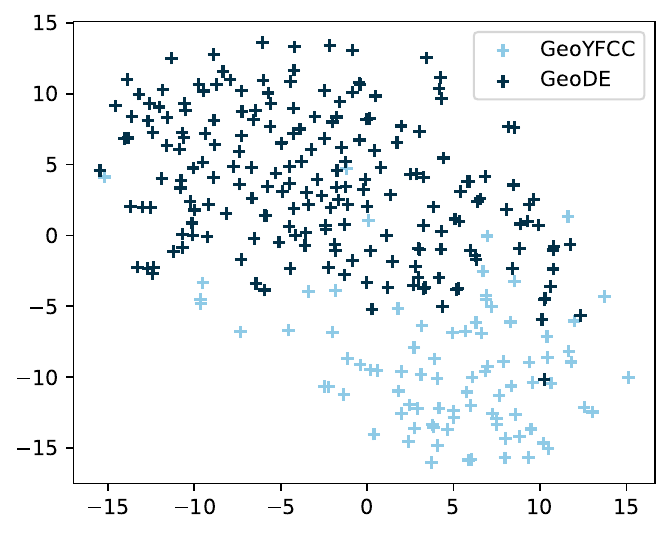}} & \raisebox{-0.6\totalheight}{\includegraphics[width=0.2\textwidth]{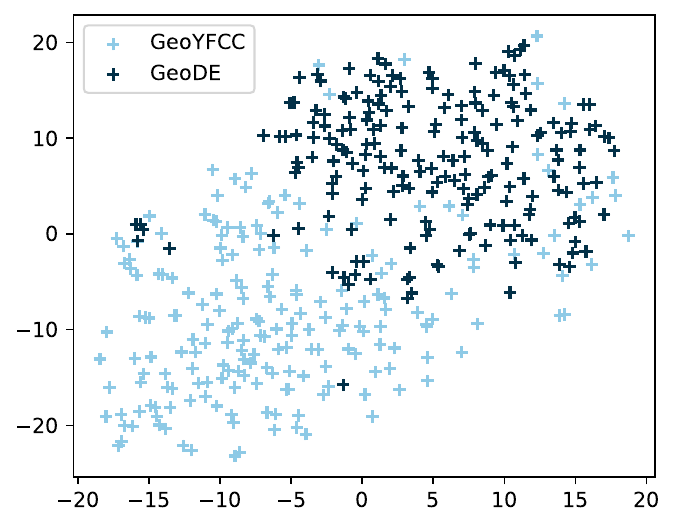}} \\

\end{tabular}}
\caption{We visualize the TSNE plots for several of object classes per region for GeoYFCC (\textcolor{SkyBlue}{light blue}) and \appen (\textcolor{MidnightBlue}{dark blue}). While the features do overlap slightly, on the whole, they are very different for dataset distributions, even within each (region, object) tuple. (See Fig.~\ref{fig:tsne_obj_reg_ds} for DollarStreet)}
\label{fig:tsne_obj_reg}
\end{figure*}

\smallsec{Object appearance} Finally, we attempt to quantify the differences in the appearance of images collected through different methods, by comparing \appen with GeoYFCC~\cite{dubey2021adaptive} and DollarStreet~\cite{DollarStreet}.
We extract features for each dataset using a ResNet50 model~\cite{he2016identity} trained with self-supervised learning SwAV~\cite{caron2020unsupervised} on the PASS dataset~\cite{asano21pass} (license: CC-BY). 
We train linear classifiers to predict the dataset given an image; the classifier achieves an accuracy of 96.3\% when trying to distinguish between GeoYFCC and \appenn, and an accuracy of 96.1\% when trying to distinguish between DollarStreet and \appenn. However, this could be the result of having different distributions of regions (in the case of GeoYFCC) and different objects (for both). To understand how the dataset distributions are different beyond just the class/region frequencies we obtain low-dimensional TSNE embeddings~\cite{vandermaaten08a} with images that restricted to a certain (region, object) pair (\cref{fig:tsne_obj_reg,fig:tsne_obj_reg_ds}). 
We see a much more pronounced difference between \appen and GeoYFCC, likely due to effects of web-scraping.

\begin{figure}[t]
   \resizebox{0.99\linewidth}{!}{ \setlength{\tabcolsep}{2pt} \begin{tabular}{L{1cm}ccc | L{1cm}ccc}
     & {\large Africa} & {\large Americas} & {\large Europe} &  & {\large Africa} & {\large Americas} & {\large Europe} \\
{\large Plate of food }& \raisebox{-0.6\totalheight}{\includegraphics[width=0.2\textwidth]{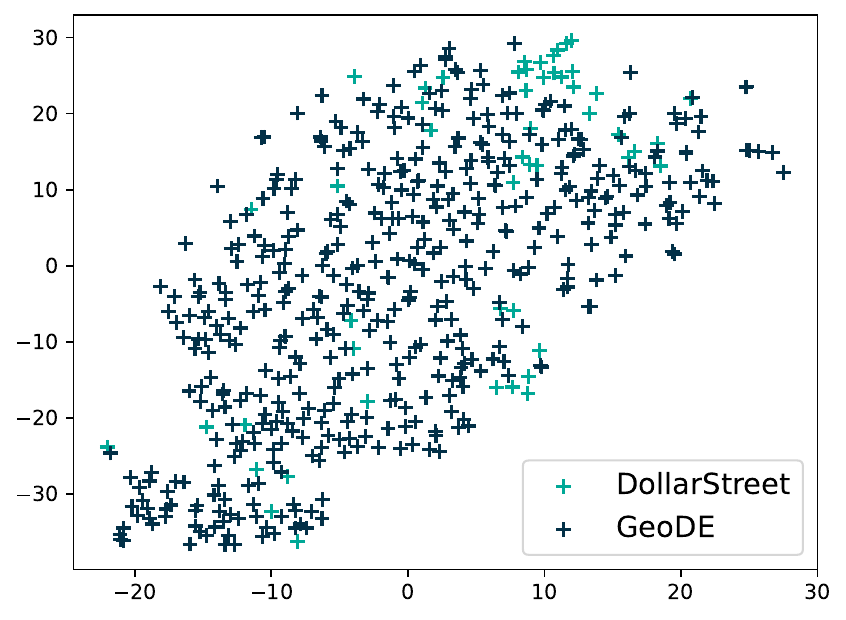}} & 
\raisebox{-0.6\totalheight}{\includegraphics[width=0.2\textwidth]{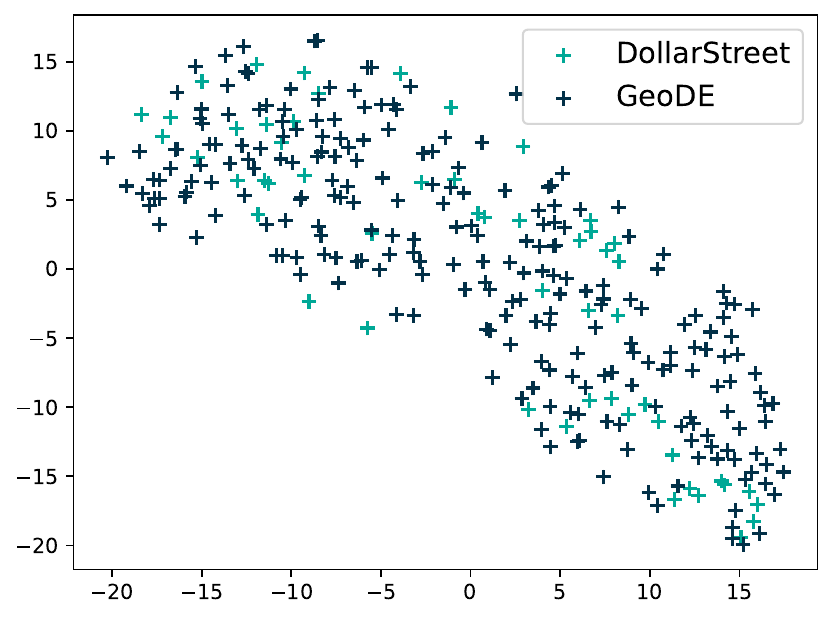}} & 
\raisebox{-0.6\totalheight}{\includegraphics[width=0.2\textwidth]{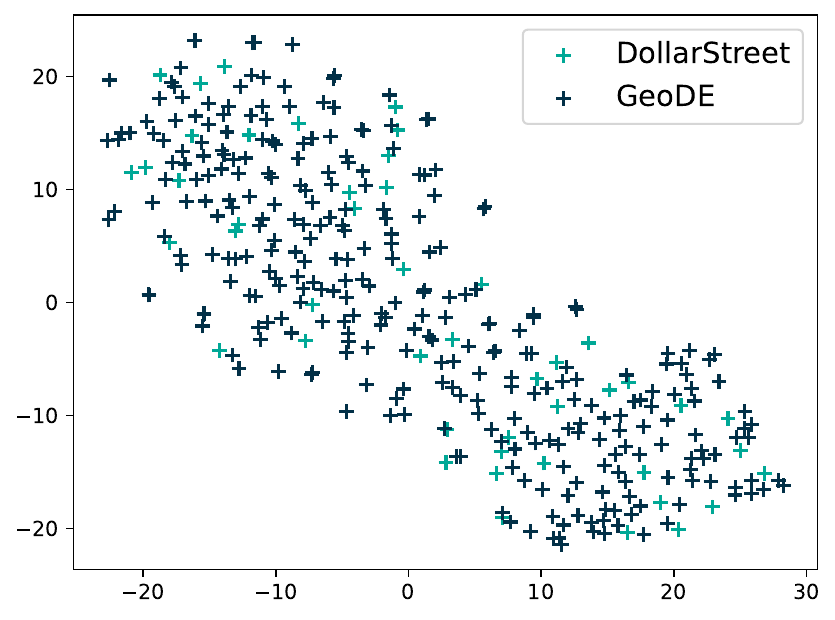}} &
{\large Stove } & \raisebox{-0.6\totalheight}{\includegraphics[width=0.2\textwidth]{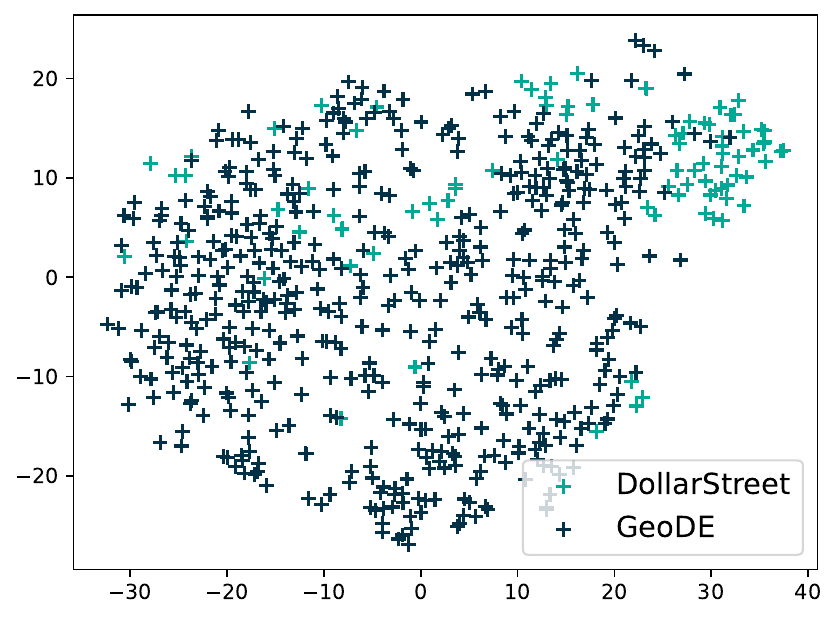}} & 
\raisebox{-0.6\totalheight}{\includegraphics[width=0.2\textwidth]{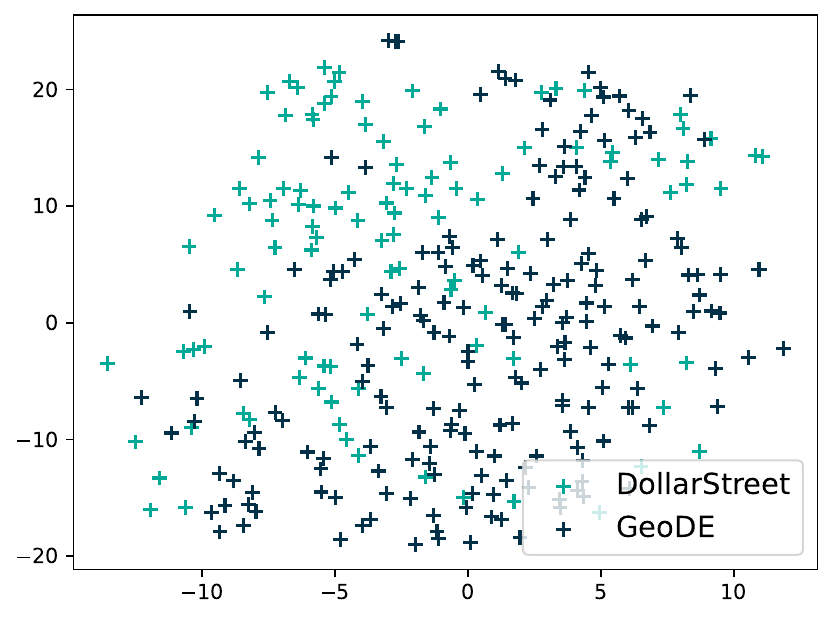}} & 
\raisebox{-0.6\totalheight}{\includegraphics[width=0.2\textwidth]{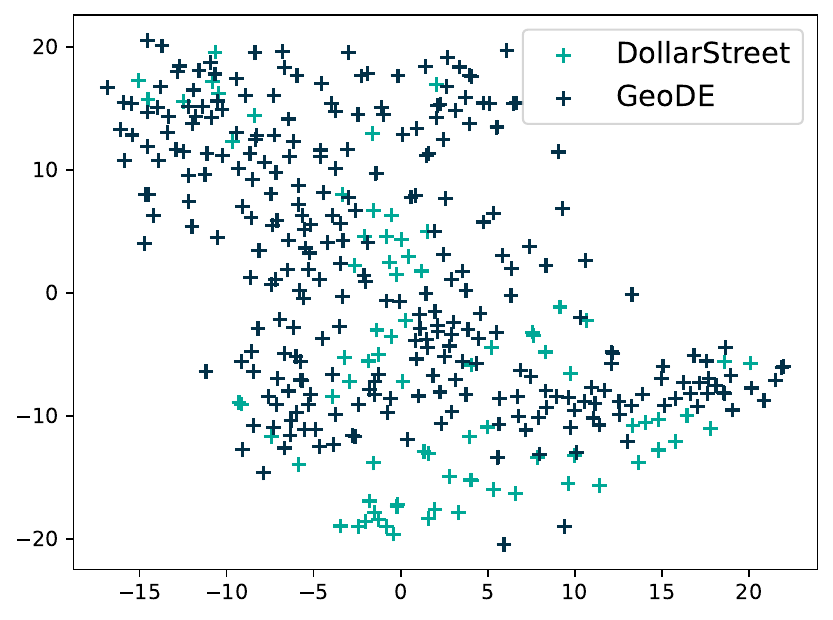}} 
\end{tabular}}
\caption{We visualize the TSNE plots for several of object classes per region for DollarStreet (\textcolor{Emerald}{cyan}) and \appen (\textcolor{MidnightBlue}{dark blue}). We see that these features overlap significantly more than that of GeoYFCC, however, there are still objects with different distributions (e.g. ``Stove'' in Americas).}
\label{fig:tsne_obj_reg_ds}
\end{figure}

\section{\appen as an evaluation dataset}
\label{sec:eval}
\begin{table*}[t]
    \centering
\caption{Per-class accuracy (in percentage; increasing order) of  CLIP~\cite{radford_clip} on  GeoDE. Objects like ``dustbin'', ``medicine'' and ``cleaning equipment'' are poorly recognized, with accuracy as low as 37\%.} 
    \label{tab:clip_accs}
\resizebox{\linewidth}{!}{
    \renewcommand{\tabcolsep}{1.5pt}
    \begin{tabular}{cccccccccccccccccccccccccccccccccccccccc}

         \toprule
    \rotatebox{90}{dustbin}& \rotatebox{90}{medicine} & \rotatebox{90}{clean. equip.} & \rotatebox{90}{spices} & \rotatebox{90}{house}
    & \rotatebox{90}{tree} & \rotatebox{90}{waste cont.} & \rotatebox{90}{candle} & \rotatebox{90}{toy} & \rotatebox{90}{backyard} & 
    \rotatebox{90}{fence} & \rotatebox{90}{streetlight} & \rotatebox{90}{stall} & \rotatebox{90}{lighter} & \rotatebox{90}{stove} &
    \rotatebox{90}{jug} & \rotatebox{90}{front door} & \rotatebox{90}{hand soap}& \rotatebox{90}{plate of food}& \rotatebox{90}{truck}&
    \rotatebox{90}{light fixture} & \rotatebox{90}{wheelbarrow}& \rotatebox{90}{storefront}& \rotatebox{90}{toothpaste}& \rotatebox{90}{toothbrush}& 
    \rotatebox{90}{flag}& \rotatebox{90}{religious bld}& \rotatebox{90}{bicycle}& \rotatebox{90}{road sign}& \rotatebox{90}{hat} &
 \rotatebox{90}{cooking pot}& \rotatebox{90}{boat}& \rotatebox{90}{hairbrush}& \rotatebox{90}{car}& \rotatebox{90}{monument}& 
 \rotatebox{90}{dog}& \rotatebox{90}{bag}& \rotatebox{90}{chair}& \rotatebox{90}{bus}& \rotatebox{90}{lightswitch} \\
   \midrule
\textcolor{red}{\underline{37}} & \textcolor{red}{\underline{54}} & \textcolor{red}{\underline{59}} & 63 & 63 & 68 & 69 & 71 & 73 & 74 & 75 & 76 & 76 & 77 & 78 & 85 & 85 & 86 & 88 & 88 & 88 & 89 & 89 & 90 & 90 & 91 & 92 & 92 & 93 & 93 & 95 & 95 & 95 & 96 & 96 & 96 & 96 & 97 & 97 & {98} \\
         \bottomrule
    \end{tabular}
\renewcommand{\tabcolsep}{6pt}
}
    
\end{table*}

\begin{figure*}[t]
    \centering
    \includegraphics[width=\linewidth]{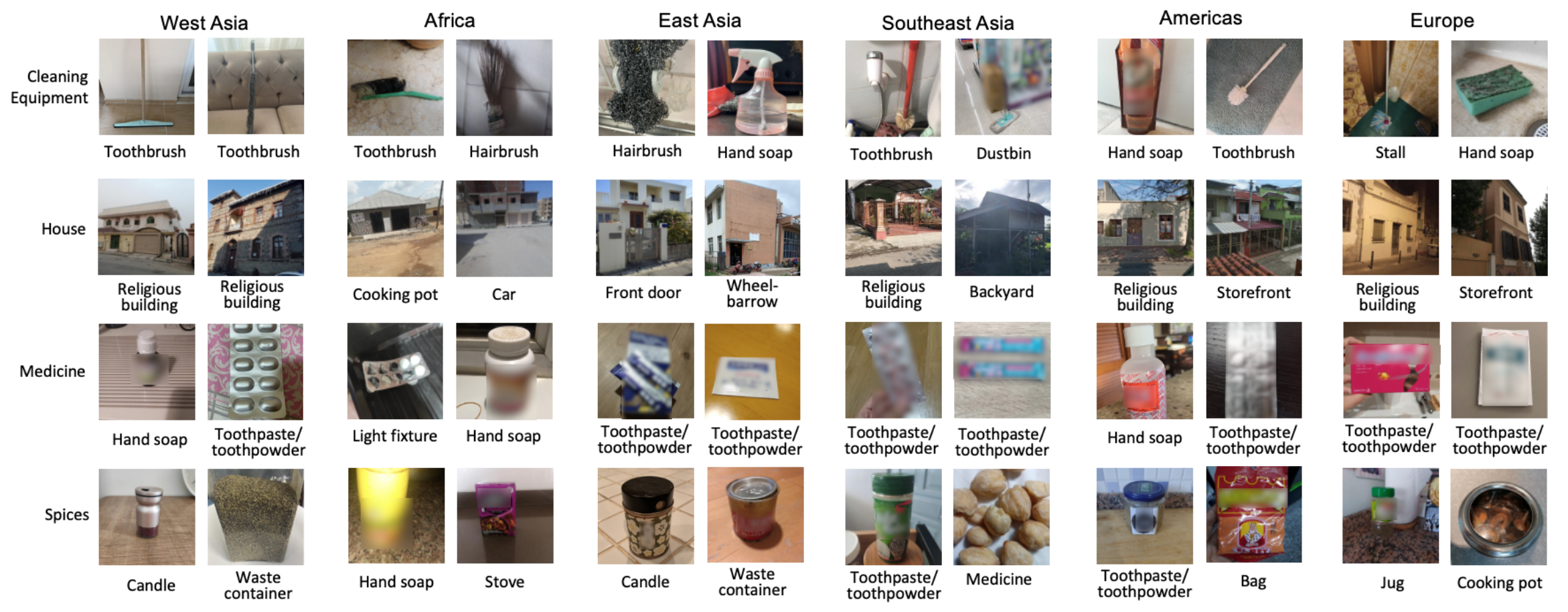}
    \caption{Example errors that the CLIP~\cite{radford_clip} model makes on \appen images (the ground truth label on the left, CLIP prediction at the bottom).  There are some systematic errors, e.g., classifying ``house'' as a ``religious building'', particularly on images from Asia. (product labels in figure are blurred).}
    \label{fig:errors}
\end{figure*}

\begin{figure*}[t]
    \centering
    \resizebox{\linewidth}{!}{
    \begin{tabular}{c c c}
    {\small spices} & {\small stove} & {\small religious building} \\
           \includegraphics[width=0.26\textwidth]{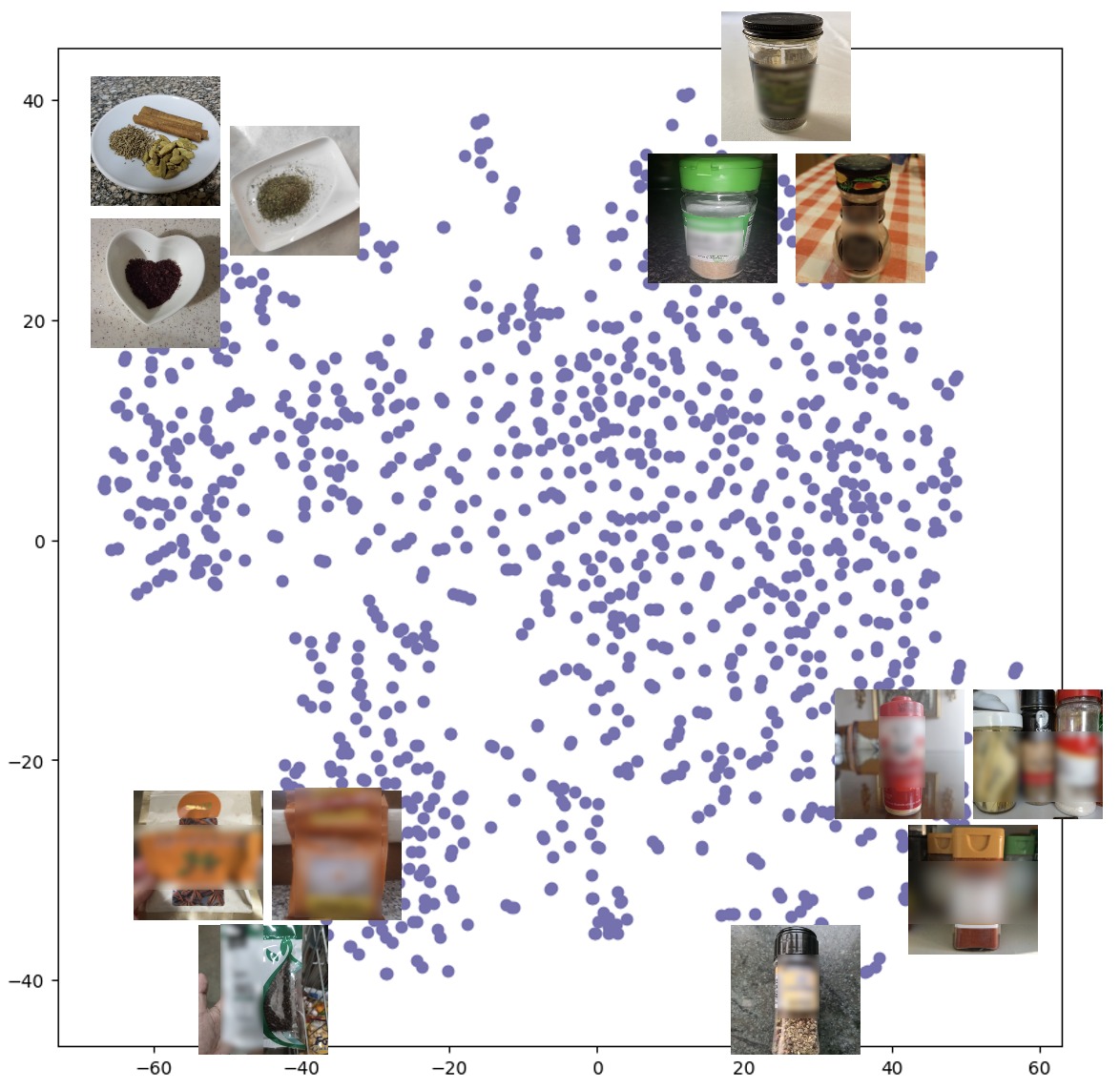} &         \includegraphics[width=0.27\textwidth]{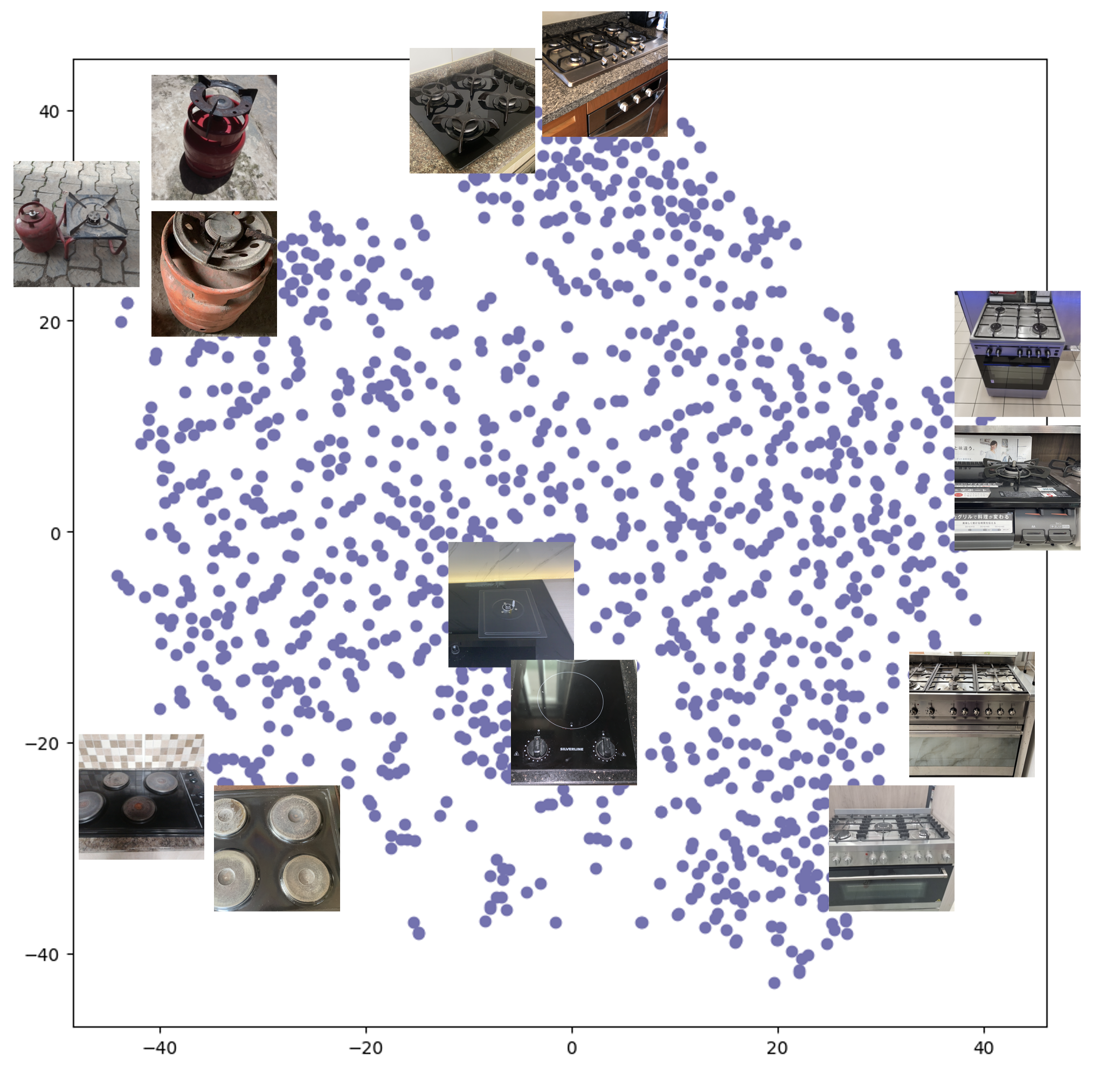} & \includegraphics[width=0.28\textwidth]{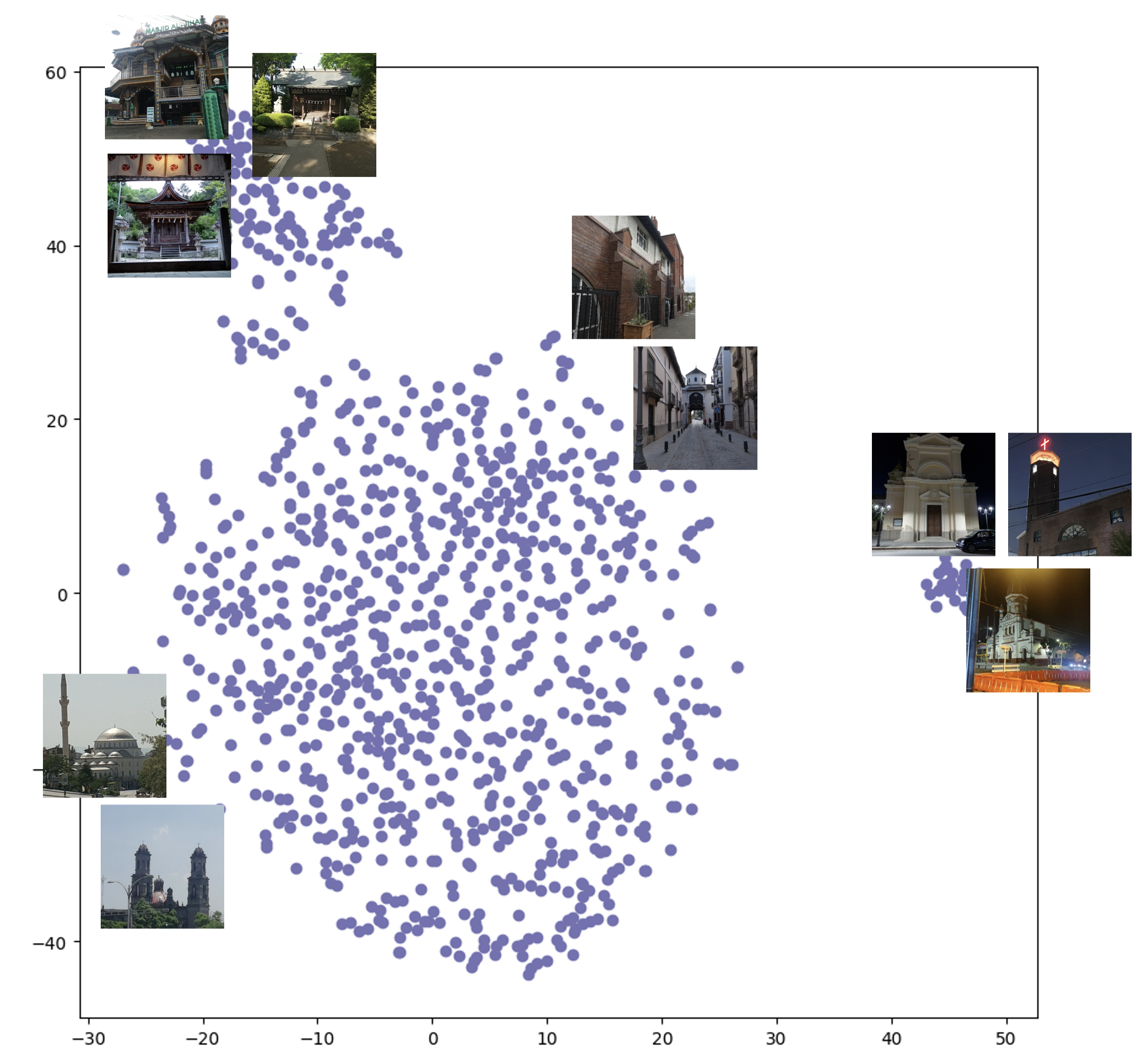}                
          \\
    \end{tabular}}
    \caption{We show the TSNE plots of objects which have large regional disparities in accuracy in the CLIP trained model, with images embedded. We see differences based on region, e.g., ``religious buildings'' contains a cluster of monasteries and temples, mostly from East and Southeast Asia.}
    \label{fig:tsne_objects}
\end{figure*}

We now analyze the use of \appen as an evaluation dataset, by using it to evaluate two canonical models: the recent CLIP~\cite{radford_clip} and an ImageNet~\cite{imagenet_cvpr09}-trained model.

\smallsec{Implementation details}
\label{subsec:training_no_appen}
For the CLIP model, we use the weights provided for the ViT-B/32 model. We use text prompts for all 40 object categories as described in the zero-shot recognition setup of~\cite{radford_clip}. To train a model on ImageNet~\cite{imagenet_cvpr09}, we first match the classes of \appen and ImageNet. We find the relevant synsets for each \appen{} class in WordNet~\cite{miller1995wordnet}, and include all images of that synset. 
For two object categories (``backyard'' and ``toothpaste/toothpowder'') we do not find any matching categories, and so we ignore these categories in the quantitative analysis. We split our filtered ImageNet~\cite{imagenet_cvpr09} dataset into train (38,353 images), validation (12,794 images), and test (12,795 images) datasets. As in Sec.~\ref{sec:comp} we extract features using a ResNet50 model~\cite{he2016identity} trained with self-supervised learning SwAV~\cite{caron2020unsupervised} on PASS~\cite{asano21pass}, and retrain the final layer.


\smallsec{Results} \cref{tab:res_clip_imgnet} (\emph{left}) shows the accuracy across different regions on these two models. Both models perform the best on images from Europe and the worst on images from Africa (difference of more than 7\% in both cases). \cref{tab:clip_accs} further breaks out the per-object accuracy for CLIP. While the average accuracy is 82.8\%, classes like ``dustbin'' (37.3\%), ``medicine'' (54.1\%), ``cleaning equipment'' (59.0\%) and ``spices'' (63.2\%) perform poorly. \cref{fig:errors} shows example errors. 


\begin{figure}[t]
    \centering
    \begin{minipage}{0.65\linewidth}
    \resizebox{\linewidth}{!}{{\small
    \setlength\tabcolsep{3 pt}
    \begin{tabular}{lcccccc}
    \toprule
    Model & WAsia & Africa & EAsia & SEAsia & Americas & Europe \\
    \midrule
    ImageNet & 69.4 & \textcolor{red}{\underline{62.7}} & 63.3 & 67.3 & 68.6 & \textbf{69.9} \\
    CLIP & 84.0 & \textcolor{red}{\underline{78.7}} & 79.9 & 81.9 & 84.4 & \textbf{85.8}\\
    \bottomrule
    \end{tabular}
    }}
    \end{minipage}%
    \begin{minipage}{0.35\linewidth}
         \includegraphics[width=\linewidth]{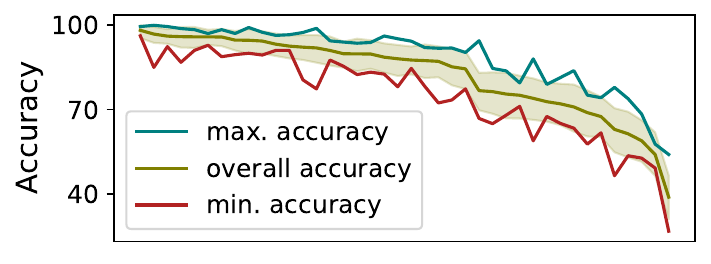}
    \end{minipage}    
    \caption{ (\textit{left}) Accuracies (in \%) on \appen of a model trained on a subset of ImageNet~\cite{imagenet_cvpr09} (details in Sec.~\ref{sec:eval}) and of CLIP~\cite{radford_clip}. The models perform \textbf{best} on images from Europe, and \textcolor{red}{\underline{worst}} on images from Africa. (\emph{right}) We compute the maximum and minimum accuracy of CLIP~\cite{radford_clip} for each object across the 6 regions in \appenn (sorted by overall accuracy). 31 of 40 objects have at least one region whose accuracy falls outside the 95\% CI, suggesting significant differences across regions. }
    \label{tab:res_clip_imgnet}
\end{figure}


In seeking to understand the accuracy variation across geographic regions, we compute the minimum and maximum per-region accuracy for each object (Fig.~\ref{tab:res_clip_imgnet} \emph{right}). We also compute the confidence interval for the \emph{expected} distribution of per-region accuracy for each object based on the object's overall accuracy.\footnote{ We draw 500 random partitions into 6 regions, and compute the resulting per-region accuracies.} 
We find that \textbf{31} of 40 objects have a minimum and/or maximum  region accuracy that falls outside the corresponding 95\% confidence interval, suggesting that these objects (including for example ``house'' and ``dustbin'') exhibit significant geographic variation (at least with respect to the visual distribution learned by CLIP).\footnote{Considering the problem of multiple hypothesis testing, we can apply the Bonferonni correction for 40 objects, with  $\alpha=0.05/40=0.001$, i.e, 99.9\% confidence intervals. Still, \textbf{21} of 40 objects fall outside their intervals, confirming that GeoDE as a whole does exhibit statistically significant geographic variation.}
We further note some differences on individual classes. ``Fence'' is  88\% accurate for images from Europe, but only 60\% and 59\% for images from Africa and SE Asia respectively. Similarly, ``stove'' is  95\% accurate in the Americas but only 67\% in East Asia. Visualizing classes using TSNE plots of the features (Fig.~\ref{fig:tsne_objects}), we see that these objects are region specific, e.g, ``religious buildings'' from East and SE Asia  uniquely include buildings like monasteries and temples; similarly, single- and two-burner ``stoves'' are primarily from Africa and SE Asia.

\section{Impact of training with \appen}
\label{sec:training}

Finally, we attempt to answer how training with \appen can improve the performance of models trained on web-scraped data. Concretely, we investigate training a model on jointly \appen and subsets of ImageNet~\cite{imagenet_cvpr09}, and demonstrate the combination improves results across geographic regions. 


\subsection{Training a model with \appen} 
\label{subsec:training_all_appen}
We would like to understand how training a model with data from \appen affects object recognition. We train a linear model using pre-trained features on a dataset comprised entirely of ImageNet images and a dataset comprised of both ImageNet and GeoDE images with the same number of images. The feature extractor is a ResNet50~\cite{He2016resnet} model trained on PASS~\cite{asano21pass} using SwAV~\cite{caron2020unsupervised}. 
\footnote{We also try a model trained from scratch and with finetuning, with similar conclusions (supp. mat.).}

\smallsec{Implementation details} We split \appen into train (4,970 images per region), validation (between 1,657 and 2,188 images per region), and test (between 1,657 and 2,189 images per region). We use the validation dataset to select training hyperparameters. The training set for our ImageNet only model is the same 38,353 image training set constructed in Sec.~\ref{subsec:training_no_appen}. To construct the training set of our ImageNet and all regions in \appen model, we add in the training sets for all 6 regions in \appen while removing the same number of images per class from ImageNet. This procedure gives a training set of 29,820 \appen images and 8,533 ImageNet~\cite{imagenet_cvpr09} images.
The models are trained using an SGD optimizer (lr $= 0.1$, momentum $=0.9$) for 500 epochs with cross entropy loss. Models were trained using a single GPU (RTX1080 or 2080) and took less than 1 hour. Results are reported on the test set.  

\smallsec{Results}
We first report results on the \appen test set, and notice a significant improvement in accuracy across all regions, as a result of training with both \appen and ImageNet (Tab.~\ref{tab:train_appen}). However, this improvement could come from the ImageNet + \appen dataset matching the domain of the \appen evaluation set and may not generalize to other datasets. Thus, we also test these models on a different dataset: the DollarStreet dataset~\cite{DollarStreet}. This dataset has been used before as an evaluation benchmark~\cite{devries2019objectrecognition}, to understand if current object recognition models can perform well on objects from a diverse set of regions. Tab.~\ref{tab:train_appen} lists the accuracy for the 4 different regions in DollarStreet, along with the per class accuracies for the object categories that overlap between \appen and DollarStreet. We see an increase in performance across most categories, suggesting that \appen is more geo-diverse than ImageNet and that there is an advantage to using geo-diverse data in the training set. 

    
  

\begin{table}[t]
    \centering
     \caption{We compare the performance of a model trained on ImageNet~\cite{imagenet_cvpr09} versus one that is trained on both ImageNet and \appen. We report results on the test set of \appen as well as the DollarStreet~\cite{DollarStreet} images. We see an improvement across all regions for both test datasets. We also report the per-class accuracies for the DollarStreet dataset and see improvement across most objects as well.} 
    \label{tab:train_appen}
\resizebox{\textwidth}{!}{
    \renewcommand{\tabcolsep}{2pt}
    \begin{tabular}{l |*{6}{c}:c| *{4}{c} :*{13}{c}:c}
   \toprule
   & \multicolumn{7}{c|}{Tested on \appen} & \multicolumn{18}{c}{Tested on DollarStreet~\cite{DollarStreet}} \\
 \midrule
  & \rotatebox{90}{West Asia} & \rotatebox{90}{Africa} & \rotatebox{90}{East Asia} & \rotatebox{90}{SE Asia} & \rotatebox{90}{Americas} & \rotatebox{90}{Europe} & \rotatebox{90}{Overall} & \rotatebox{90}{Africa} & \rotatebox{90}{America} & \rotatebox{90}{Asia} & \rotatebox{90}{Europe} & \rotatebox{90}{bicycle}& \rotatebox{90}{chair} & \rotatebox{90}{clean.equip} & \rotatebox{90}{cooking pot} & \rotatebox{90}{dustbin} & \rotatebox{90}{hand soap} & \rotatebox{90}{house} & \rotatebox{90}{light fixture} & \rotatebox{90}{light switch} & \rotatebox{90}{medicine} & \rotatebox{90}{plate of food} & 
    \rotatebox{90}{stove} & \rotatebox{90}{toy} & \rotatebox{90}{Overall}\\
   \midrule
ImageNet &  69 & 63 & 63 & 67 & 69 & 70 & 67  & 45 & 64 & 58 & 75 & 92 & 86 & 19 & 49 & \textbf{76} & 49 & 88 & 36 & 77 & \textbf{80} & 84 & \textbf{89} & 50 & 60 \\
+GeoDE & \textbf{88} & \textbf{87} & \textbf{86} & \textbf{87} & \textbf{89} & \textbf{90} & \textbf{88} & \textbf{55} & \textbf{74} & \textbf{68} & \textbf{80} & \textbf{95} & \textbf{88} & \textbf{36} & \textbf{61} & 68 & \textbf{65} & \textbf{91} & \textbf{63} & \textbf{79} & 78 & \textbf{96} & 85 & \textbf{58} & \textbf{69}\\\bottomrule
\end{tabular}
\renewcommand{\tabcolsep}{6pt}
}
   
\end{table}


\subsection{Cost-vs-Diversity tradeoffs}
The main drawback of \appen is the cost of this dataset: images collected in this way cost more than the standard pipeline of web-scraping and crowd-sourcing annotations. Thus, it is important to identify which classes and regions contribute most to the overall model. To investigate this, we start with the filtered ImageNet dataset described above, vary the amount of \appen data from a particular region, and analyze the change in overall regional performance.

\smallsec{Implementation Details} We start with the 38,353 filtered ImageNet images and add a region of \appenn's data back into the dataset and remove the same number of ImageNet images to keep the number of images and class balance the same. Other training details remain the same as in Sec.~\ref{subsec:training_all_appen}.  

\smallsec{Evaluation} As we are evaluating on the \appen test set, there are two possible sources of performance gain: (1) the model is able to take advantage of the additional regional information from the \appen data; and (2) the \appen images were collected using the same collection method as the test set and from Sec.~\ref{sec:comp}, we saw that there is a difference in the feature space that can be attributed to the collection method itself (deliberately taking photos rather than web-scraping). In order to distinguish between these two sources, we measure the accuracy on both the region in the train set \emph{and} accuracy on the images from Europe\footnote{We use Europe as this region had the best performance when using a model trained on just ImageNet.}. We also measure the increase in AP for specific objects to better understand which objects benefit most from \appen data.

\smallsec{Results} 
We find that the performance within each specific region and in Europe increase with the additional \appen data. The relative increase in performance for the specific region is larger than the increase for Europe, showing the value of data for each region, moreover, the improvements do not saturate, suggesting that more data could lead to further gains. Full results are presented in the supp. mat. We examine the classes that have the largest increase in average precision (AP) as the regional \appen images are added to the dataset in Fig.~\ref{fig:object_ap_improvement} (\emph{left}). We also present the object classes that see the most improvement in Fig.~\ref{fig:object_ap_improvement} (\emph{right}). In general, we see that specific objects such as ``spices'', ``waste container'' and ``cleaning equipment'' benefit most from regional \appen data.

\begin{figure}
    \centering
    \begin{minipage}{0.55\linewidth}
        \includegraphics[width=1\linewidth]{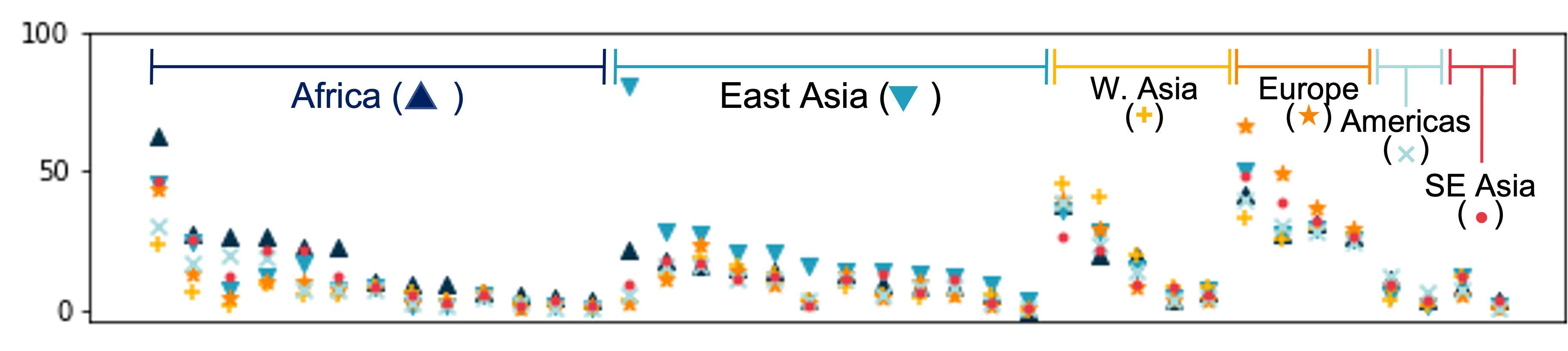}
    \end{minipage}%
    \begin{minipage}{0.45\linewidth}
    \begingroup
    \setlength{\tabcolsep}{3pt}
        \resizebox{\linewidth}{!}{
    \begin{tabular}{ll}\toprule
      & {Classes with largest \% inc. in AP} \\ \midrule
\textit{Africa} & waste cont., spices, dustbin, {clean. equip}.\\
\textit{E. Asia} & relig. blg., spices, dustbin, waste cont.\\
\textit{W. Asia} & dustbin, hand soap, clean. equip., spices, \\
\textit{Americas} & dustbin, spices, {clean. equip}., medicine\\
\textit{SE. Asia} & waste cont., spices, medicine, clean. equip.\\
\bottomrule
\end{tabular}}
\endgroup
    \end{minipage}
    \caption{(\emph{left}) We measure the relative improvement in AP per object when \appen images from that region are included in training. Each vertical line represents an object (sorted by region of max. improvement). Africa and East Asia see the largest improvement for the most classes. (\emph{right}) We highlight the classes with the largest increases in the AP when adding in training images from \appenn.}
    \label{fig:object_ap_improvement}
\end{figure}

\section{Conclusion}
\label{sec:conclusion}
We introduced a new dataset \appen which uses crowd-sourcing for image collection, a significant departure from the popular computer vision dataset collection paradigm of web-scraping for image collection. Through this collection method, we ensured that this dataset does not contain personally identifiable information, we own the rights to the images, the image creators were compensated for their work, and were able to control for geographic diversity and object distribution in the dataset.
We show that \appen is a useful dataset for highlighting shortcomings in common models (e.g., CLIP) and can improve performance when added to the training dataset.
Also, \appen shows that crowd-sourcing is a viable image collection method for creating diverse and responsible datasets. 


\smallsec{Acknowledgements} This material is based upon work partially supported by the National Science Foundation under Grant No. 2145198. Any opinions, findings, and conclusions or recommendations expressed in this material are those of the author(s) and do not necessarily reflect the views of the National Science Foundation. We also acknowledge support from Meta AI and the Princeton SEAS Howard B. Wentz, Jr. Junior Faculty Award to OR. We thank Dhruv Mahajan for his valuable insights during the project development phase. We also thank Jihoon Chung, Nicole Meister, Angelina Wang and the Princeton Visual AI Lab for their helpful comments and feedback during the writing process.

{\small
\bibliographystyle{ieee_fullname}
\bibliography{egbib}
}

\appendix

\section*{Appendix}
Here, we provide some more details about our experiments. 
\begin{itemize}
\item In Sec~\ref{supp_sec:datasheet}, we provide the datasheet for \appen, as in \cite{gebru2021datasheets}.
\item In Sec~\ref{supp_sec:obj_select}, we describe our heuristic to select object categories in more detail. 
\item In Sec~\ref{supp_sec:imagenet_comp}, we compare the \appen feature space to that of ImageNet~\cite{imagenet_cvpr09}
\item In Sec.~\ref{supp_sec:more_train}, we provide results when finetuning pre-trained models rather than just training the final layer of a ResNet. 
\item In Sec.~\ref{supp_sec:sample_images}, we give more details about \appen, including the counts of images of different regions and categories, as well as more sample images from this dataset.
\end{itemize}

\section{Datasheet for \appen}
\label{supp_sec:datasheet}
We include the datasheet for \appen below, based on Datasheets for Datasets~\cite{gebru2021datasheets}\footnote{The template used is from \url{https://github.com/AudreyBeard/Datasheets-for-Datasets-Template}}

\begin{mdframed}[style=MyFrame, linecolor=\sectioncolor]
\section*{\textcolor{\sectioncolor}{
    MOTIVATION
}}
\end{mdframed}

\textcolor{\sectioncolor}{\textbf{For what purpose was the dataset created?
    }
    Was there a specific task in mind? Was there
    a specific gap that needed to be filled? Please provide a description.
    } \\
 \appen was created for 2 purposes: (1) To construct a more geographically diverse dataset for training and evaluation and (2) to understand what it would take to crowd-source an image dataset from scratch.   \\
    
    \textcolor{\sectioncolor}{\textbf{Who created this dataset (e.g., which team, research group) and on behalf
    of which entity (e.g., company, institution, organization)?
    }
    } \\
   \appen was created via a collaboration between the Princeton Visual AI lab and Meta research. \\
    
    \textcolor{\sectioncolor}{\textbf{What support was needed to make this dataset?
    }
    (e.g.who funded the creation of the dataset? If there is an associated
    grant, provide the name of the grantor and the grant name and number, or if
    it was supported by a company or government agency, give those details.)
    } \\
Creation of \appen was partially supported by the National Science Foundation under Grant No. 2145198. It was also supported by Meta AI and the Princeton SEAS Howard B. Wentz, Jr. Junior Faculty Award to OR.
    
   \textcolor{\sectioncolor}{\textbf{
    Any other comments?
    }} \\
    N/A \\


\begin{mdframed}[linecolor=\sectioncolor]
\section*{\textcolor{\sectioncolor}{
    COMPOSITION
}}
\end{mdframed}
    \textcolor{\sectioncolor}{\textbf{What do the instances that comprise the dataset represent (e.g., documents,
    photos, people, countries)?
    }
    Are there multiple types of instances (e.g., movies, users, and ratings;
    people and interactions between them; nodes and edges)? Please provide a
    description.
    } \\
\appen consists of images of 40 different categories from 6 different regions. These categories and regions are fully listed in tables 2 and 3 of the main paper. 
Additionally, meta data for each image includes the type of phone used to take the picture, the GPS coordinates of the image, whether there are people present in the background of the image (note that there are no recognizable people in the dataset), and if a large fraction of the image consists of trees. 
    
    \textcolor{\sectioncolor}{\textbf{How many instances are there in total (of each type, if appropriate)?
    }
    } \\
There are 61, 940 images. 
    
    \textcolor{\sectioncolor}{\textbf{Does the dataset contain all possible instances or is it a sample (not
    necessarily random) of instances from a larger set?
    }
    If the dataset is a sample, then what is the larger set? Is the sample
    representative of the larger set (e.g., geographic coverage)? If so, please
    describe how this representativeness was validated/verified. If it is not
    representative of the larger set, please describe why not (e.g., to cover a
    more diverse range of instances, because instances were withheld or
    unavailable).
    } \\
\appen consists of a sample of images. The larger set would include all possible objects from all possible regions of the world.
\appen is balanced across 6 regions and 40 objects,(regions chosen to maximise geodiversity).\\

    \textcolor{\sectioncolor}{\textbf{What data does each instance consist of?
    }
    “Raw” data (e.g., unprocessed text or images) or features? In either case,
    please provide a description.
    } \\
    Each instance consists of an image, location, object in the image, whether there are people in the background, whether a significant portion of the image contains trees. \\ 
    
    \textcolor{\sectioncolor}{\textbf{Is there a label or target associated with each instance?
    }
    If so, please provide a description.
    } \\
Yes, each image is labelled with one of 40 objects. \\
    
    \textcolor{\sectioncolor}{\textbf{Is any information missing from individual instances?
    }
    If so, please provide a description, explaining why this information is
    missing (e.g., because it was unavailable). This does not include
    intentionally removed information, but might include, e.g., redacted text.
    } \\
    No \\
    
    \textcolor{\sectioncolor}{\textbf{Are relationships between individual instances made explicit (e.g., users’
    movie ratings, social network links)?
    }
    If so, please describe how these relationships are made explicit.
    } \\
    N/A\\
    
    \textcolor{\sectioncolor}{\textbf{Are there recommended data splits (e.g., training, development/validation,
    testing)?
    }
    If so, please provide a description of these splits, explaining the
    rationale behind them.
    } \\
No, we created training, validation and test splits by randomly splitting the dataset.\\
    
    
    \textcolor{\sectioncolor}{\textbf{Are there any errors, sources of noise, or redundancies in the dataset?
    }
    If so, please provide a description.
    } \\
Possible errors include issues with labels of objects and countries. To the best of our knowledge this is limited to lesser than 1\% of the dataset.  \\
    
    \textcolor{\sectioncolor}{\textbf{Is the dataset self-contained, or does it link to or otherwise rely on
    external resources (e.g., websites, tweets, other datasets)?
    }
    If it links to or relies on external resources, a) are there guarantees
    that they will exist, and remain constant, over time; b) are there official
    archival versions of the complete dataset (i.e., including the external
    resources as they existed at the time the dataset was created); c) are
    there any restrictions (e.g., licenses, fees) associated with any of the
    external resources that might apply to a future user? Please provide
    descriptions of all external resources and any restrictions associated with
    them, as well as links or other access points, as appropriate.
    } \\
\appen is self contained    
\\

    \textcolor{\sectioncolor}{\textbf{Does the dataset contain data that might be considered confidential (e.g.,
    data that is protected by legal privilege or by doctor-patient
    confidentiality, data that includes the content of individuals’ non-public
    communications)?
    }
    If so, please provide a description.
    } \\
No, \appen does not contain confidential data.\\
    
    \textcolor{\sectioncolor}{\textbf{Does the dataset contain data that, if viewed directly, might be offensive,
    insulting, threatening, or might otherwise cause anxiety?
    }
    If so, please describe why.
    } \\
No\\
    
    \textcolor{\sectioncolor}{\textbf{Does the dataset relate to people?
    }
    If not, you may skip the remaining questions in this section.
    } \\
No \\
    
    \textcolor{\sectioncolor}{\textbf{Does the dataset identify any subpopulations (e.g., by age, gender)?
    }
    If so, please describe how these subpopulations are identified and
    provide a description of their respective distributions within the dataset.
    } \\
    N/A\\
    
    \textcolor{\sectioncolor}{\textbf{Is it possible to identify individuals (i.e., one or more natural persons),
    either directly or indirectly (i.e., in combination with other data) from
    the dataset?
    }
    If so, please describe how.
    } \\
N/A\\
    
    \textcolor{\sectioncolor}{\textbf{Does the dataset contain data that might be considered sensitive in any way
    (e.g., data that reveals racial or ethnic origins, sexual orientations,
    religious beliefs, political opinions or union memberships, or locations;
    financial or health data; biometric or genetic data; forms of government
    identification, such as social security numbers; criminal history)?
    }
    If so, please provide a description.
    } \\
  N/A\\

    \textcolor{\sectioncolor}{\textbf{Any other comments?
    }} \\
    N/A \\

\begin{mdframed}[linecolor=\sectioncolor]
\section*{\textcolor{\sectioncolor}{
    COLLECTION
}}
\end{mdframed}

    \textcolor{\sectioncolor}{\textbf{How was the data associated with each instance acquired?
    }
    Was the data directly observable (e.g., raw text, movie ratings),
    reported by subjects (e.g., survey responses), or indirectly
    inferred/derived from other data (e.g., part-of-speech tags, model-based
    guesses for age or language)? If data was reported by subjects or
    indirectly inferred/derived from other data, was the data
    validated/verified? If so, please describe how.
    } \\
    Participants from across the world took photos of different objects and submitted it. They were compensated for their efforts. This data was manually checked and verified to contain the object.  \\
    
    \textcolor{\sectioncolor}{\textbf{Over what timeframe was the data collected?
    }
    Does this timeframe match the creation timeframe of the data associated
    with the instances (e.g., recent crawl of old news articles)? If not,
    please describe the timeframe in which the data associated with the
    instances was created. Finally, list when the dataset was first published.
    } \\
   Data was collected in 2022. It was first publicly released in January 2023.  \\
    
    \textcolor{\sectioncolor}{\textbf{What mechanisms or procedures were used to collect the data (e.g., hardware
    apparatus or sensor, manual human curation, software program, software
    API)?
    }
    How were these mechanisms or procedures validated?
    } \\
 We used manual human curation: participants took photos of different objects. All images were verified by Appen's quality analysis team.  \\
    
    \textcolor{\sectioncolor}{\textbf{What was the resource cost of collecting the data?
    }
    (e.g. what were the required computational resources, and the associated
    financial costs, and energy consumption - estimate the carbon footprint.
    See Strubell \textit{et al.}\cite{strubell2019energy} for approaches in this area.)
    } \\
    {Total cost for all images was \$54,000, not including researcher time. There were no models involved in collecting the dataset. } \\
    
    \textcolor{\sectioncolor}{\textbf{If the dataset is a sample from a larger set, what was the sampling
    strategy (e.g., deterministic, probabilistic with specific sampling
    probabilities)?
    }
    } \\
    It is not a sample from a larger dataset. \\
    
    \textcolor{\sectioncolor}{\textbf{Who was involved in the data collection process (e.g., students,
    crowdworkers, contractors) and how were they compensated (e.g., how much
    were crowdworkers paid)?
    }
    } \\
    Participants from across the world were tasked with taking photos of specific objects and paid for their time. We partnered with Appen (\url{www.appen.com}), who recruited and compensated the workers. Compensation varied depending on the region, but we got assurances that the pay was appropriate for the work.\\    
    
    \textcolor{\sectioncolor}{\textbf{Were any ethical review processes conducted (e.g., by an institutional
    review board)?
    }
    If so, please provide a description of these review processes, including
    the outcomes, as well as a link or other access point to any supporting
    documentation.
    } \\
    No, \\
    
    \textcolor{\sectioncolor}{\textbf{Does the dataset relate to people?
    }
    If not, you may skip the remainder of the questions in this section.
    } \\
    No \\
    
    \textcolor{\sectioncolor}{\textbf{Did you collect the data from the individuals in question directly, or
    obtain it via third parties or other sources (e.g., websites)?
    }
    } \\
    N/A \\
    
    \textcolor{\sectioncolor}{\textbf{Were the individuals in question notified about the data collection?
    }
    If so, please describe (or show with screenshots or other information) how
    notice was provided, and provide a link or other access point to, or
    otherwise reproduce, the exact language of the notification itself.
    } \\
    N/A  \\
    
    \textcolor{\sectioncolor}{\textbf{Did the individuals in question consent to the collection and use of their
    data?
    }
    If so, please describe (or show with screenshots or other information) how
    consent was requested and provided, and provide a link or other access
    point to, or otherwise reproduce, the exact language to which the
    individuals consented.
    } \\
N/A \\
    
    \textcolor{\sectioncolor}{\textbf{If consent was obtained, were the consenting individuals provided with a
    mechanism to revoke their consent in the future or for certain uses?
    }
     If so, please provide a description, as well as a link or other access
     point to the mechanism (if appropriate)
    } \\
    N/A \\
    
    \textcolor{\sectioncolor}{\textbf{Has an analysis of the potential impact of the dataset and its use on data
    subjects (e.g., a data protection impact analysis)been conducted?
    }
    If so, please provide a description of this analysis, including the
    outcomes, as well as a link or other access point to any supporting
    documentation.
    } \\
    N/A\\
    
    \textcolor{\sectioncolor}{\textbf{Any other comments?
    }} \\
N/A \\

\begin{mdframed}[linecolor=\sectioncolor]
\section*{\textcolor{\sectioncolor}{PREPROCESSING / CLEANING / LABELING
}}
\end{mdframed}

    \textcolor{\sectioncolor}{\textbf{Was any preprocessing/cleaning/labeling of the data
    done(e.g.,discretization or bucketing, tokenization, part-of-speech
    tagging, SIFT feature extraction, removal of instances, processing of
    missing values)?
    }
    If so, please provide a description. If not, you may skip the remainder of
    the questions in this section.
    } \\
 No. \\

    \textcolor{\sectioncolor}{\textbf{Was the “raw” data saved in addition to the preprocessed/cleaned/labeled
    data (e.g., to support unanticipated future uses)?
    }
    If so, please provide a link or other access point to the “raw” data.
    } \\
    N/A \\

    \textcolor{\sectioncolor}{\textbf{Is the software used to preprocess/clean/label the instances available?
    }
    If so, please provide a link or other access point.
    } \\
    N/A \\

     \textcolor{\sectioncolor}{\textbf{Any other comments?
    }} \\
    N/A \\

\begin{mdframed}[linecolor=\sectioncolor]
\section*{\textcolor{\sectioncolor}{
    USES
}}
\end{mdframed}

    \textcolor{\sectioncolor}{\textbf{Has the dataset been used for any tasks already?
    }
    If so, please provide a description.
    } \\
    \appen has been used to evaluate large scale models for geographical bias in this paper.  \\

\textcolor{\sectioncolor}{\textbf{Is there a repository that links to any or all papers or systems that use the dataset?
    }
    If so, please provide a link or other access point.
    } \\
    No. \\

    \textcolor{\sectioncolor}{\textbf{What (other) tasks could the dataset be used for?
    }
    } \\
    Additional uses of \appen could be trying to understand geodiversity of current datasets, to understand how webscraping and crowd-collected images differ (a small analysis done in section 5 of the main paper). \\

    \textcolor{\sectioncolor}{\textbf{Is there anything about the composition of the dataset or the way it was
    collected and preprocessed/cleaned/labeled that might impact future uses?
    }
    For example, is there anything that a future user might need to know to
    avoid uses that could result in unfair treatment of individuals or groups
    (e.g., stereotyping, quality of service issues) or other undesirable harms
    (e.g., financial harms, legal risks) If so, please provide a description.
    Is there anything a future user could do to mitigate these undesirable
    harms?
    } \\
    All images in \appen were collected by participants with smart phones. Thus, the dataset does not exhibit economic diversity.  \\

    \textcolor{\sectioncolor}{\textbf{Are there tasks for which the dataset should not be used?
    }
    If so, please provide a description.
    } \\
    N/A \\

    \textcolor{\sectioncolor}{\textbf{Any other comments?
    }} \\
    N/A \\

\begin{mdframed}[linecolor=\sectioncolor]
\section*{\textcolor{\sectioncolor}{
    DISTRIBUTION
}}
\end{mdframed}

    \textcolor{\sectioncolor}{\textbf{Will the dataset be distributed to third parties outside of the entity
    (e.g., company, institution, organization) on behalf of which the dataset
    was created?
    }
    If so, please provide a description.
    } \\
    Yes, \appen is freely available to download at \url{https://geodiverse-data-collection.cs.princeton.edu/} \\

    \textcolor{\sectioncolor}{\textbf{How will the dataset will be distributed (e.g., tarball on website, API,
    GitHub)?
    }
    Does the dataset have a digital object identifier (DOI)?
    } \\
    \appen is available as a .zip file to download.  \\

    \textcolor{\sectioncolor}{\textbf{When will the dataset be distributed?
    }
    } \\
    \appen is currently available. \\

    \textcolor{\sectioncolor}{\textbf{Will the dataset be distributed under a copyright or other intellectual
    property (IP) license, and/or under applicable terms of use (ToU)?
    }
    If so, please describe this license and/or ToU, and provide a link or other
    access point to, or otherwise reproduce, any relevant licensing terms or
    ToU, as well as any fees associated with these restrictions.
    } \\
    No, \appen is released under a CC-BY license. \\

    \textcolor{\sectioncolor}{\textbf{Have any third parties imposed IP-based or other restrictions on the data
    associated with the instances?
    }
    If so, please describe these restrictions, and provide a link or other
    access point to, or otherwise reproduce, any relevant licensing terms, as
    well as any fees associated with these restrictions.
    } \\
    No \\

    \textcolor{\sectioncolor}{\textbf{Do any export controls or other regulatory restrictions apply to the
    dataset or to individual instances?}
    If so, please describe these restrictions, and provide a link or other
    access point to, or otherwise reproduce, any supporting documentation.
    } \\
  No \\

    \textcolor{\sectioncolor}{\textbf{Any other comments?
    }} \\
    N/A \\

\begin{mdframed}[linecolor=\sectioncolor]
\section*{\textcolor{\sectioncolor}{
    MAINTENANCE
}}
\end{mdframed}

    \textcolor{\sectioncolor}{\textbf{Who is supporting/hosting/maintaining the dataset?
    }
    } \\
    Currently, \appen is being hosted by the Princeton computer science department, specifically, Dr. Vikram Ramaswamy and Prof. Olga Russakovaky. For the long term, we are considering one of two options: partnering with Common Visual Data Foundation (CVDF; \url{http://www.cvdfoundation.org/}) or utilizing \url{https://researchdata.princeton.edu/news/2023-05-25/coming-soon-princeton-data-commons} (we've seen internal versions which look great for our use cases but are waiting for it to become public)\\

    \textcolor{\sectioncolor}{\textbf{How can the owner/curator/manager of the dataset be contacted (e.g., email
    address)?
    }
    } \\
    Questions can be emailed to \href{mailto:vr23cs.princeton.edu}{vr23@cs.princeton.edu} \\

    \textcolor{\sectioncolor}{\textbf{Is there an erratum?
    }
    If so, please provide a link or other access point.
    } \\
    No. \\

    \textcolor{\sectioncolor}{\textbf{Will the dataset be updated (e.g., to correct labeling errors, add new
    instances, delete instances)?
    }
    If so, please describe how often, by whom, and how updates will be
    communicated to users (e.g., mailing list, GitHub)?
    } \\
    Yes, the dataset will be updated as needed, by Vikram Ramaswamy. Updates will be posted on the GitHub repo along with the website, on how to access the corrected version.  \\

    \textcolor{\sectioncolor}{\textbf{If the dataset relates to people, are there applicable limits on the
    retention of the data associated with the instances (e.g., were individuals
    in question told that their data would be retained for a fixed period of
    time and then deleted)?
    }
    If so, please describe these limits and explain how they will be enforced.
    } \\
    N/A \\

    \textcolor{\sectioncolor}{\textbf{Will older versions of the dataset continue to be
    supported/hosted/maintained?
    }
    If so, please describe how. If not, please describe how its obsolescence
    will be communicated to users.
    } \\
    No, older versions will not continue to be hosted, however, we will provide information on our github as well as the webpage, with a script to update the dataset (if applicable). \\

    \textcolor{\sectioncolor}{\textbf{If others want to extend/augment/build on/contribute to the dataset, is
    there a mechanism for them to do so?
    }
    If so, please provide a description. Will these contributions be
    validated/verified? If so, please describe how. If not, why not? Is there a
    process for communicating/distributing these contributions to other users?
    If so, please provide a description.
    } \\
    No, there is no current mechanism to do so. Users can provide feedback / corrections to the dataset on github, which we will use to update the dataset.  \\

    \textcolor{\sectioncolor}{\textbf{
    Any other comments?
    }} \\
    N/A \\

\section{Selecting object categories for \appen}
\label{supp_sec:obj_select}

In this section, we provide more details about the object selection heuristic we employed. We mainly used the GeoYFCC~\cite{dubey2021adaptive} dataset that was constructed to be geodiverse. 

\smallsec{Implementation details} Features for GeoYFCC were extracted using a ResNet50~\cite{He2016resnet} pretrained on ImageNet~\cite{imagenet_cvpr09}. We used Logistic regression, Linear SVM and KMeans clustering implementations from the sklearn library~\cite{scikit-learn}. We used continents as regions. GeoYFCC~\cite{dubey2021adaptive} contains over 1200 tags, we ignored all tags with counts in the bottom 20th percentile, giving us a total of 745 tags. 

First, we apply each of these methods to GeoYFCC to identify candidate tags.
\begin{itemize}[itemsep=-2pt, leftmargin=*]
    \item For each region $R$, we train a linear model using a feature extractor and images from all regions except $R$ and a linear model trained on all images from all regions, to predict the presence or absence of each tag. We then applied both models to images from $R$. The difference in performance between these models allows us to measure the difference in appearance of the tag. We selected tags where in the weighted average precision on the region was less than 0.8* the performance on other regions. This gave us a set of 277 tags such as ``footstool'', ``chili'', ``case'', etc. 
   \item For each tag $T$, we train a linear SVM to predict the region given the features of images containing tag $T$. If this model has high accuracy, this suggests that this tag is visually different across regions. We selected tags that had an accuracy of over 50\%. 223 tags were identified in this manner.  ``Cork'', ``bowler\_hat'' and ``mountain\_bike'' are examples of tags found in this way. 
  \item We clustered features of images containing tag $T$. We then computed the Gini impurity of each world region, and selected tags that had a median Gini value of at least 0.5. This gave us 75 tags in total. Examples of tags found in this way were ``chili'', ``footstool'' and ``stove''.
\end{itemize}

\begin{table*}[t]
    \centering
    {\small
    \resizebox{\linewidth}{!}{
    \begin{tabular}{L{\linewidth}}
         \toprule 
         \textbf{Leave one out training} \\\midrule
         curler, fan, footstool, chili, coconut, toilet, canoe, motorboat, mountain-bike, stupa, villa, backpack, baseball-glove, basin, basket, 
         bat, bathtub, battery, beer-mug, belt, blade, bowl, bowl, broom, bucket, carryall, case, cash-machine, cassette, cleaver, cologne, cooler, counter, dinner-dress, dinner-jacket, dish, gown, grille, hammer, jacket, kettle, 
         microphone, parka, porch, rack, remote-control, sandal, scale, shelf, shot-glass, stereo, stocking, stool, sweater, tape, teddy, 
         timer, tripod, trouser, turntable, wardrobe, weight, wok, \textcolor{red}{woodcarving}, hot-pot, chewing-gum, cucumber, lime, fig, pineapple, jackfruit, 
         kiwi, mango, basil, garlic, sage, lager, ale, porter, stout, champagne, rum, tequila, vodka, whiskey, mocha \\\toprule
         \textbf{Linear SVM for region} \\
         \midrule
        mountain-bike, bicycle, raft, ferry, ship, kayak, streetcar, bus, impala, car, footstool, bench, chair, mushroom, breakfast, vegetable, dessert, dinner, door, bowler-hat, house, building, chandelier, lamp, light, castle, acropolis, fortress, tower, palace, dome, architecture, memorial, statue, sculpture, gravestone, arch, temple, stupa, monastery, church, cathedral, chapel, mosque, signboard, grocery-store, shop, kitchen, lantern, doll, \textcolor{NavyBlue}{coati}, cork, \textcolor{NavyBlue}{primate}, alp, shore, curler, cologne, seashore, \textcolor{NavyBlue}{gnu}, \textcolor{NavyBlue}{hog}, \textcolor{NavyBlue}{giraffe},  \textcolor{red}{arctic}, ice-rink, ski, \textcolor{NavyBlue}{elephant}, guinness, makeup, circuit, geyser, skyscraper, \textcolor{NavyBlue}{hippopotamus}, basketball, paintball, sword, hijab, fortification, craft, clock, stage, tractor, dagger, \textcolor{red}{defile}, bikini, swing, windmill, motor, brick, snowboard, course, volleyball, display, opera, railing, playground, veranda, wind-instrument, city-hall, ruin, portfolio, newspaper, airbus, bridge, airfield, \textcolor{red}{global-positioning-system}, brake-drum, \textcolor{NavyBlue}{kid}, mangrove, motor-scooter, crane, intersection, plain, column, wardrobe, interface, guitar, costume, grand-piano, aircraft, factory, seaside, ball, sweet, gravy-boat, spotlight, \textcolor{NavyBlue}{american-bison}, sail, beer, pier, road, tulip, grass, miniskirt, willow, flood, street, roof, slide, cliff, track, train, vehicle, boot, world, patio, window, rainbow, beacon, sidewalk, organ-pipe, tank, cable-car, grey, hall, map, cattle, airport, school, mountain, promontory, \textcolor{NavyBlue}{monkey}, motorcycle, bubble, black, mirror, golf-club, skateboard, computer, university, denim, sky, rock, earphone, \textcolor{red}{descent}, garden, hill, library, tea, blush-wine, radio, bill, sunglass, ballpark, apparel, web, field-glass, reef, fountain, \textcolor{red}{downhill}, pen, cable, step, graffito, conveyance, fabric, hovel, umbrella, iron, cloud, strand, toilet, walker, valley, airplane, cup, base, wire, \textcolor{NavyBlue}{camel}, pizza, bathroom, lounge, dock, van, circuit-board, bell, \textcolor{NavyBlue}{sheep}, book, fish, canyon, fire, \textcolor{red}{array}, rangefinder, coca-cola \\\toprule
        \textbf{Clustering} \\\midrule
        acropolis, cork, coati, footstool, stupa, impala, chili, \textcolor{NavyBlue}{primate}, cologne, \textcolor{NavyBlue}{gnu}, guinness, alp, \textcolor{NavyBlue}{hog}, shore, boater, walker, plain, \textcolor{NavyBlue}{hippopotamus}, raft, chandelier, curler, \textcolor{NavyBlue}{giraffe}, \textcolor{red}{arctic}, bowler-hat, castle, geyser, boot, streetcar, rum, hijab, ski, temple, windmill, dagger, \textcolor{red}{fortification}, snowboard, coffee, ice-rink, display, cathedral, bench, bikini, lantern, slope, \textcolor{NavyBlue}{elephant}, strand, sword, paintball, gravestone, tulip, golf-club, \textcolor{red}{downhill}, swing, volleyball, mushroom, monastery, \textcolor{NavyBlue}{american-bison}, stage, cup, church, wardrobe, wind-instrument, skyscraper, sweet, \textcolor{red}{course}, tower, opera, sketch, circuit, chapel, \textcolor{red}{col}, motor, clock, railing, mangrove \\\bottomrule
    \end{tabular}}}
    \caption{Prospective tags identified from GeoYFCC~\cite{dubey2021adaptive}. Tags in \textcolor{red}{red} seemed hard to picture. Tags in \textcolor{NavyBlue}{blue} are of animals that might be hard to crowdsource}. 
    \label{tab:my_label}
\end{table*}

After identifying these tags, we first pruned them by removing tags that did not appear to correspond to an object. Examples of this include ``arctic'', ``descent'', etc. Second, we removed tags corresponding to wild animals, since these would not be found in all regions. Examples of tags removed in this manner were ``gnu'', ``camel'', etc. This gave us a list of 265 tags. Third, these tags were sometimes variants of objects, for example, we had tags like ``stupa'', ``temple'', ``church'', ``mosque'' and ``chapel''. Thus we grouped tags based on meaning. Other examples included ``breakfast'', ``dinner'' and ``dessert'' as ``plate of food'', ``stool'', ``footstool'' and ``bench'' as ``chair'', etc. This gave us a list of objects we could collect, e.g. ``religious buildings'', ``plate of food'', ``toy''.

We also provide the user demographics of the participants (\cref{fig:demographics}). As shown, images in \appen were provided by people of varying genders, ethnicities and ages.  
\begin{figure}
    \centering
    \includegraphics[width=0.75\linewidth]{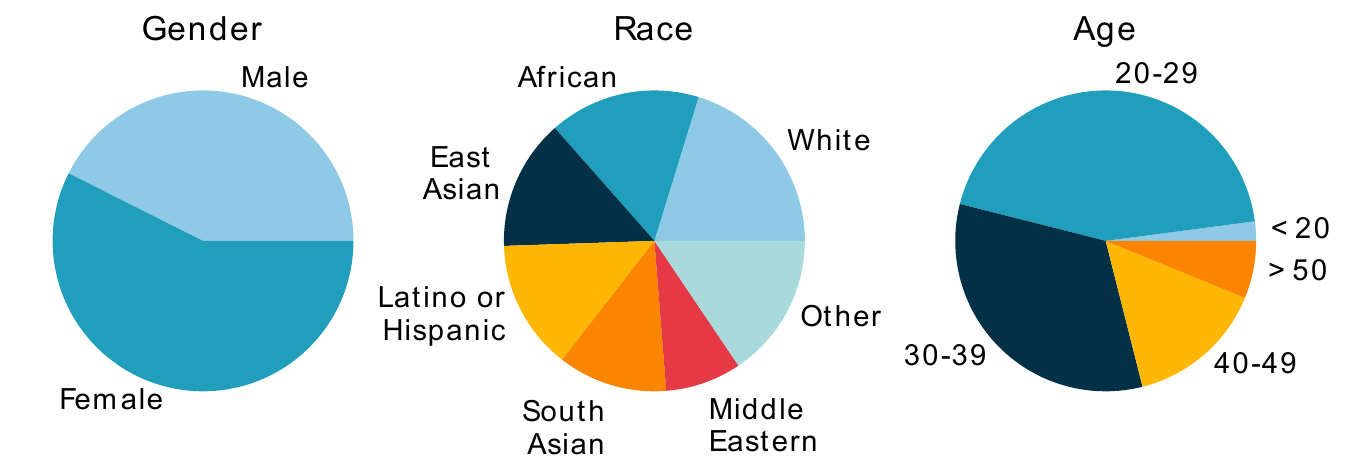}
    \caption{Participant demographics}
    \label{fig:demographics}
\end{figure}

\vspace*{0.75cm}

\section{Comparison with ImageNet}
\label{supp_sec:imagenet_comp}
In this section, we run more comparisons of GeoDE with ImageNet~\cite{imagenet_cvpr09} and DollarStreet~\cite{DollarStreet}.

\subsection{Comparison to ImageNet}
We note that the comparison to GeoYFCC~\cite{dubey2021adaptive} in Sec. 5 in the main text required us to use tags which are noisy. Thus, to compare \appen to another web-scraped dataset, we compare \appen to ImageNet in this section, checking how much the feature spaces differ. 

We find a subset of ImageNet21k as outlined in Sec. 6, and extract features using a PASS~\cite{asano21pass} trained ResNet50~\cite{He2016resnet} model. Other implementation details remain the same as in Sec 5 in the main paper. 

We first use a Logistic regression model to predict the dataset that the features are taken from and this has an accuracy of 96.0\%, showing that the feature space is very different. We also visualize TSNE plots of different objects in figure~\ref{fig:tsne_imagenet}.

\begin{figure}[t]
    \centering
    \begin{tabular}{cc}
     \includegraphics[width=0.4\linewidth]{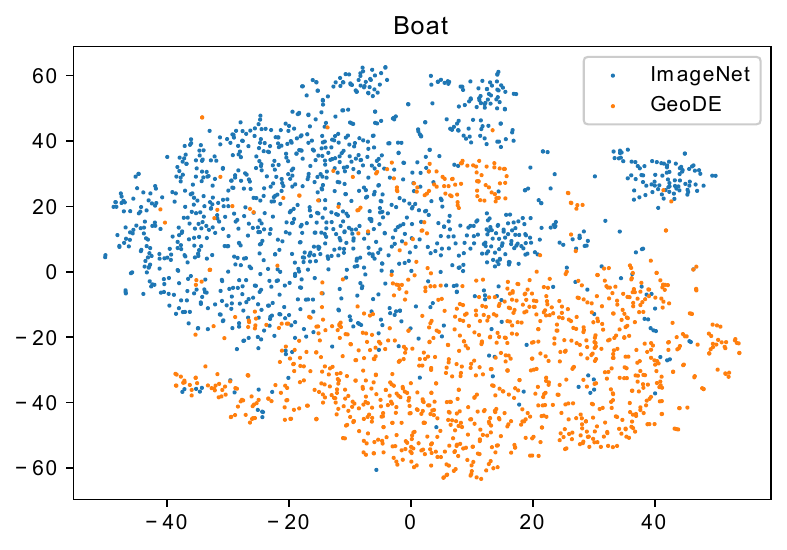}   &      \includegraphics[width=0.4\linewidth]{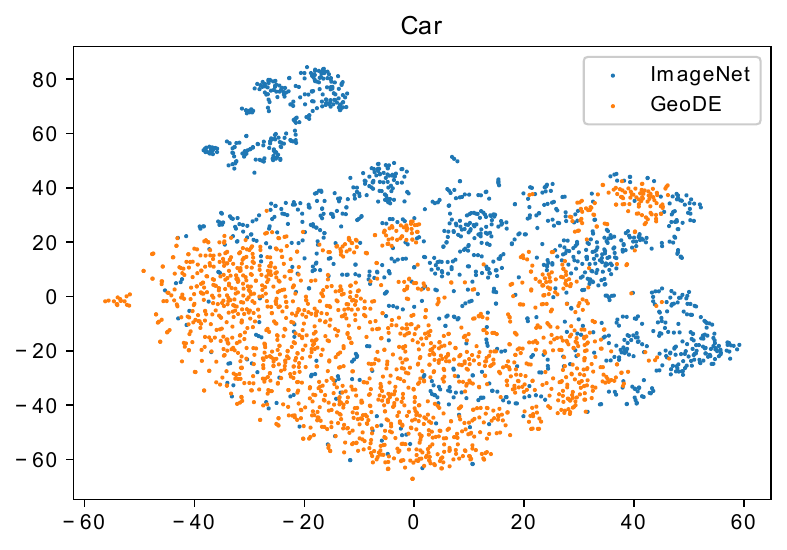} \\    
     \includegraphics[width=0.4\linewidth]{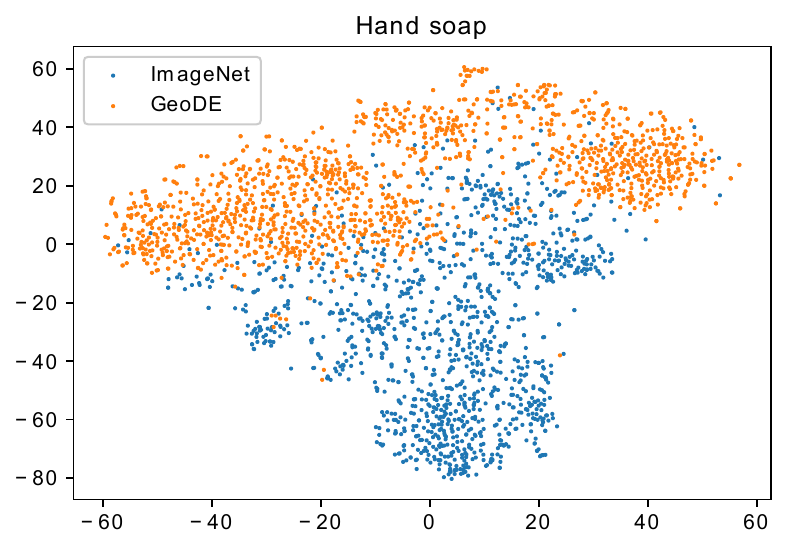}   &      \includegraphics[width=0.4\linewidth]{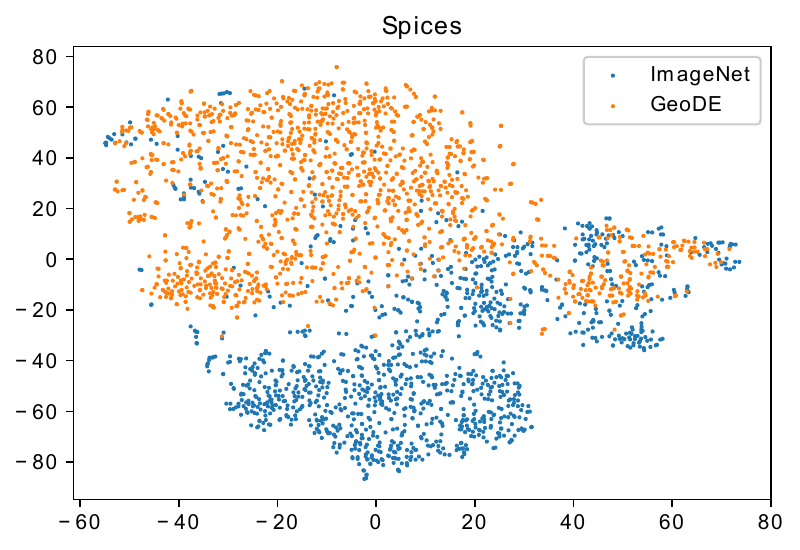} \\    
    \end{tabular}

    \caption{We visualize the TSNE plots for several of object classes using ImageNet and \appen. While the features do overlap slightly, on the whole, they are very different for dataset distributions, even within each category.}
    \label{fig:tsne_imagenet}
\end{figure}

\new{
\subsection{Comparison with DollarStreet}

Compared to GeoDE, DollarStreet~\cite{DollarStreet} contains a lot more categories, thus resulting in much fewer images per category (on average, the top 40 object categories in DollarStreet contain only 382 images, whereas GeoDE has an average of 1548 images per category). Thus, when filtering images to comprise of only common categories between both datasets, we end up with much fewer images for DollarStreet. We run small scale tests to understand how these images differ from those in GeoDE. 

\smallsec{Relative value of an image} Similar to the canonical work on dataset bias~\cite{torralba2011unbiased}, we measure the relative value of an image from \appen and DollarStreet. 
That is, we measure the number of training images needed for strong cross-dataset generalization. Concretely, we select 13 classes from DollarStreet which (1) appear in \appenn, and (2) have more than $100$ images. We restrict both datasets to these 13 classes, resulting in $4,788$ images for DollarStreet and $17,245$ images for \appenn. We now extract features using a PASS trained network, and train linear models to predict the 13 classes. First, we train a baseline model on 250 randomly sampled DollarStreet images and evaluate it on the remaining DollarStreet images; we then train models with increasing numbers of images from \appen until we match the accuracy of the DollarStreet-trained baseline. This occurs with 3,000 \appen training images. Next, we do a similar process for a baseline trained on 250 randomly sampled \appen images. However, we are unable to match its accuracy on \appen using DollarStreet training images, even after using all $4,788$ images of these classes, showing a higher relative value per image in \appen.  
}

\section{More results using \appen as an evaluation dataset}
\label{supp_sec:more_eval}
Here, we provide more more analysis performed on using GeoDE as an evaluation dataset. As shown in sec. 5 of the main text, CLIP models perform worse on certain objects (e.g ``house'', ``spices'', ``medicine'', etc.). Visualizing the probabilities assigned to different images from the same class as a box plot, we see different scenarios emerge: the variance of scores is large for all regions, as in the case of spices; the variance is large for certain regions, as in the case of stoves from Africa and East and Southeast Asia, or the scores are much lower for a specific region, as in the case of religious buildings in East Asia (\ref{fig:clip_variance}). 

\begin{figure*}[t]
    \centering
    \includegraphics[width=0.75\linewidth]{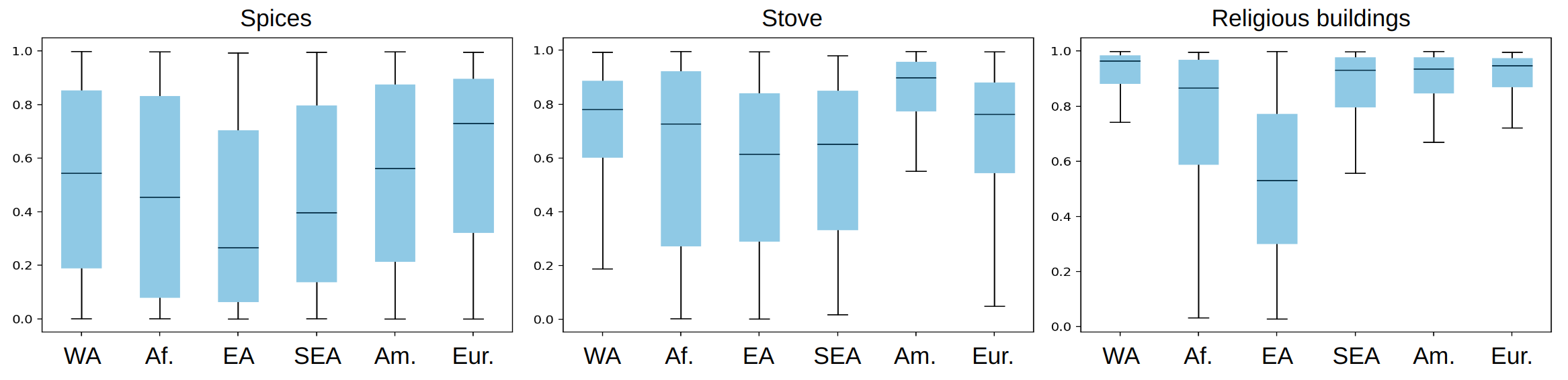}
    \caption{We visualize the probabilities assigned by CLIP to different images from the same class as a box plot. We see that the images across different regions have scores that vary in different ways: for ``spices'', we see a large variance for all regions; for ``stoves'', the variance is large for stoves from Africa, and Southeast and East Asia, but much smaller for other regions. For ``religious buildings'', we see that the scores are just much lower for buildings in East Asia.   }
    \label{fig:clip_variance}
\end{figure*}

\section{More results when training with \appen}
\label{supp_sec:more_train}
In this section, we provide results for the incremental training with \appen for different regions that were not presented in Sec 7, and provide results when fine-tuning a ResNet50 model, rather than freezing the layer weights. 

\subsection{Results from incrementally adding additional regions}
We visualize the improvement in the accuracy as we incrementally add in images from different regions (Sec. 6.2 in the main text). We can see that the performance both within the specific region and in Europe (compared to Americas when considering Europe) increase with the additional \appen data. We see that the increase within the region is larger than that of the control, showing that these images are from different domains. (Fig.~\ref{fig:incr_rest})

\begin{figure}[t]
    \centering
    \begin{tabular}{ccc}
        \includegraphics[width=0.3\linewidth]{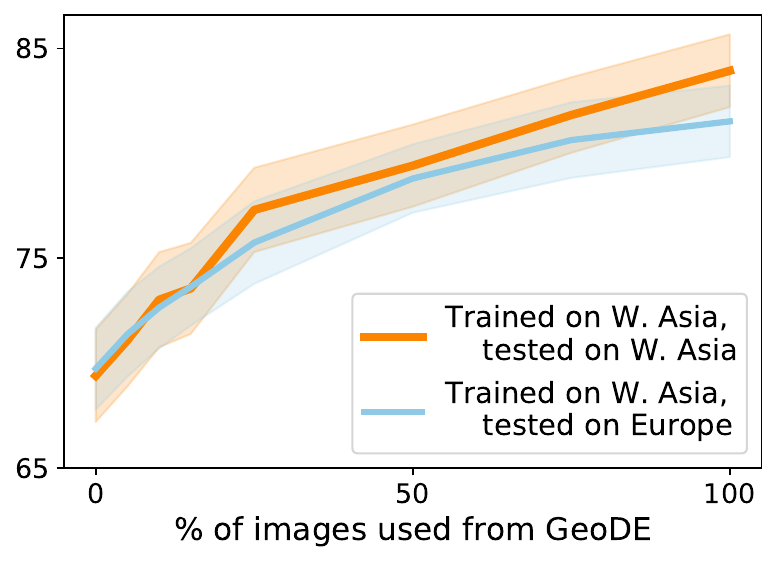} &         \includegraphics[width=0.3\linewidth]{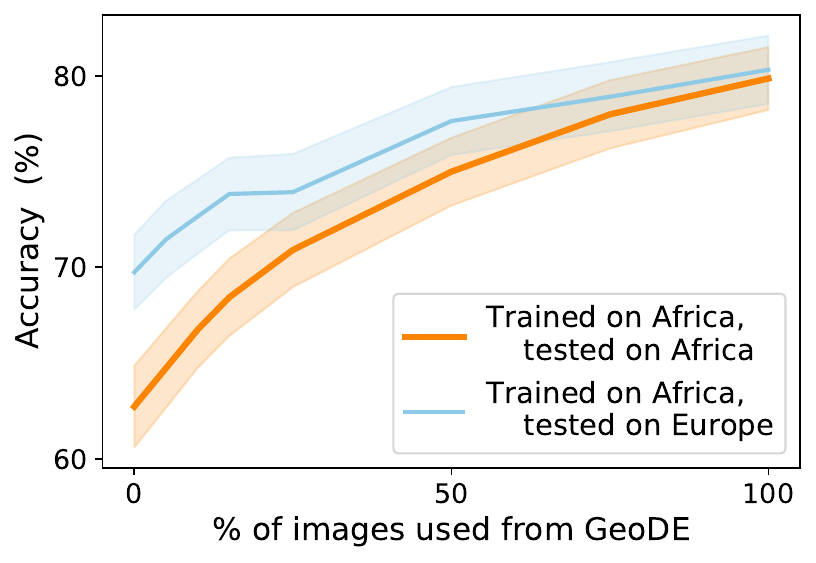}  &
        \includegraphics[width=0.3\linewidth]{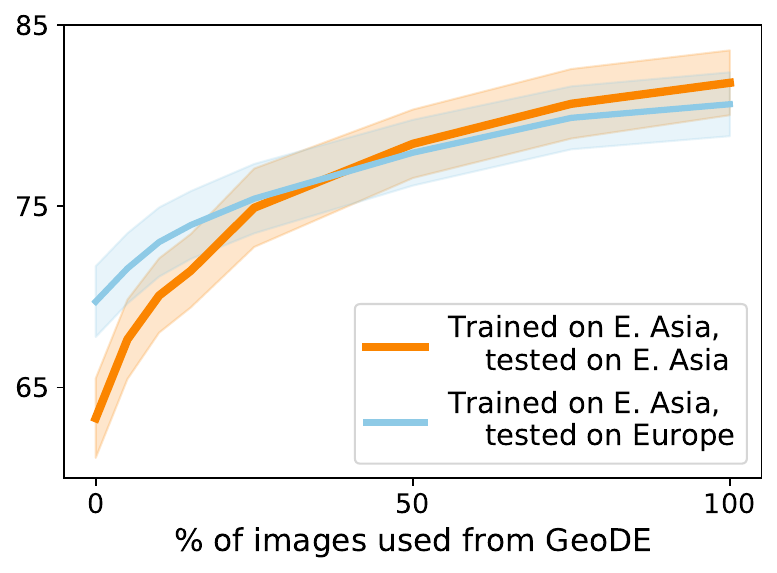} \\         \includegraphics[width=0.3\linewidth]{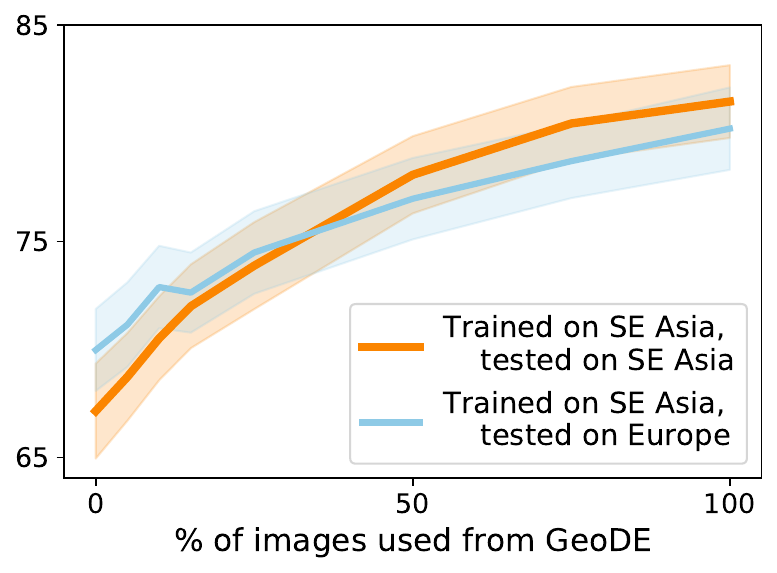}  &
        \includegraphics[width=0.3\linewidth]{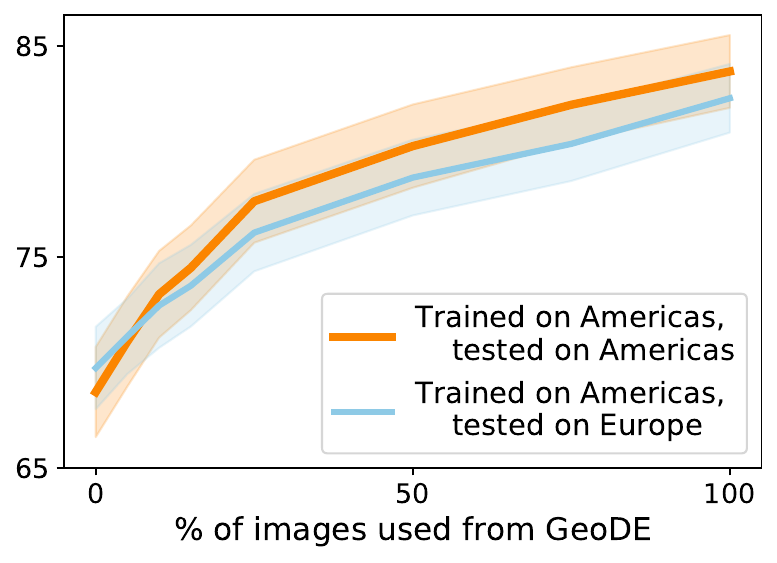} &         \includegraphics[width=0.3\linewidth]{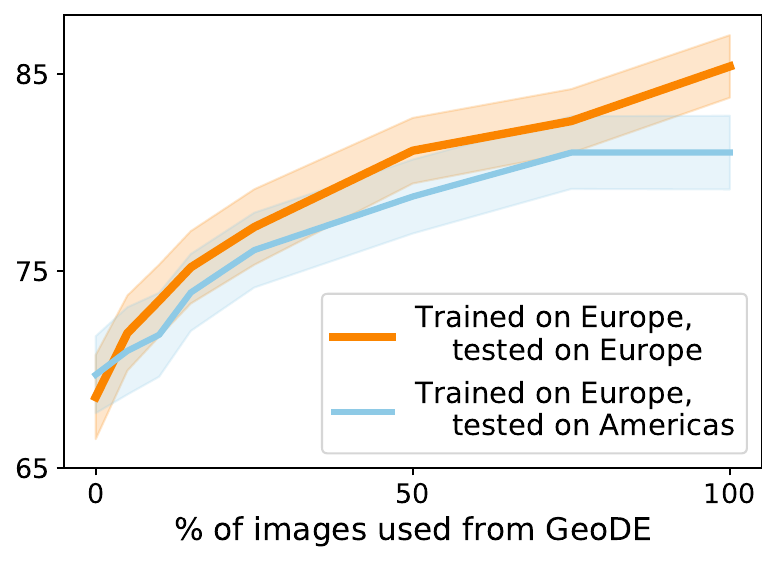}  \\    \end{tabular}
    \caption{We visualize the increase in accuracy as images are incrementally added in from a region. We find that while adding any \appen regional images increases the performance of the model on European images, it has a larger effect on the region the images were drawn from. }
    \label{fig:incr_rest}
\end{figure}

\subsection{Results from finetuning a ResNet50 model}
\smallsec{Implementation details} We use a ResNet50~\cite{He2016resnet} model pretrained on Imagenet and fine tune the weights using different fractions of the ImageNet and \appen datasets as mentioned in Sec. 7 in the main paper. We train the model with an SGD optimizer, learning rate = 0.1, and momentum=0.9. Other implementation details remain the same as before. 

\smallsec{Results}
While the overall trend of the results are the same, we see that these results are slightly noisier, potentially because the model overfits to the small training set (Fig.~\ref{fig:incr_finetune}). 
\begin{figure}
    \centering
    \begin{tabular}{ccc}
        \includegraphics[width=0.3\linewidth]{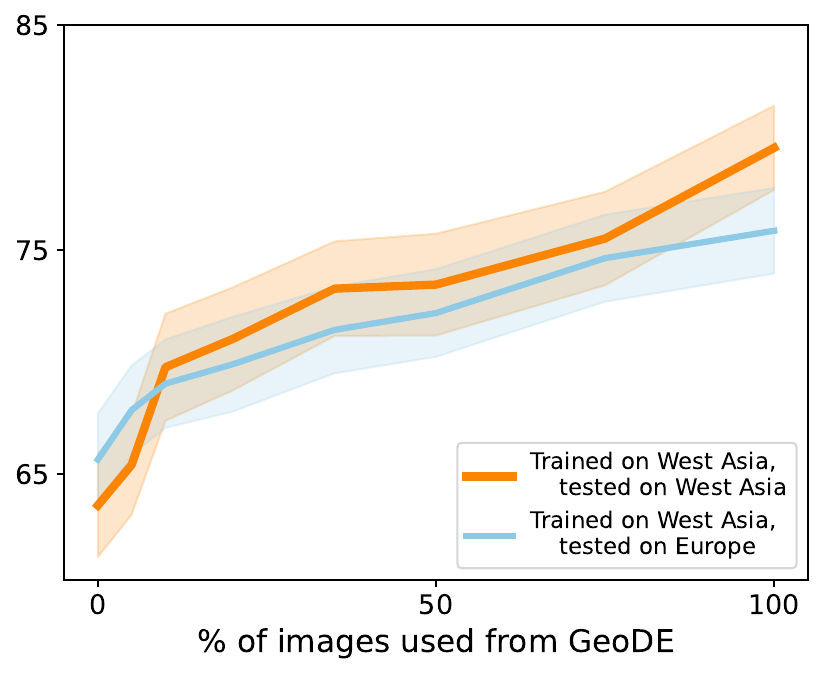} &         \includegraphics[width=0.3\linewidth]{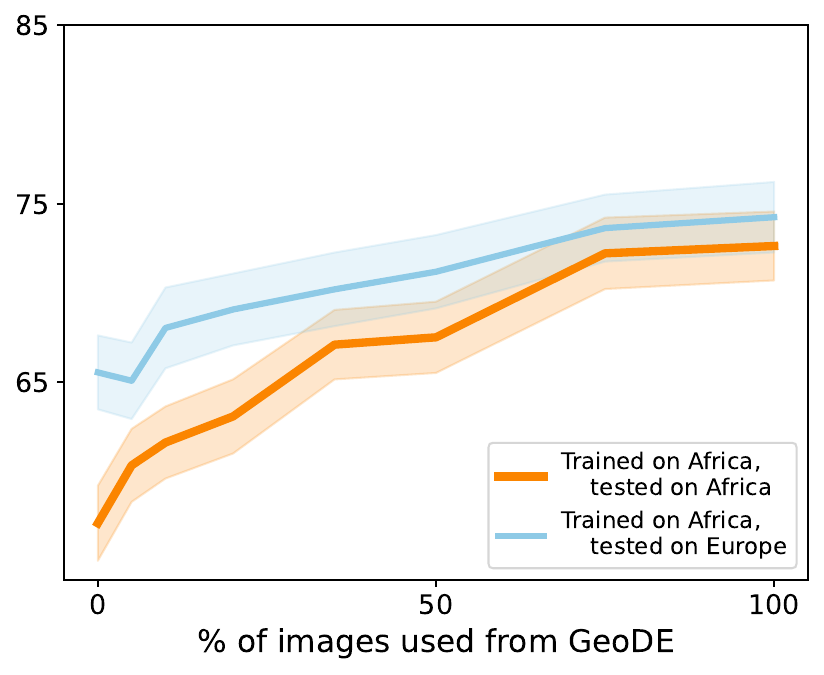}  &
        \includegraphics[width=0.3\linewidth]{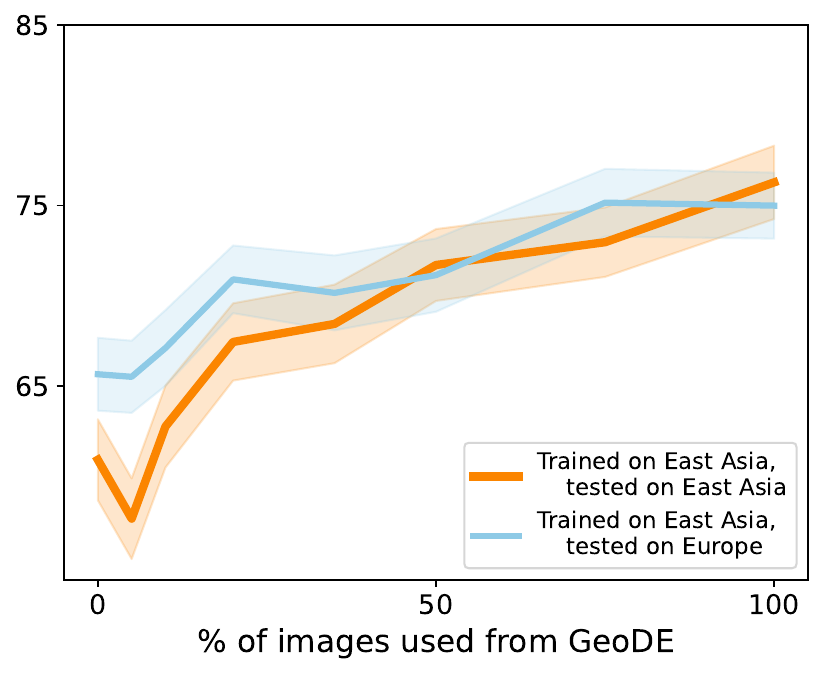} \\
        \includegraphics[width=0.3\linewidth]{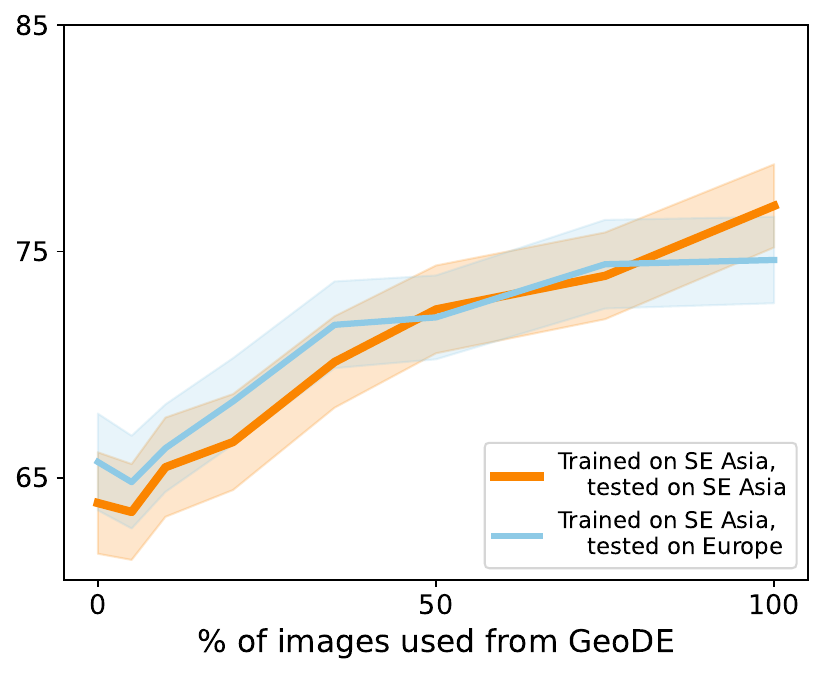}  &
        \includegraphics[width=0.3\linewidth]{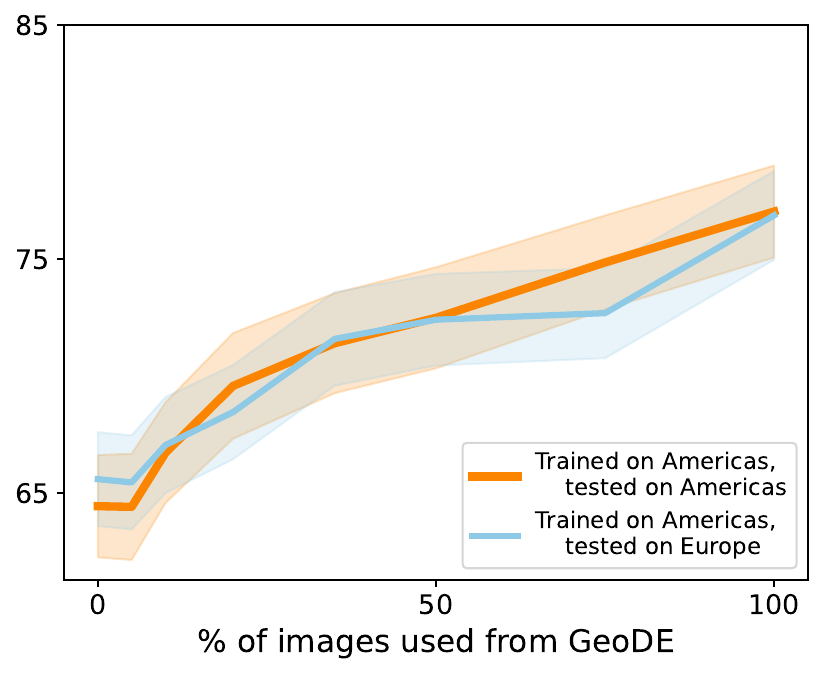} &        \includegraphics[width=0.3\linewidth]{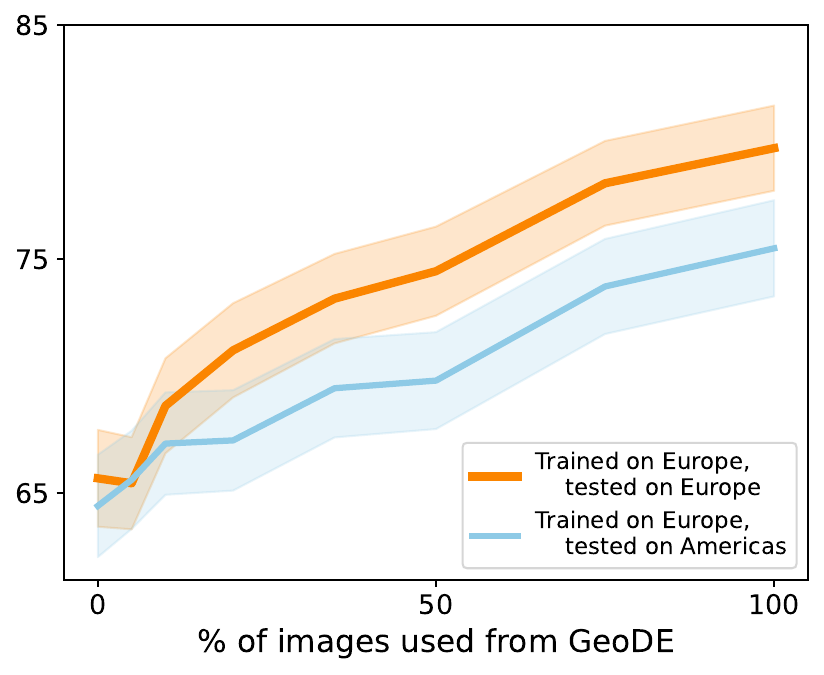}  \\
                \end{tabular}
    \caption{We visualize the increase in accuracy as images are incrementally added in from a region when finetuning a ResNet50 model. Similar to Fig. 10 in the main text, we see that adding in \appen images increases performance more in the region than a control.  }
    \label{fig:incr_finetune}
\end{figure}

\new{
\subsection{Performance on individual countries}

We measure the performance of models on individual countries within each region of GeoDE. We use two different models: first, we use the CLIP~\cite{radford_clip} model to understand performance of current models on GeoDE. Next, we train a simple linear model of features extracted from PASS~\cite{asano21pass} for the GeoDE dataset, and measure performance on individual countries. Implementation details for the CLIP model are the same as in Sec. 6 in the main paper, implementation details for the GeoDE model are as in section 7.1 of the main paper. Accuracies are computed for all countries with at least 25 images in the test set. 

Tab.~\ref{tab:ind_countries} summarizes our results for both. When using a pre-trained model like CLIP, we do notice differences between accuracies of countries within each region, however, we note that the overall trend in accuracy remains roughly the same. When training with images from GeoDE, discrepancies between countries within each region further reduces.  

\begin{table}[t]
\resizebox{\linewidth}{!}{
\begin{tabular}{ccccccccc}
\toprule 
 & \multicolumn{4}{c}{CLIP~\cite{radford_clip} model} & \multicolumn{4}{c}{Train on GeoDE}       \\
\cmidrule(l{1em}r{1em}){2-5} \cmidrule(l{1em}r{1em}){6-9}
 & Average  & STD  & Minimum & Country & Average & STD & Minimum & Country \\ 
 \midrule
Africa & 79.4 & 2.9 & 75.4 & Nigeria & 86.7 & 1.6 & 84.5 & Egypt \\
Americas & 84.4 & 0.2 & 84.1 & Argentina & 88.8 & 1.0 & 87.7 & Argentina \\
EastAsia & 80.2 & 2.0 & 77.2 & China & 89.0 & 1.1 & 87.9 & Japan \\
Europe & 85.9 & 3.6 & 78.3 & Portugal & 91.9 & 1.8 & 89.2 & United Kingdom \\
SouthEastAsia & 82.7 & 1.1 & 81.4 & Indonesia & 88.3 & 1.0 & 87.0 & Indonesia \\
WestAsia & 82.3 & 2.0 & 79.1 & Jordan & 88.9 & 3.6 & 84.2 & Saudi Arabia\\ \bottomrule   
\end{tabular}}
    \caption{We measure the performance of a CLIP based model as well as model trained on GeoDE on individual countries within each region. For each region, we report the average accuracy, the standard deviation, and the country with the minimum accuracy for all countries with over 25 images. We find that while a pretrained model does show significant discrepancies among countries within the same region, training on GeoDE data does reduce this. }
    \label{tab:ind_countries}

\end{table}
}

\section{More details about the dataset}
\label{supp_sec:sample_images}
In this supplementary section, we provide counts of the objects per region in \appen as well as more examples of images from this dataset. 

As mentioned before, \appen is mostly balanced across both region and object: for most part, we were able to get atleast 150 images per region per object, with a few exceptions (``wheelbarrow'' in 2 regions; ``monument'', ``boat'' and ``flag'' in 1 region). See Tab.~\ref{tab:counts} for full counts. 

\begin{table*}[t]
    \centering
\resizebox{\linewidth}{!}{\begin{tabular}{l| cccccc}
\toprule
 & West Asia & Africa & East Asia & Southeast Asia & Americas & Europe \\
 \midrule
backyard & 216 & 670 & 192 & 218 & 217 & 226 \\
bag & 267 & 397 & 370 & 593 & 298 & 437 \\
bicycle & 237 & 257 & 298 & 235 & 228 & 241 \\
boat & \textbf{162} & 227 & \textbf{174} & 237 & \textbf{84} & 222 \\
bus & 203 & 240 & 223 & 214 & 217 & 226 \\
candle & 232 & 244 & 220 & 239 & 188 & 270 \\
car & 242 & 331 & 276 & 235 & 273 & 363 \\
chair & 279 & 365 & 326 & 512 & 344 & 349 \\
cleaning equipment & 259 & 284 & 307 & 305 & 270 & 361 \\
cooking pot & 216 & 270 & 228 & 202 & 213 & 304 \\
dog & 219 & 194 & 185 & 244 & 206 & 193 \\
dustbin & 220 & 423 & 266 & 203 & 271 & 294 \\
fence & 259 & 322 & 244 & 302 & 226 & 282 \\
flag & 206 & 265 & \textbf{139} & 223 & 206 & 272 \\
front\_door & 210 & 254 & 216 & 224 & 200 & 235 \\
hairbrush/comb & 269 & 255 & 307 & 300 & 290 & 431 \\
hand soap & 222 & 208 & 277 & 191 & 245 & 362 \\
hat & 209 & 297 & 337 & 316 & 294 & 336 \\
house & 199 & 437 & 208 & 195 & 277 & 194 \\
jug & 217 & 211 & 186 & 249 & 236 & 194 \\
light fixture & 234 & 344 & 248 & 209 & 191 & 300 \\
light switch & 215 & 240 & 246 & 273 & 273 & 234 \\
lighter & 221 & 312 & 225 & 237 & 217 & 268 \\
medicine & 242 & 286 & 310 & 330 & 328 & 300 \\
monument & \textbf{161} & 191 & 186 & 183 & 254 & 245 \\
plate of food & 211 & 480 & 294 & 364 & 241 & 304 \\
religious building & 222 & 230 & 204 & 226 & 197 & 229 \\
road sign & 226 & 416 & 258 & 270 & 235 & 284 \\
spices & 243 & 250 & 331 & 216 & 290 & 300 \\
stall & \textbf{143} & 215 & 203 & 227 & 197 & 221 \\
storefront & 209 & 306 & 191 & 240 & 243 & 204 \\
stove & 199 & 553 & 191 & 262 & 206 & 282 \\
streetlight / lantern & 202 & 346 & 211 & 196 & 208 & 227 \\
toothbrush & 264 & 258 & 330 & 361 & 337 & 270 \\
toothpaste / toothpowder & 209 & 288 & 269 & 230 & 245 & 315 \\
toy & 224 & 221 & 280 & 292 & 323 & 287 \\
tree & 226 & 308 & 245 & 357 & 300 & 328 \\
truck & 205 & 246 & 207 & 231 & 212 & 225 \\
waste container & 231 & 213 & 209 & 213 & 211 & 253 \\
wheelbarrow & \textbf{122} & 267 & \textbf{130} & 197 & \textbf{152} & 243 \\
\bottomrule
\end{tabular}}
    \caption{ We show the counts of objects per region in \appenn. \textbf{Bolded} are the ones categories for which we were not able to get 175 images per region.}
    \label{tab:counts}
\end{table*}

We also provide more examples of the images from \appen in the Figures~\cref{fig:backyard1,fig:backyard2,fig:bicycle1,fig:bicycle2,fig:boat1,fig:boat2,fig:cleaning_equipment1,fig:cleaning_equipment2,fig:spices1,fig:spices2,fig:stove1,fig:stove2,fig:waste_container1,fig:waste_container2}. 

\begin{figure*}
    \centering
    \begin{tabular}{lc}
    \toprule
    \multicolumn{2}{c}{\textbf{Backyard}}\\\toprule
    West Asia & \raisebox{-0.5\totalheight}{\includegraphics[width=0.72\textwidth]{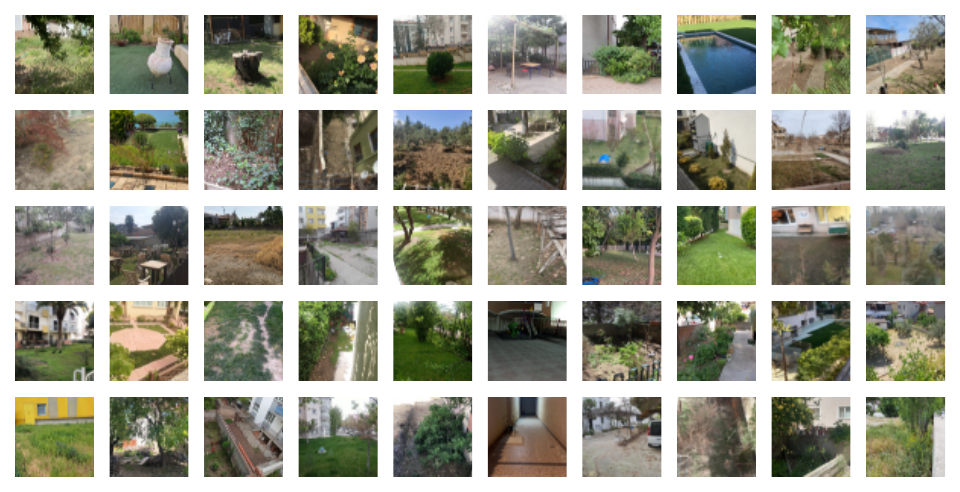}} \\ \midrule
    Africa & \raisebox{-0.5\totalheight}{\includegraphics[width=0.72\textwidth]{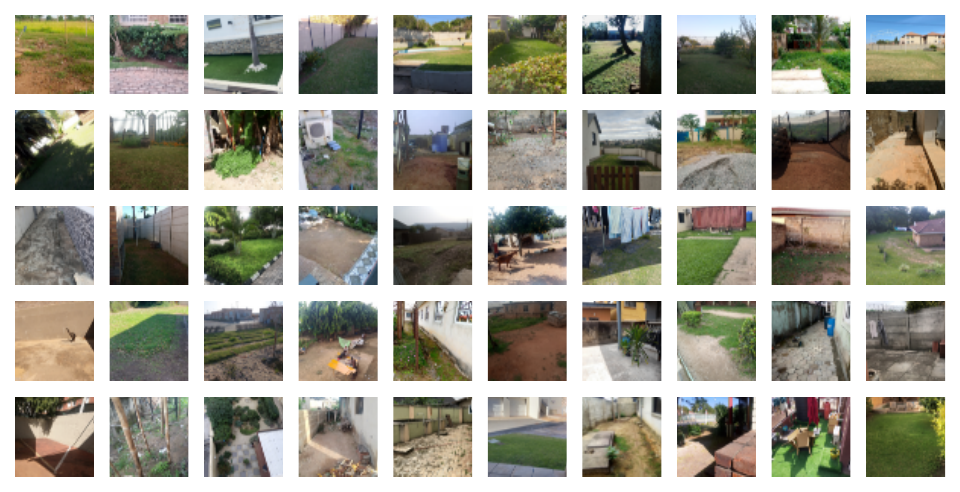}} \\ \midrule
    East Asia & \raisebox{-0.5\totalheight}{\includegraphics[width=0.72\textwidth]{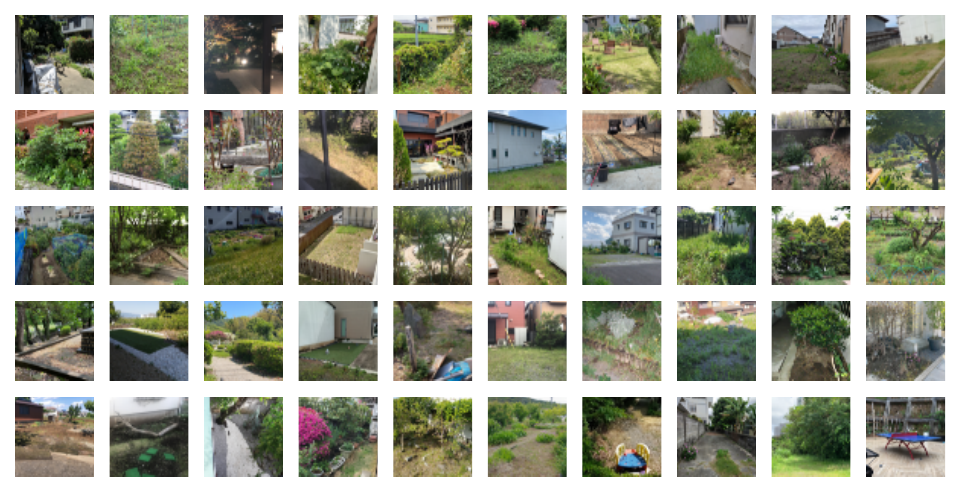}} \\ \midrule  
    \end{tabular}
    
    \caption{Randomly chosen images for ``backyard'' for 3 regions. We notice that some of these are backyards made of concrete (West Asia: r2c1, r3c4, etc., Africa: r3c1 ,r4c9, r5c3, etc.,) 
 or do not contain lawns (West Asia: r3c1, r2c5, etc., Africa:r5c1-5, etc., East Asia: r2c4, r5c1, r5c4-6, etc.)}
    \label{fig:backyard1}
\end{figure*}

\begin{figure*}
    \centering
    \begin{tabular}{lc}
    \toprule
    \multicolumn{2}{c}{\textbf{Backyard}}\\\toprule
    Southeast Asia & \raisebox{-0.5\totalheight}{\includegraphics[width=0.72\textwidth]{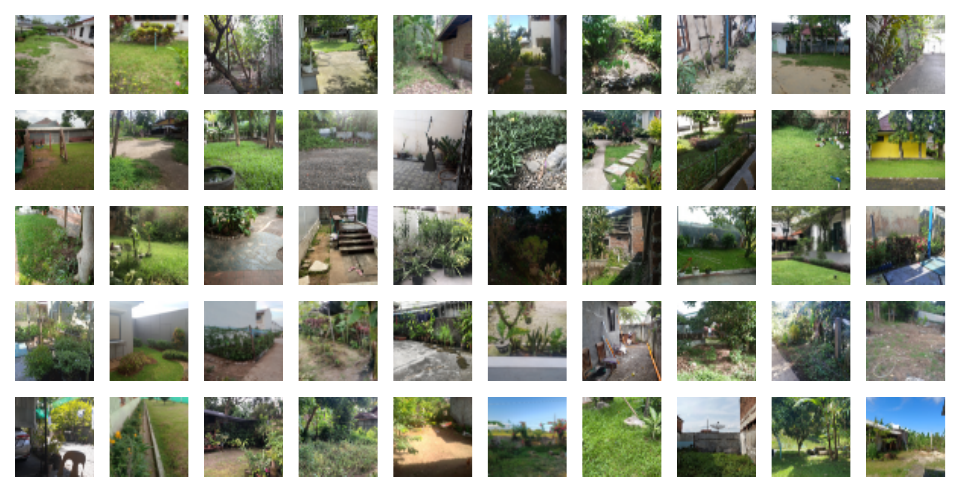}} \\ \midrule
    Americas & \raisebox{-0.5\totalheight}{\includegraphics[width=0.72\textwidth]{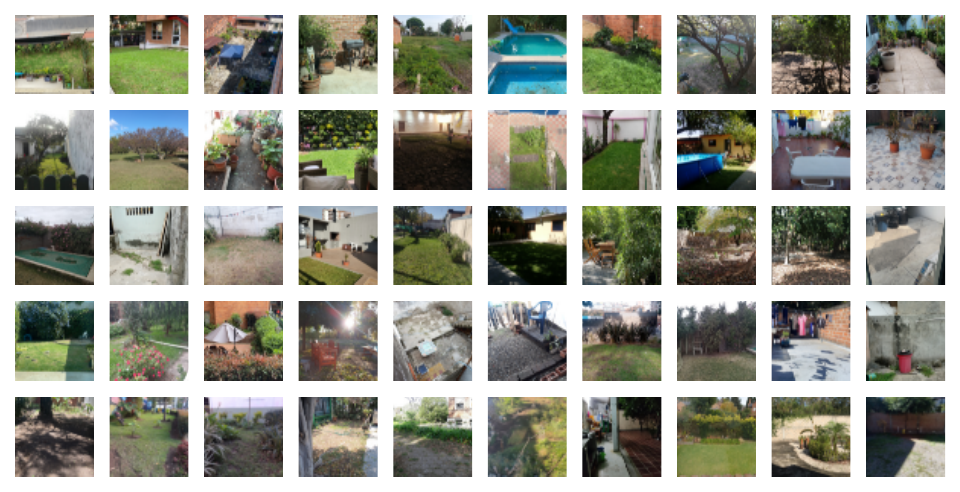}} \\ \midrule
    Europe & \raisebox{-0.5\totalheight}{\includegraphics[width=0.72\textwidth]{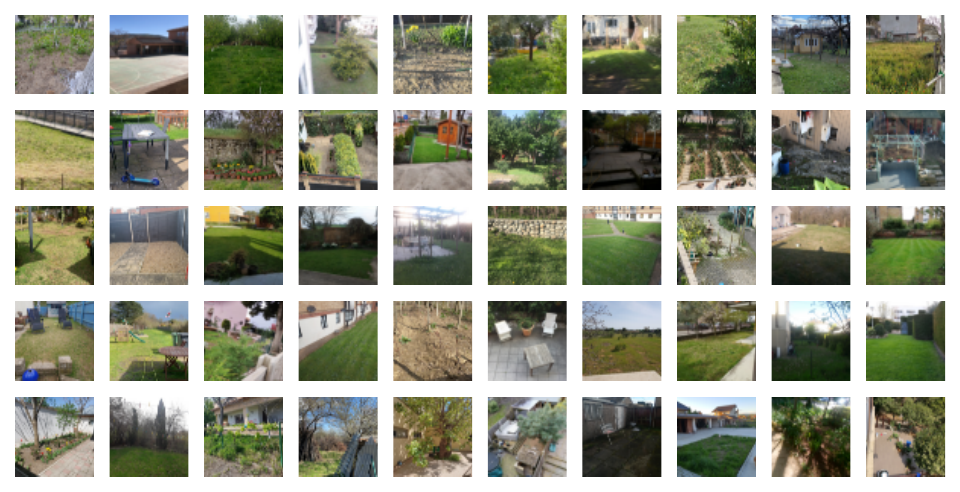}} \\ \midrule  
    \end{tabular}
    \caption{Randomly chosen images for ``backyard'' for the 3 other regions. Again, we see that regions tend to have backyards made of concrete or paved (Southeast Asia: r1c8, r2c5, r3c3 as examples, Americas: r1c9, r1c10, r2c10, etc., Europe: r3c2, r4c5, etc ), or do not contain lawns (Southeast Asia: r1c1, r1c9, etc., Americas:r3c2, r3c3, etc., Europe: r3c2, r5c9, etc.) }
    \label{fig:backyard2}
\end{figure*}

\begin{figure*}
    \centering
    
    \begin{tabular}{lc}
    \toprule
    \multicolumn{2}{c}{\textbf{Bicycle}}\\\toprule
    West Asia & \raisebox{-0.5\totalheight}{\includegraphics[width=0.72\textwidth]{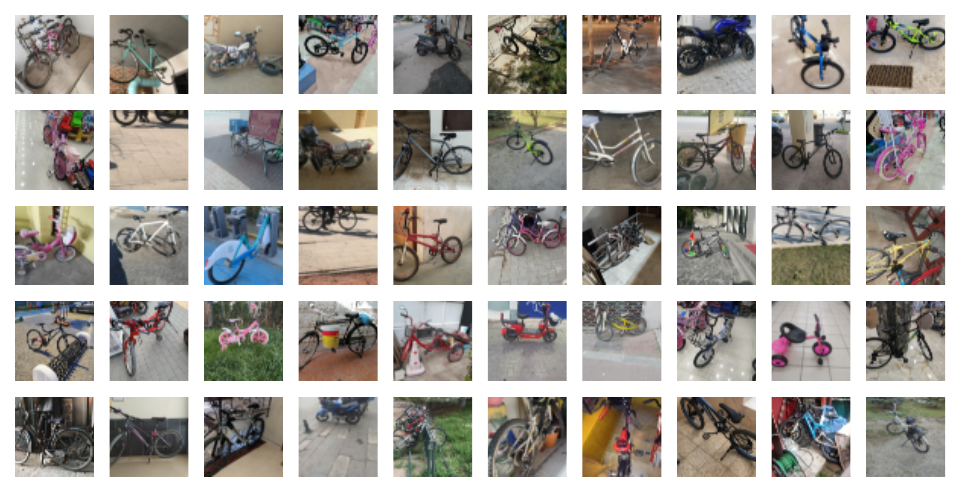}} \\ \midrule
    Africa & \raisebox{-0.5\totalheight}{\includegraphics[width=0.72\textwidth]{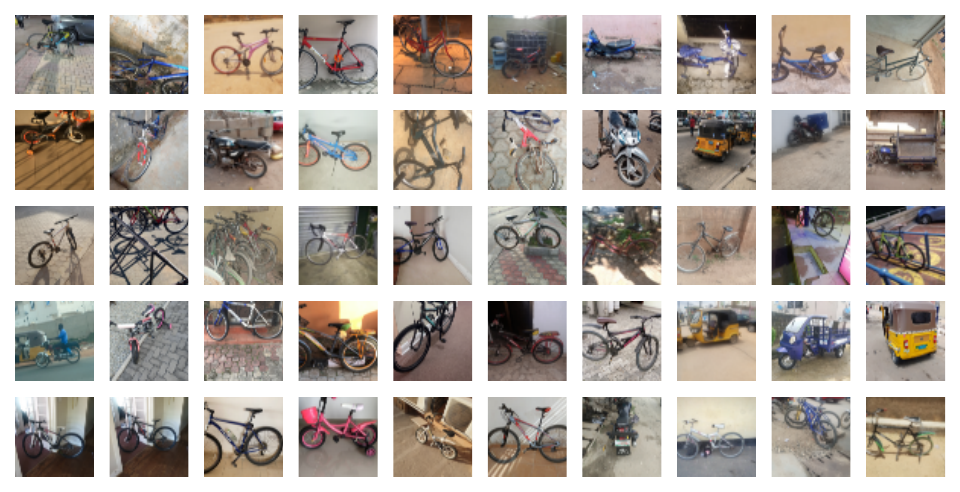}} \\ \midrule
    East Asia & \raisebox{-0.5\totalheight}{\includegraphics[width=0.72\textwidth]{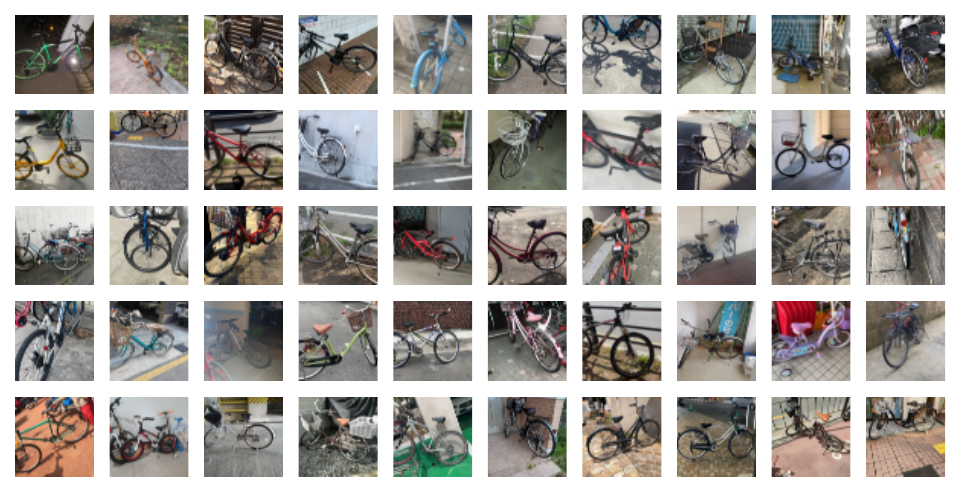}} \\ \midrule  
    \end{tabular}
    
    \caption{Randomly chosen images for ``bicycle'' for 3 regions. While most images are of standard bicycles, we notice a couple of interesting images: tricycle (West Asia: r4c9), rickshaws (Africa: r2c8, r2c10, r4c8, r4c10), and motorized cycles (West Asia: r1c8). There are also a lot of children's bicycles.}
    \label{fig:bicycle1}
\end{figure*}

\begin{figure*}
    \centering
    \begin{tabular}{lc}
    \toprule
    \multicolumn{2}{c}{\textbf{Bicycle}}\\\toprule
    Southeast Asia & \raisebox{-0.5\totalheight}{\includegraphics[width=0.72\textwidth]{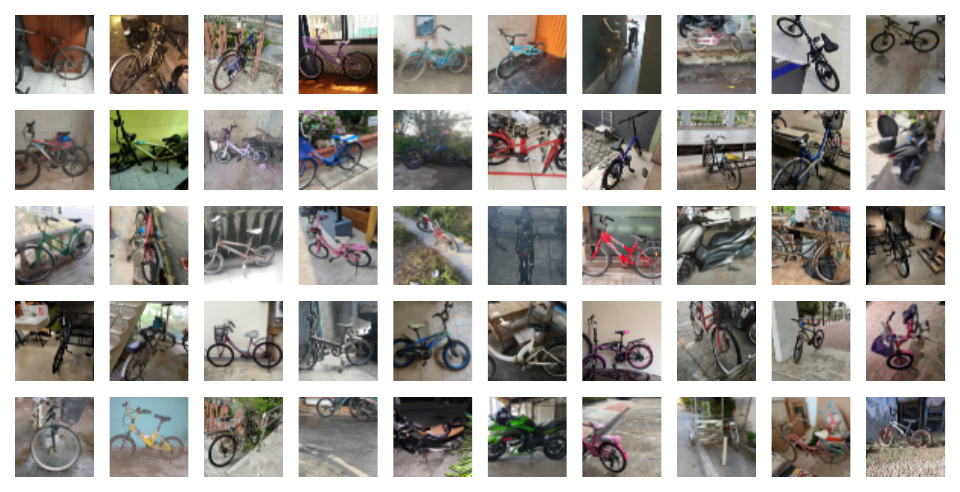}} \\ \midrule
    Americas & \raisebox{-0.5\totalheight}{\includegraphics[width=0.72\textwidth]{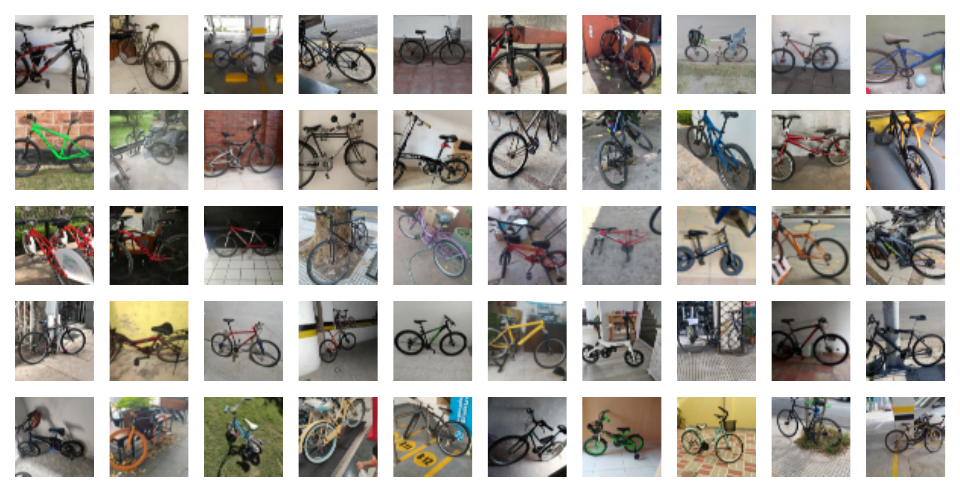}} \\ \midrule
    Europe & \raisebox{-0.5\totalheight}{\includegraphics[width=0.72\textwidth]{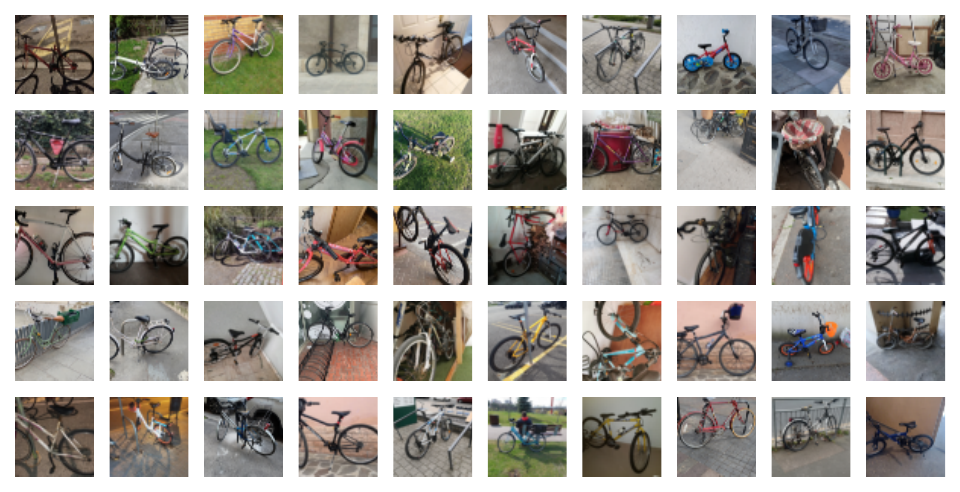}} \\ \midrule  
    \end{tabular}
    \caption{Randomly chosen images for ``bicycle'' for the 3 other regions. We see more motorized cycles (Southeast Asia: r5c6) as well as several children's bicycles. }
    \label{fig:bicycle2}
\end{figure*}

\begin{figure*}
    \centering
    
    \begin{tabular}{lc}
    \toprule
    \multicolumn{2}{c}{\textbf{Boat}}\\\toprule
    West Asia & \raisebox{-0.5\totalheight}{\includegraphics[width=0.72\textwidth]{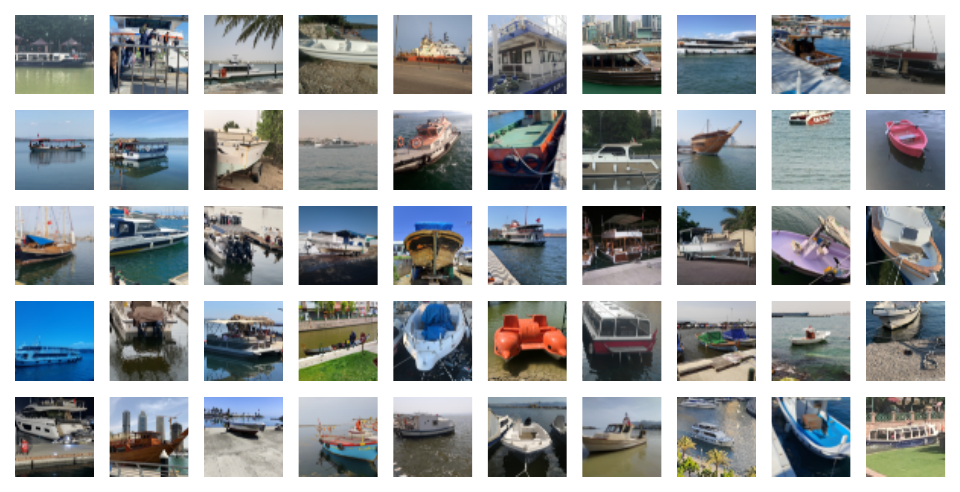}} \\ \midrule
    Africa & \raisebox{-0.5\totalheight}{\includegraphics[width=0.72\textwidth]{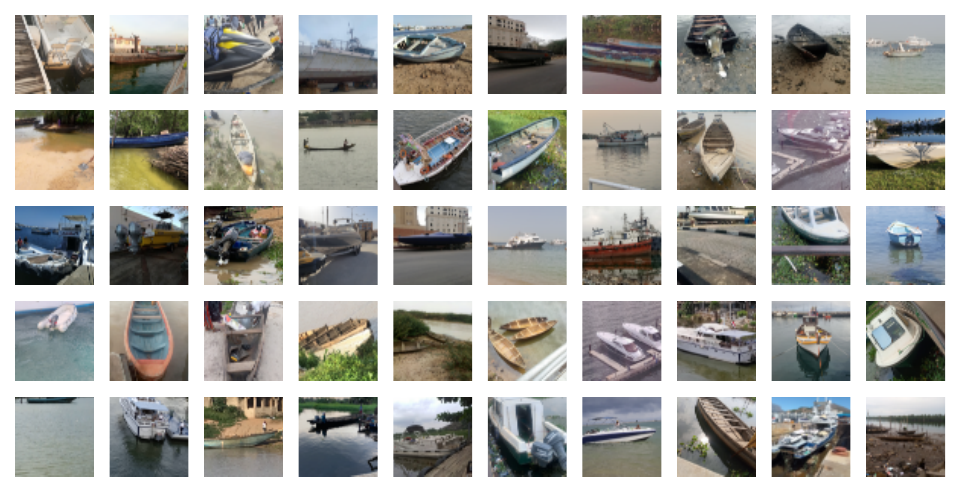}} \\ \midrule
    East Asia & \raisebox{-0.5\totalheight}{\includegraphics[width=0.72\textwidth]{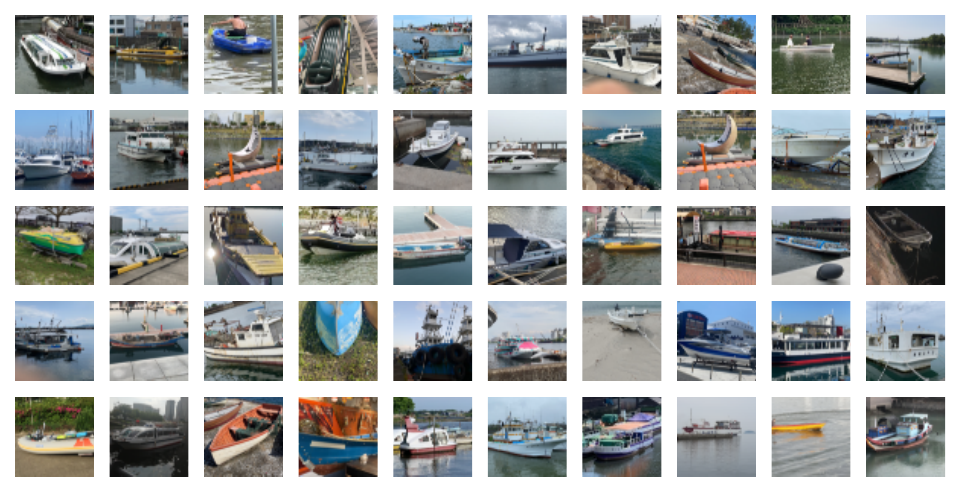}} \\ \midrule  
    \end{tabular}
    
    \caption{Randomly chosen images for ``boat'' for 3 regions. We see a variety of boats including larger ships in West Asia (r1c1, r1c2, r1c5, r4c1,r5c1), smaller kayaks and canoes in Africa (r1c8-9, r2c1-4, r4c2 , etc ), and a mix in East Asia.}
    \label{fig:boat1}
\end{figure*}

\begin{figure*}
    \centering
    \begin{tabular}{lc}
    \toprule
    \multicolumn{2}{c}{\textbf{Boat}}\\\toprule
    Southeast Asia & \raisebox{-0.5\totalheight}{\includegraphics[width=0.72\textwidth]{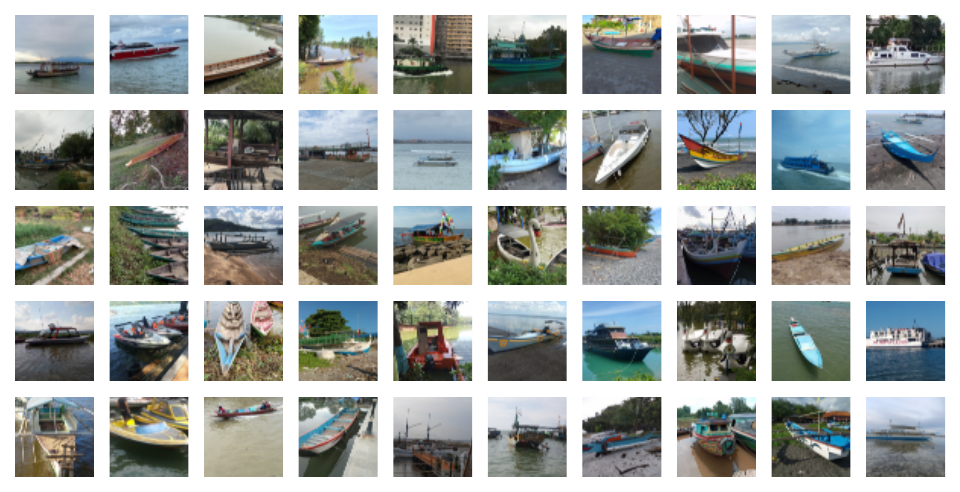}} \\ \midrule
    Americas & \raisebox{-0.5\totalheight}{\includegraphics[width=0.72\textwidth]{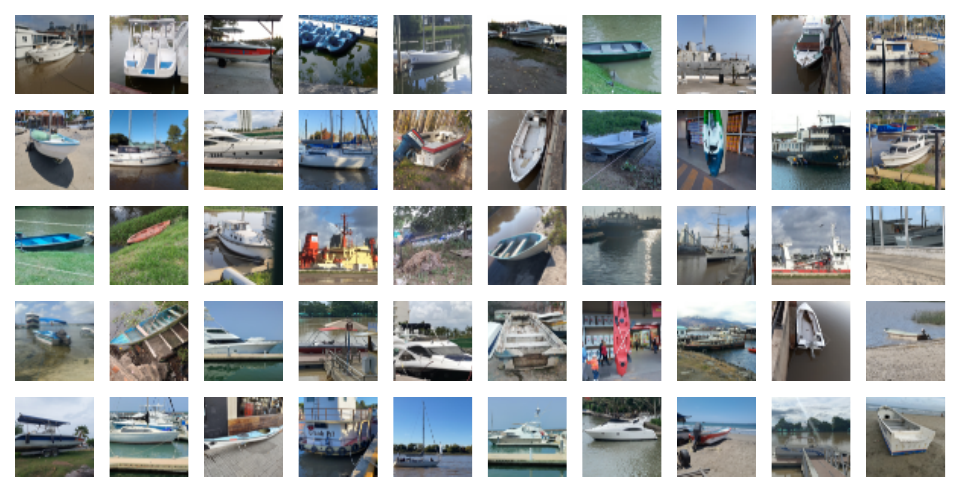}} \\ \midrule
    Europe & \raisebox{-0.5\totalheight}{\includegraphics[width=0.72\textwidth]{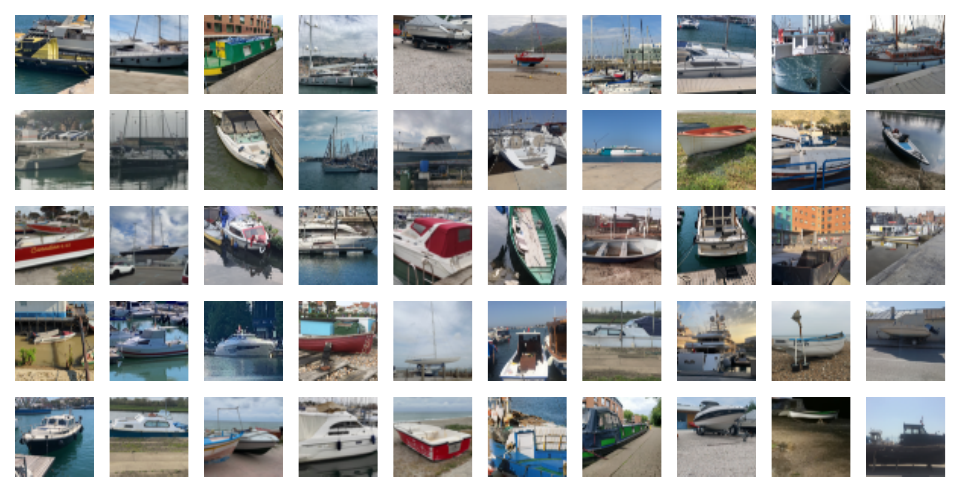}} \\ \midrule  
    \end{tabular}
    \caption{Randomly chosen images for ``boat'' for the 3 other regions. We again see a variety of boats ranging from motor boats in Europe and the Americas to smaller boats in Southeast Asia. }
    \label{fig:boat2}
\end{figure*}

\begin{figure*}
    \centering
    
    \begin{tabular}{lc}
    \toprule
    \multicolumn{2}{c}{\textbf{Cleaning equipment}}\\\toprule
    West Asia & \raisebox{-0.5\totalheight}{\includegraphics[width=0.72\textwidth]{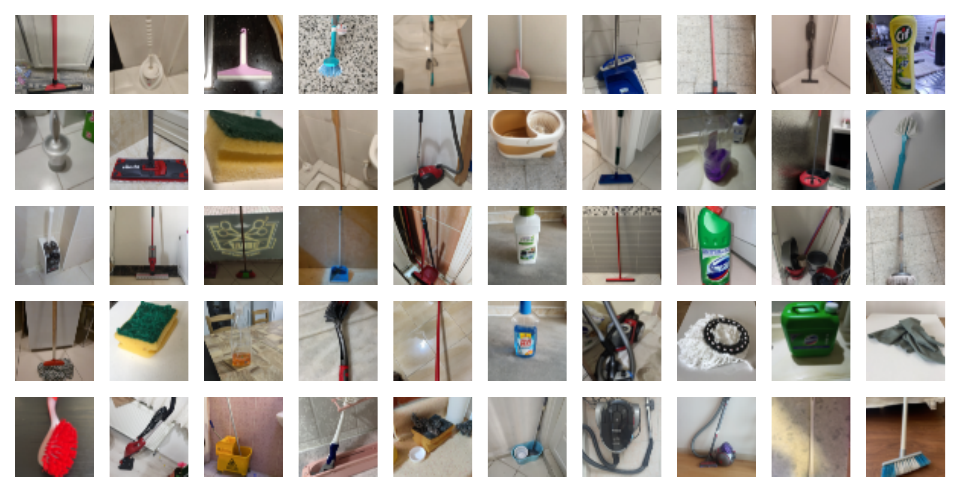}} \\ \midrule
    Africa & \raisebox{-0.5\totalheight}{\includegraphics[width=0.72\textwidth]{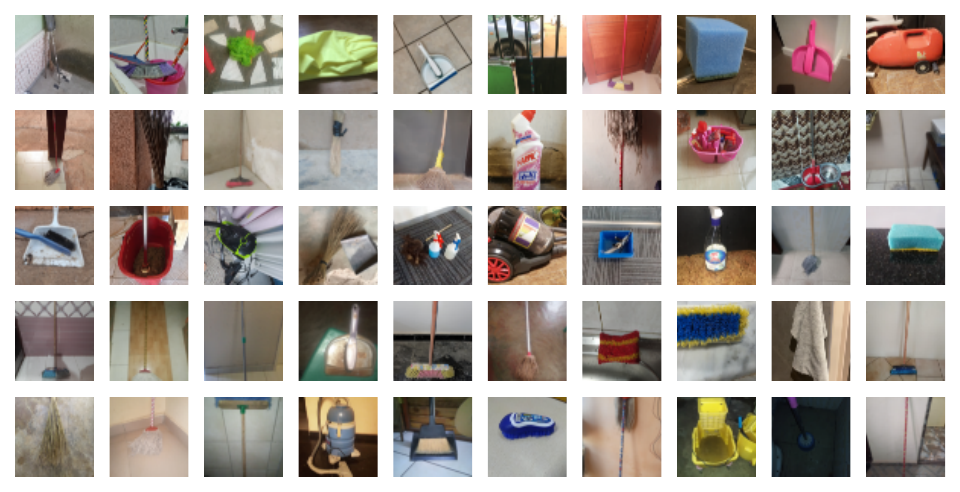}} \\ \midrule
    East Asia & \raisebox{-0.5\totalheight}{\includegraphics[width=0.72\textwidth]{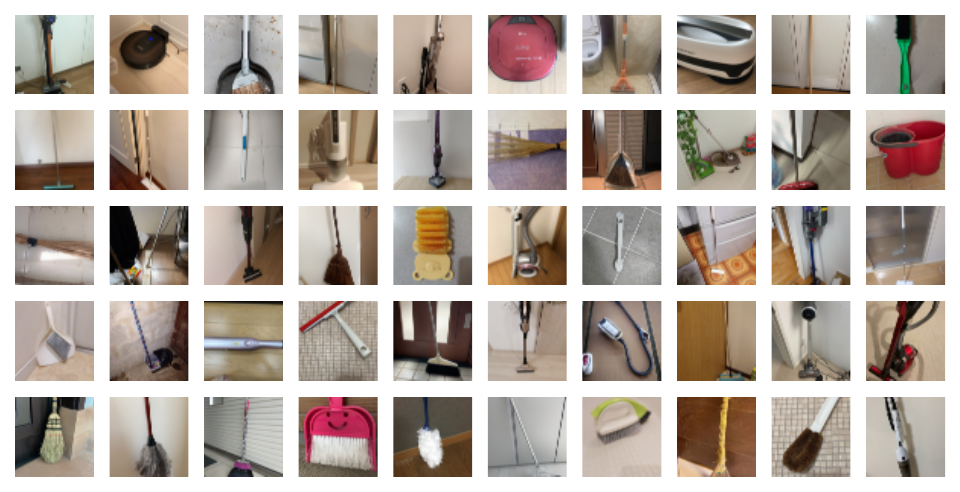}} \\ \midrule  
    \end{tabular}
    
    \caption{Randomly chosen images for ``cleaning equipment'' for 3 regions. This appears to be a diverse category within all regions containing images of mops, buckets, products, brooms, etc.}
    \label{fig:cleaning_equipment1}
\end{figure*}

\begin{figure*}
    \centering
    \begin{tabular}{lc}
    \toprule
    \multicolumn{2}{c}{\textbf{Cleaning equipment}}\\\toprule
    Southeast Asia & \raisebox{-0.5\totalheight}{\includegraphics[width=0.72\textwidth]{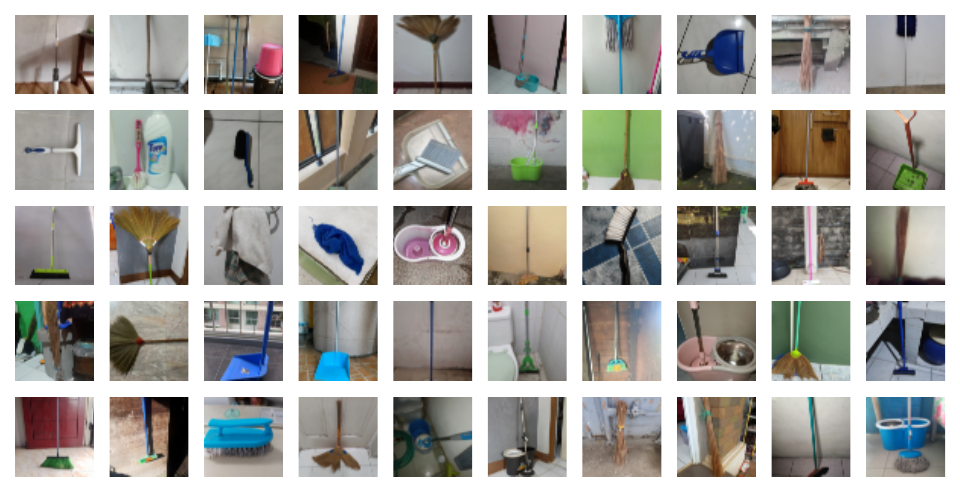}} \\ \midrule
    Americas & \raisebox{-0.5\totalheight}{\includegraphics[width=0.72\textwidth]{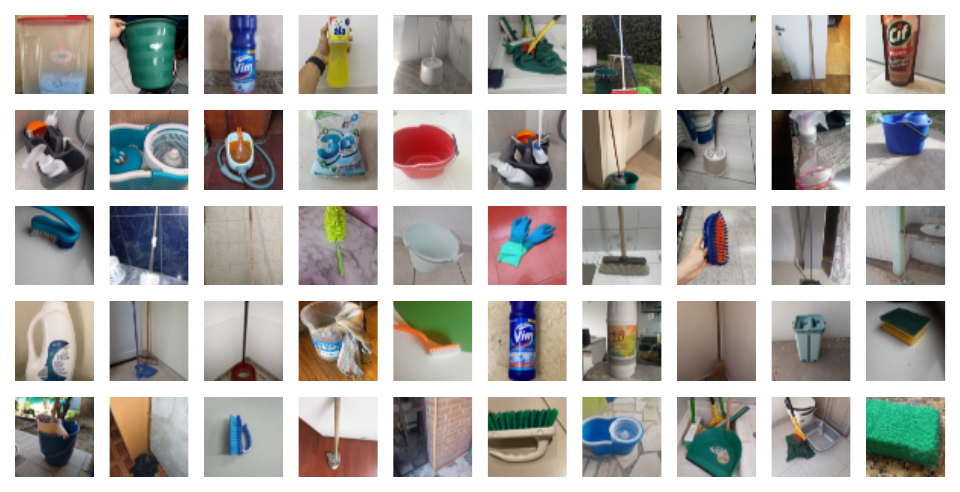}} \\ \midrule
    Europe & \raisebox{-0.5\totalheight}{\includegraphics[width=0.72\textwidth]{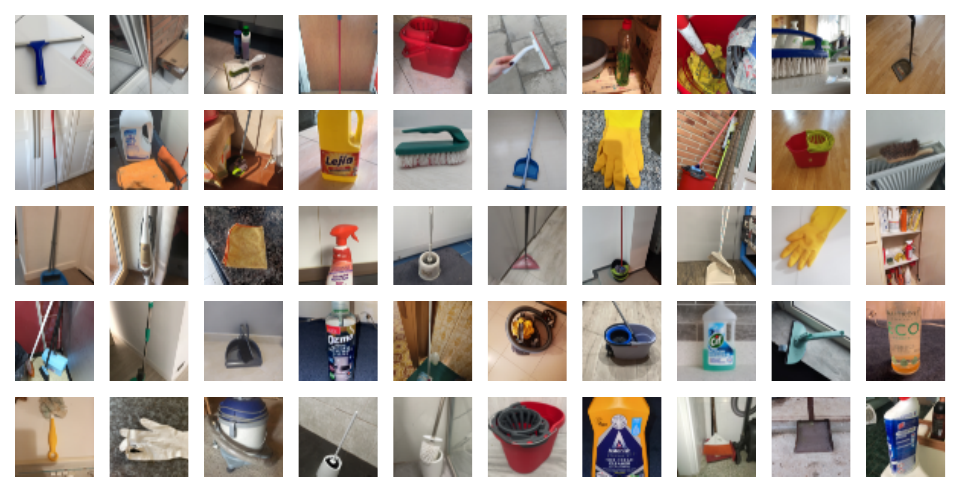}} \\ \midrule  
    \end{tabular}
    \caption{Randomly chosen images for ``cleaning equipment'' for the 3 other regions. This appears to be a diverse category within all regions containing images of mops, buckets, products, brooms, etc. }
    \label{fig:cleaning_equipment2}
\end{figure*}

\begin{figure*}
    \centering
    
    \begin{tabular}{lc}
    \toprule
    \multicolumn{2}{c}{\textbf{Spices}}\\\toprule
    West Asia & \raisebox{-0.5\totalheight}{\includegraphics[width=0.72\textwidth]{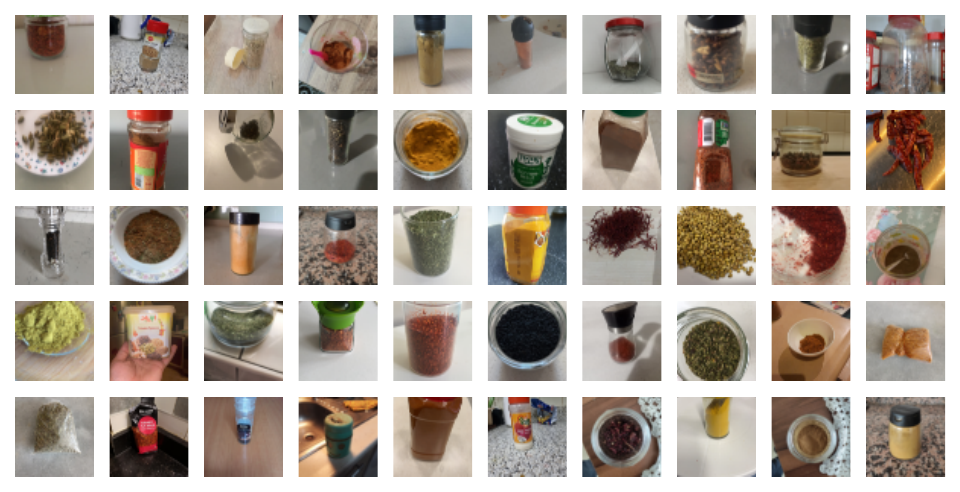}} \\ \midrule
    Africa & \raisebox{-0.5\totalheight}{\includegraphics[width=0.72\textwidth]{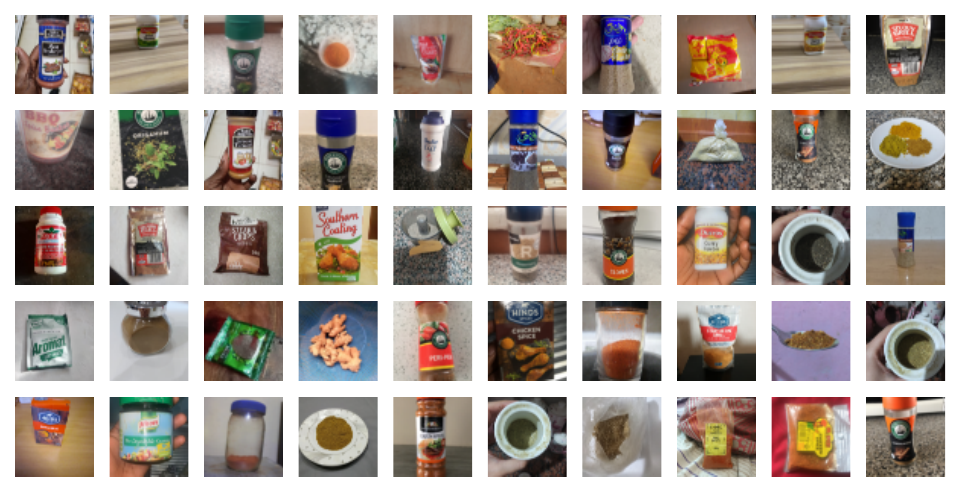}} \\ \midrule
    East Asia & \raisebox{-0.5\totalheight}{\includegraphics[width=0.72\textwidth]{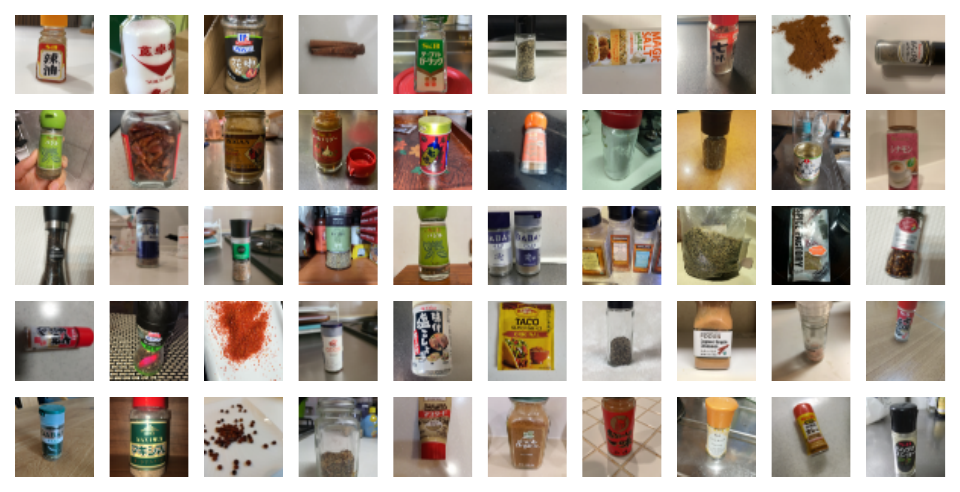}} \\ \midrule  
    \end{tabular}
    
    \caption{Randomly chosen images for ``spices'' for 3 regions. We see a wide range of containers, ranging from packets (mostly in Africa), glass jars (in West Asia) to some bottles (all regions).}
    \label{fig:spices1}
\end{figure*}

\begin{figure*}
    \centering
    \begin{tabular}{lc}
    \toprule
    \multicolumn{2}{c}{\textbf{Spices}}\\\toprule
    Southeast Asia & \raisebox{-0.5\totalheight}{\includegraphics[width=0.72\textwidth]{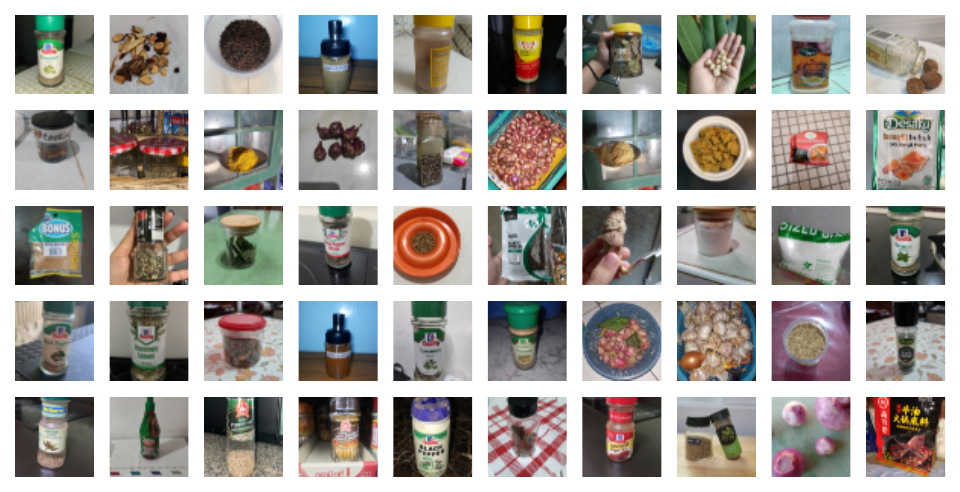}} \\ \midrule
    Americas & \raisebox{-0.5\totalheight}{\includegraphics[width=0.72\textwidth]{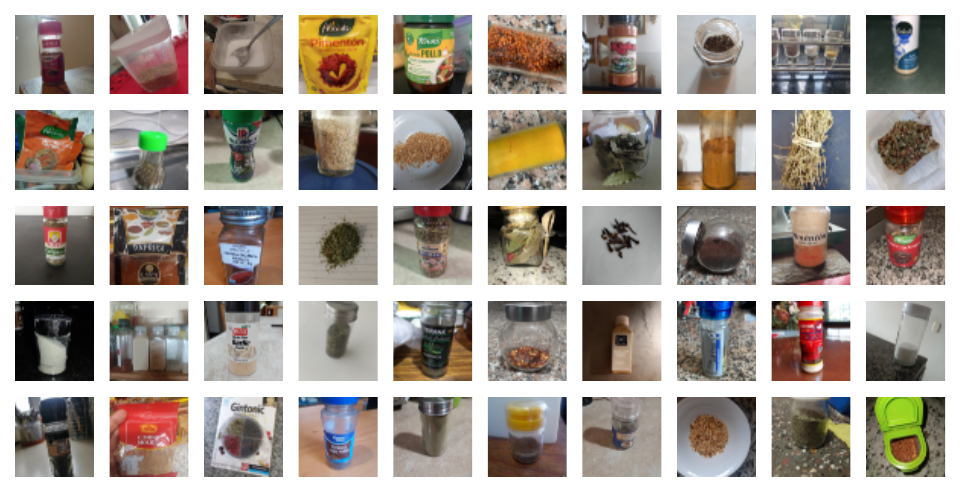}} \\ \midrule
    Europe & \raisebox{-0.5\totalheight}{\includegraphics[width=0.72\textwidth]{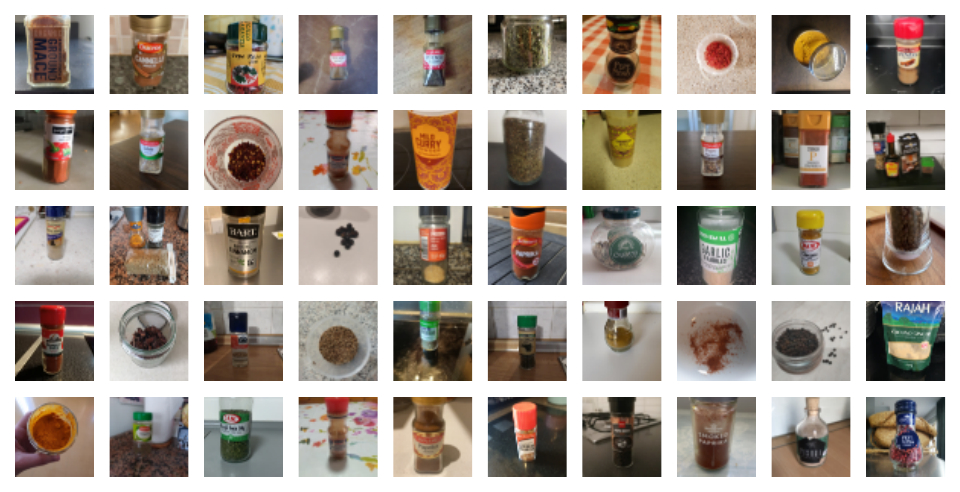}} \\ \midrule  
    \end{tabular}
    \caption{Randomly chosen images for ``spices'' for 3 regions. We see a wide range of containers, ranging from packets (some in Southeast Asia and Americas) to bottles (some in Southeast Asia)}
    \label{fig:spices2}
\end{figure*}

\begin{figure*}
    \centering
    
    \begin{tabular}{lc}
    \toprule
    \multicolumn{2}{c}{\textbf{Stove}}\\\toprule
    West Asia & \raisebox{-0.5\totalheight}{\includegraphics[width=0.72\textwidth]{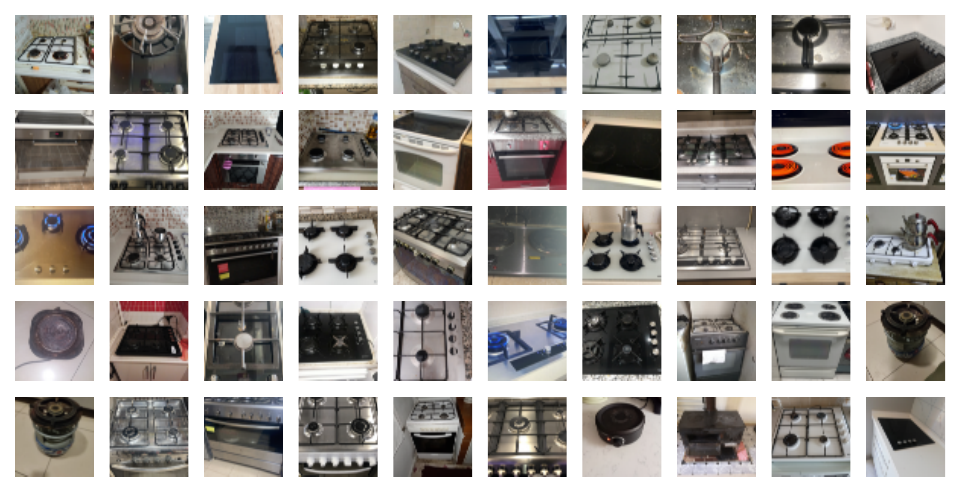}} \\ \midrule
    Africa & \raisebox{-0.5\totalheight}{\includegraphics[width=0.72\textwidth]{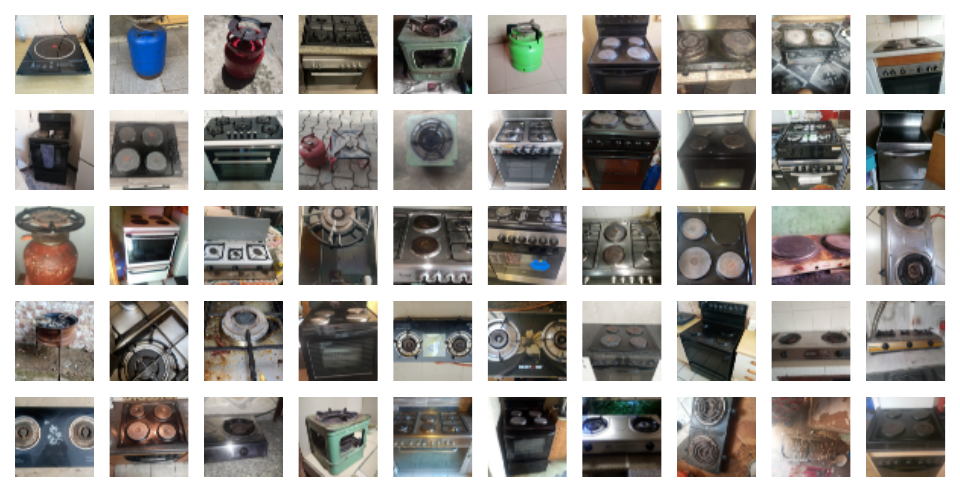}} \\ \midrule
    East Asia & \raisebox{-0.5\totalheight}{\includegraphics[width=0.72\textwidth]{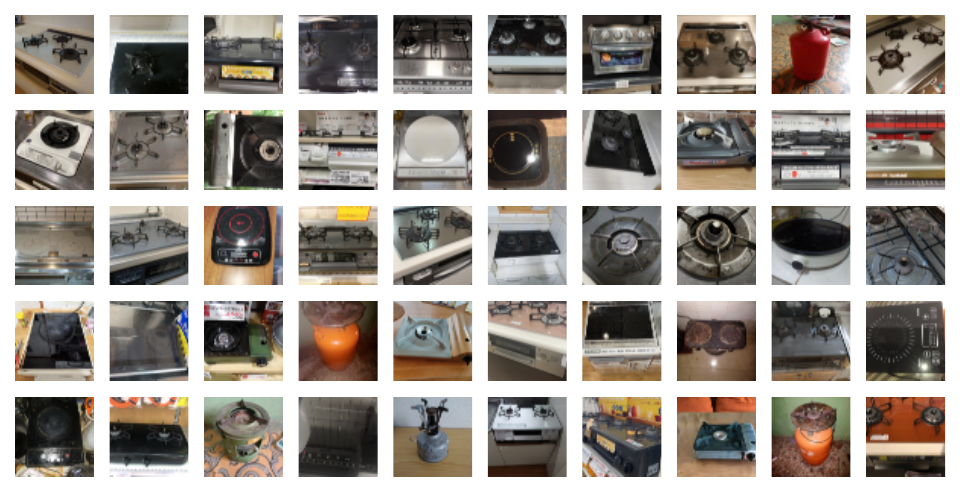}} \\ \midrule  
    \end{tabular}
    
    \caption{Randomly chosen images for ``stove'' for 3 regions. We see that Africa and East Asia contain one-burner and two burner stoves (along with 4 burner stoves). We also see a variety of stoves in terms of induction, gas, ovens, etc. }
    \label{fig:stove1}
\end{figure*}

\begin{figure*}
    \centering
    \begin{tabular}{lc}
    \toprule
    \multicolumn{2}{c}{\textbf{Stove}}\\\toprule
    Southeast Asia & \raisebox{-0.5\totalheight}{\includegraphics[width=0.72\textwidth]{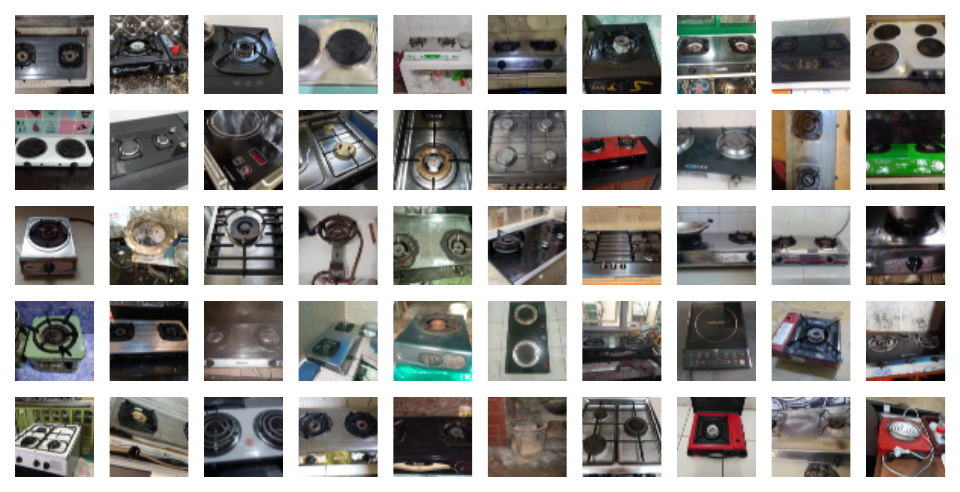}} \\ \midrule
    Americas & \raisebox{-0.5\totalheight}{\includegraphics[width=0.72\textwidth]{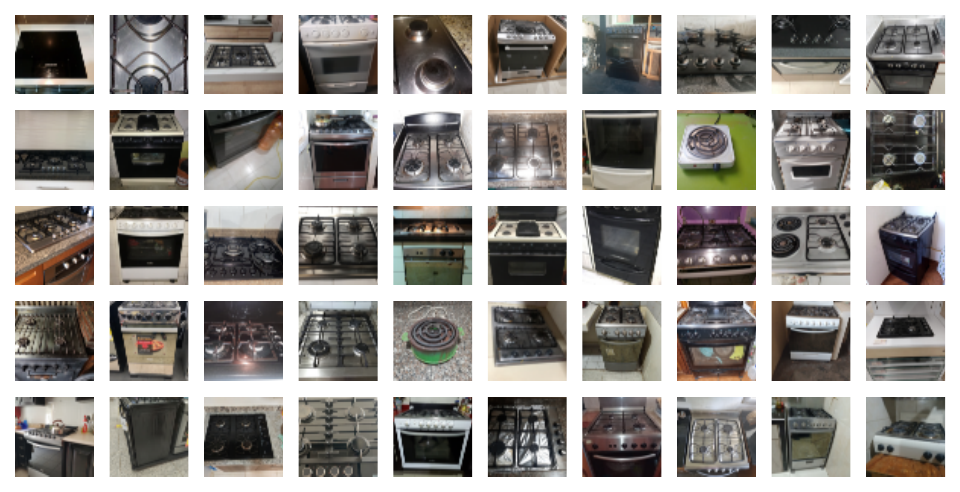}} \\ \midrule
    Europe & \raisebox{-0.5\totalheight}{\includegraphics[width=0.72\textwidth]{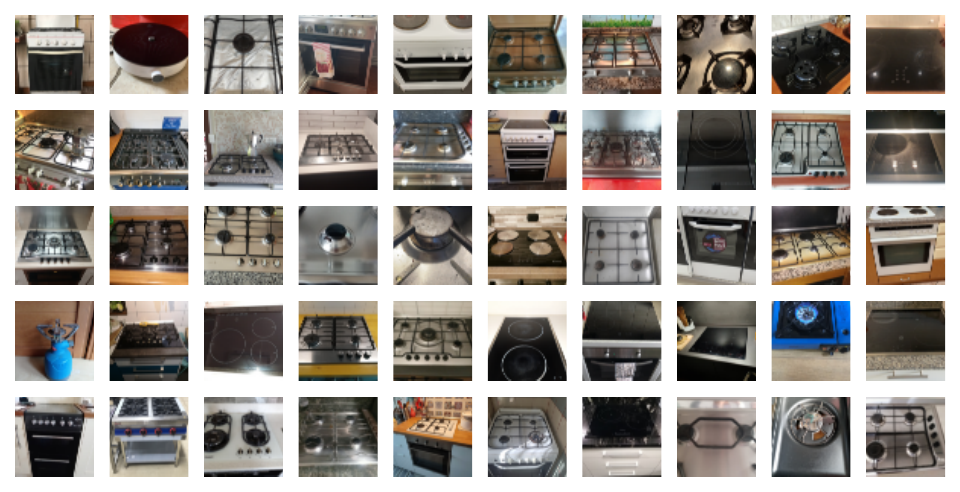}} \\ \midrule  
    \end{tabular}
    \caption{Randomly chosen images for ``stove'' for 3 regions. We see that Southeast Asia contains one-burner and two burner stoves (along with 4 burner stoves). We also see a variety of stoves in terms of induction, coils, gas, ovens, etc. }
    \label{fig:stove2}
\end{figure*}

\begin{figure*}
    \centering
    
    \begin{tabular}{lc}
    \toprule
    \multicolumn{2}{c}{\textbf{Waste container}}\\\toprule
    West Asia & \raisebox{-0.5\totalheight}{\includegraphics[width=0.72\textwidth]{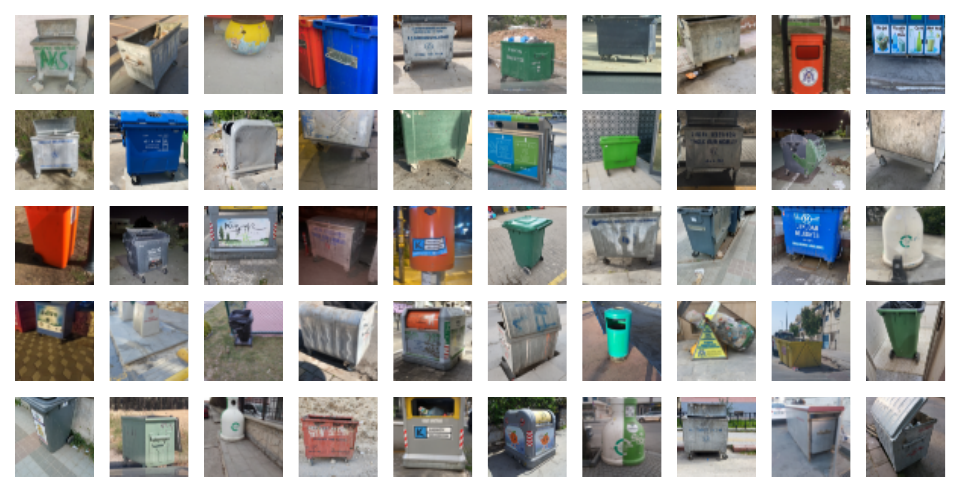}} \\ \midrule
    Africa & \raisebox{-0.5\totalheight}{\includegraphics[width=0.72\textwidth]{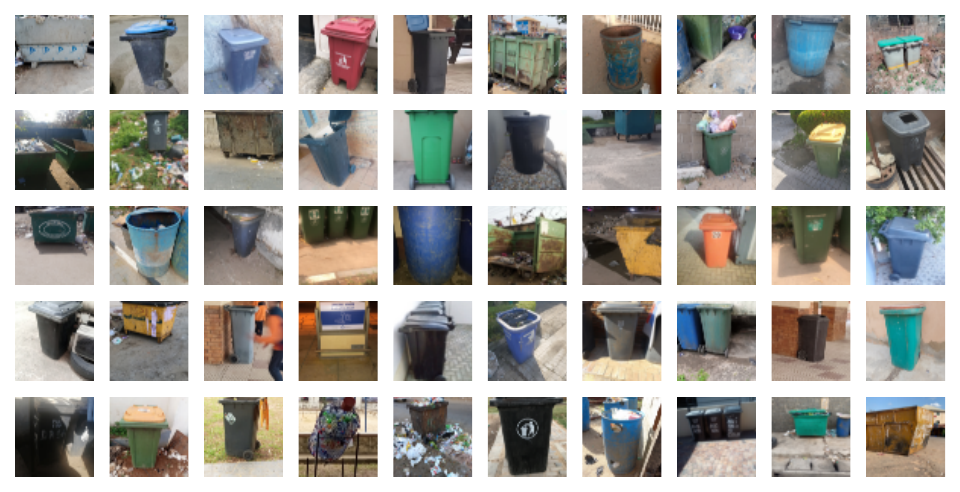}} \\ \midrule
    East Asia & \raisebox{-0.5\totalheight}{\includegraphics[width=0.72\textwidth]{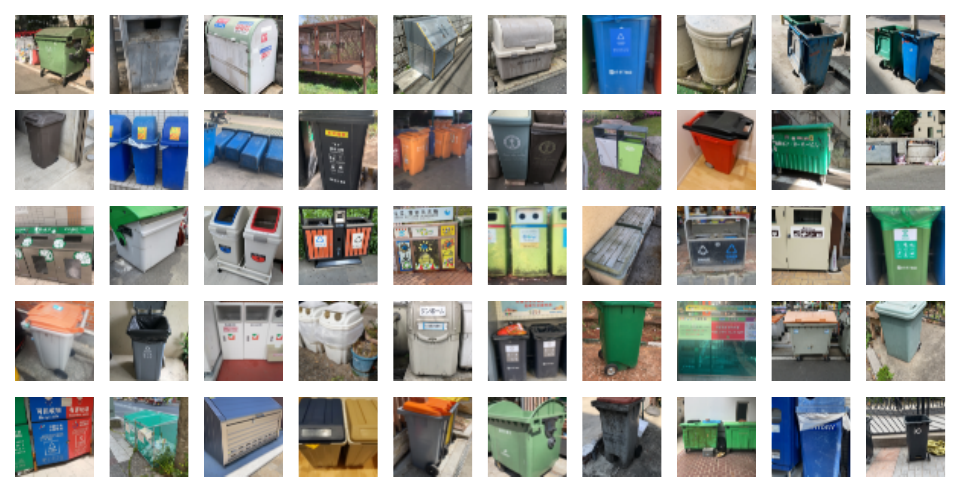}} \\ \midrule  
    \end{tabular}
    
    \caption{Randomly chosen images for ``waste container'' for 3 regions. We see that different regions have containers of varying sizes (Africa seems to be smaller than West Asia or East Asia), and have different closing mechanisms (see West Asia r5c6 as an interesting example.) East Asia also tends to have segregated waste containers. }
    \label{fig:waste_container1}
\end{figure*}

\begin{figure*}
    \centering
    \begin{tabular}{lc}
    \toprule
    \multicolumn{2}{c}{\textbf{Waste containers}}\\\toprule
    Southeast Asia & \raisebox{-0.5\totalheight}{\includegraphics[width=0.72\textwidth]{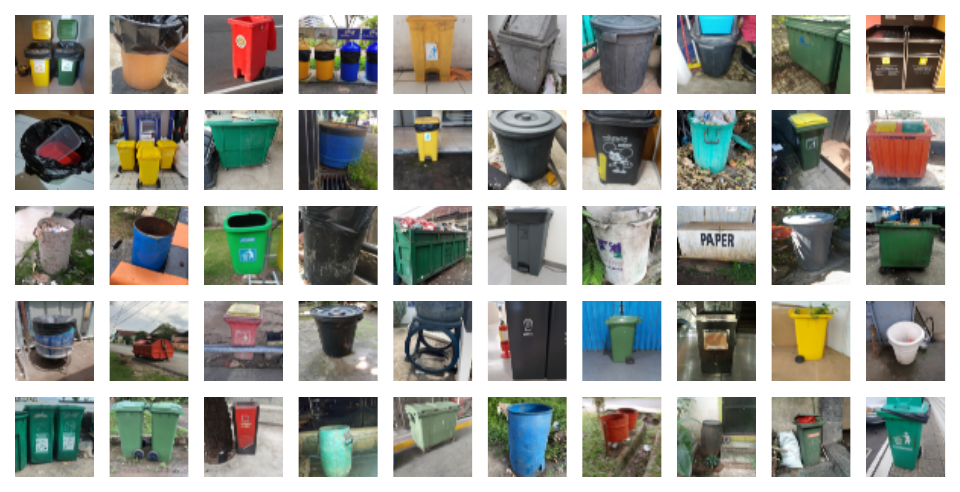}} \\ \midrule
    Americas & \raisebox{-0.5\totalheight}{\includegraphics[width=0.72\textwidth]{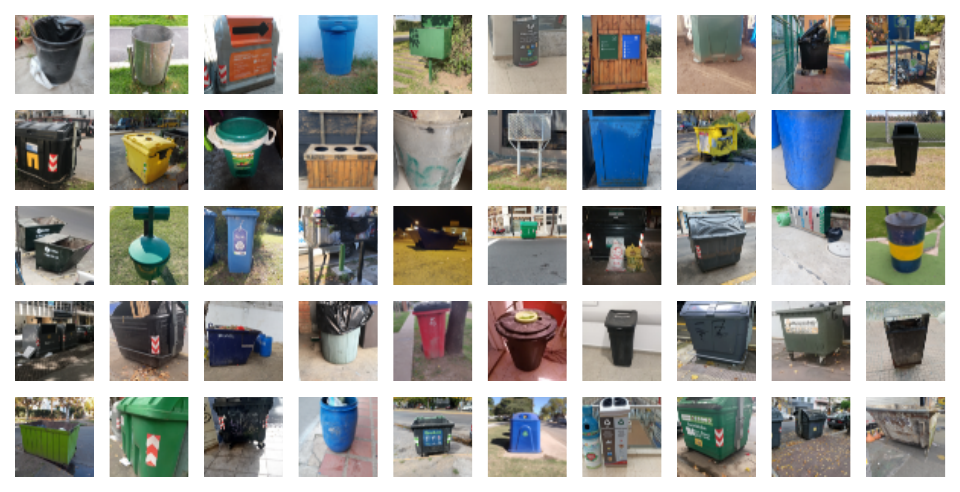}} \\ \midrule
    Europe & \raisebox{-0.5\totalheight}{\includegraphics[width=0.72\textwidth]{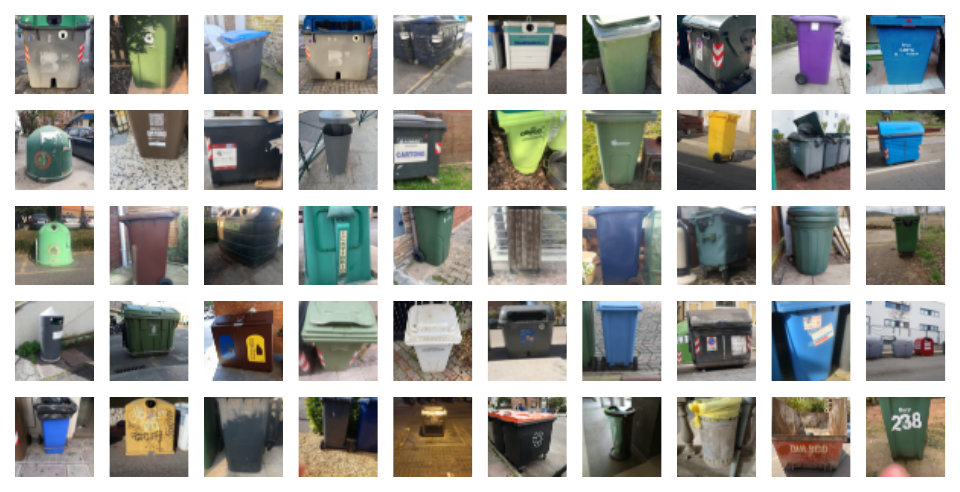}} \\ \midrule  
    \end{tabular}
    \caption{Randomly chosen images for ``waste container'' for 3 regions. We see that different regions have containers of varying sizes (Europe seems to have containers of very different sizes) and have different closing mechanisms (see Southeast Asia r2c6 as an interesting example.) }
    \label{fig:waste_container2}
\end{figure*}

\end{document}